\documentclass[lettersize,journal]{IEEEtran}
\usepackage{amsmath,amsfonts}
\usepackage{algorithmic}
\usepackage{array}
\usepackage[caption=false,font=normalsize,labelfont=sf,textfont=sf]{subfig}
\usepackage{textcomp}
\usepackage{stfloats}
\usepackage{url}
\usepackage{xurl}
\usepackage{verbatim}
\usepackage{graphicx}
\usepackage{cite}
\usepackage{subcaption}
\usepackage{booktabs} 
\usepackage{multirow} 
\usepackage{dsfont}
\usepackage[capitalize,nameinlink]{cleveref}
\usepackage{footnote}
\makesavenoteenv{figure*}
\def\BibTeX{{\rm B\kern-.05em{\sc i\kern-.025em b}\kern-.08em
    T\kern-.1667em\lower.7ex\hbox{E}\kern-.125emX}}
\usepackage{balance}
\begin{document}
\title{Initialization-Free Bundle Adjustment Revisited: \\ A Controlled Experimental Study}
\author{Simon Weber\textsuperscript{*}, Mateo de Mayo\textsuperscript{*}, Je Hyeong Hong, Carl Olsson, Daniel Cremers, Ronald Clark%
\thanks{\textsuperscript{*}Equal contribution.}
\thanks{Simon Weber and Ronald Clark are in the department of Computer Science at the University of Oxford, United Kingdom. E-mail: \{simon.weber,ronald.clark\}@cs.ox.ac.uk.}
\thanks{Mateo de Mayo and Daniel Cremers are with the Technical University of Munich and the Munich Center for Machine Learning, Germany. E-mail: \{mateo.demayo,cremers\}@tum.de.}
\thanks{Je Hyeong Hong is in the department of Electronic Engineering at Hanyang University, South Korea. E-mail: jhh37@hanyang.ac.kr}
\thanks{Carl Olsson is at the Centre for Mathematical Sciences at Lund University, Sweden. E-mail: carl.olsson@math.lth.se} \thanks{This work has been submitted to the IEEE for possible publication. 
Copyright may be transferred without notice, after which this version may 
no longer be accessible.}}

\markboth{}%
{Simon Weber \MakeLowercase{\textit{et al.}}: Initialization-Free Bundle Adjustment Revisited}

\maketitle
\IEEEpubidadjcol

\begin{abstract}
Initialization-free bundle adjustment (InitFree BA) aims to recover camera poses and scene structure directly from image observations, avoiding the geometric initialization stages of conventional structure-from-motion pipelines. Recent methods based on Object-Space Error (OSE) formulations and Variable Projection (VarPro) show encouraging optimization behavior from random camera configurations. However, existing evaluations primarily measure optimization success, leaving unclear whether a low OSE objective yields a valid metric 3D reconstruction. We revisit InitFree BA experimentally through a unified evaluation framework combining a C++ implementation of existing OSE formulations with a Blender-based dataset generator providing exact ground truth and controlled camera configurations and observation densities. Our experiments reveal a previously overlooked optimization--reconstruction gap: projective solutions with similarly low OSE values can lead to substantially different Euclidean reconstructions after metric upgrade. We identify initialization priors, landmark observation density, and metric-upgrade stability as key factors governing reconstruction success. Overall, our results suggest that the main challenge of InitFree BA is not merely minimizing OSE objectives, but obtaining projective reconstructions that admit reliable metric upgrade. We believe that the proposed benchmark, implementation, and analysis establish stronger experimental foundations for future research on initialization-free bundle adjustment, a problem largely unexplored within the computer vision community. Project page is available at \url{https://github.com/simonwebertum/InitFreeBA.git}.
\end{abstract}


\begin{figure*}[t]
\centering
\setlength{\tabcolsep}{4pt}
\renewcommand{\arraystretch}{0.95}
\begin{tabular}{cc}
\includegraphics[width=0.48\textwidth]{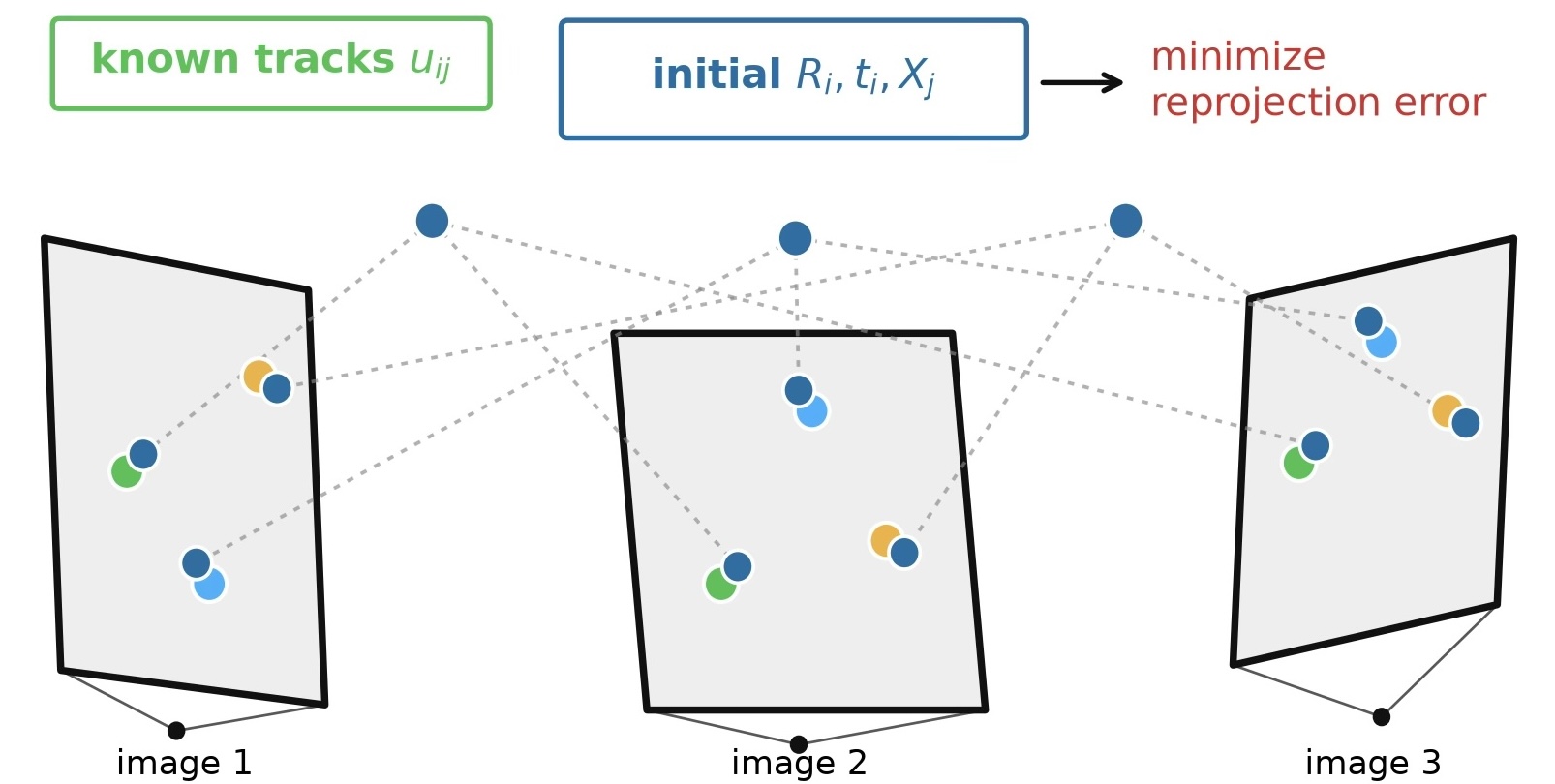} &
\includegraphics[width=0.48\textwidth]{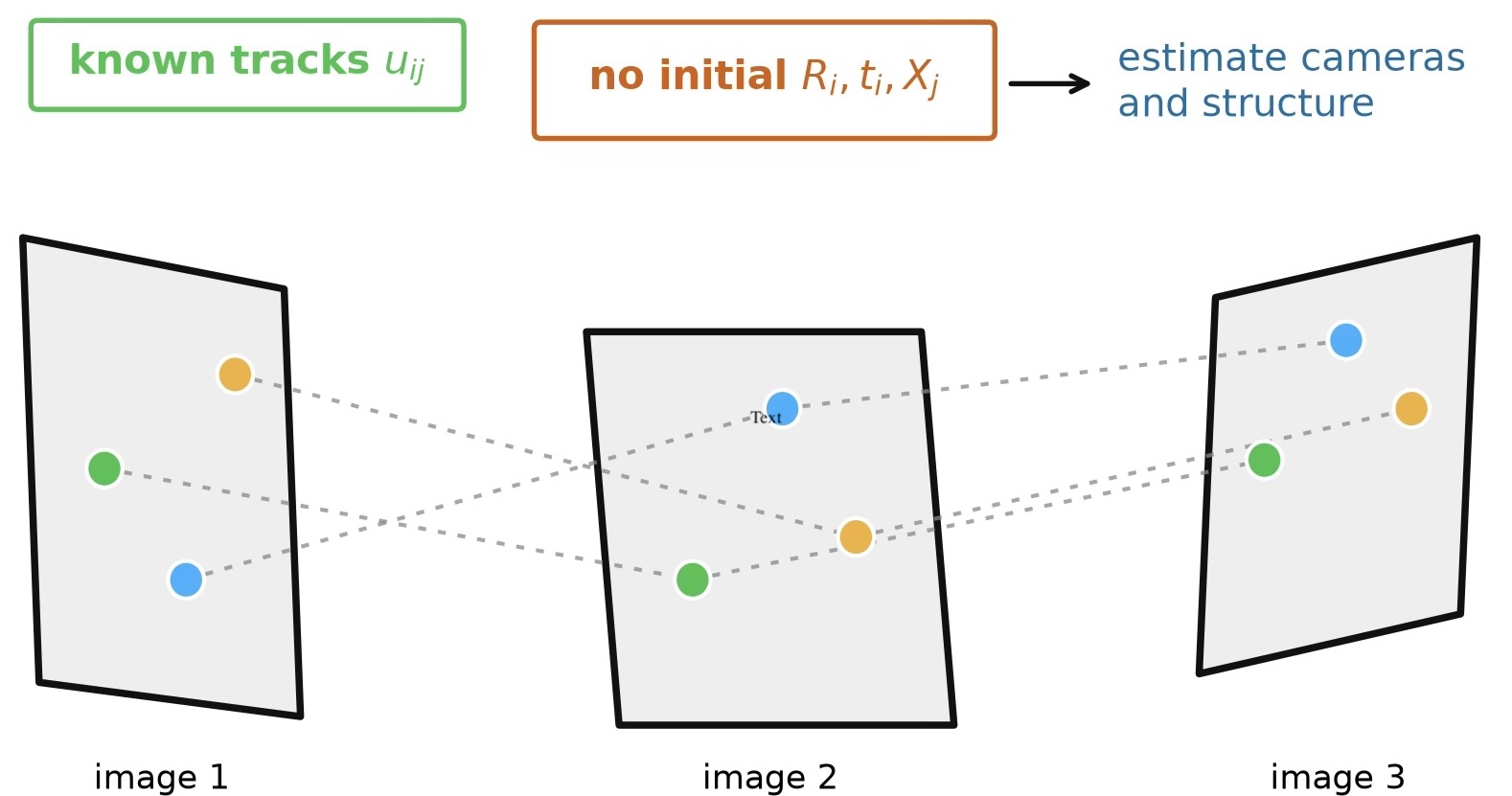}
\\[-1mm]
\small (a) Traditional BA input: observations + initialization &
\small (b) InitFree BA input: observations only
\end{tabular}
\vspace{-1mm}
\caption{\textbf{Traditional BA and initialization-free BA solve different input problems.}
(a) Traditional BA minimizes reprojection error from image observations and
an initial estimate of cameras and landmarks. It is a well-studied problem \cite{triggs1999bundle}. (b) In contrast, initialization-free BA starts only from image observations and must recover cameras and structure without a geometric initialization. This problem is recent \cite{hong2016projective,hong2018pose} and still largely uncharted.}
\label{fig:traditional_vs_initfree_problem}
\vspace{-2mm}
\end{figure*}

\begin{figure*}[t]
\centering
\setlength{\tabcolsep}{4pt}
\renewcommand{\arraystretch}{0.95}
\begin{tabular}{ccc}
\includegraphics[width=0.31\textwidth]{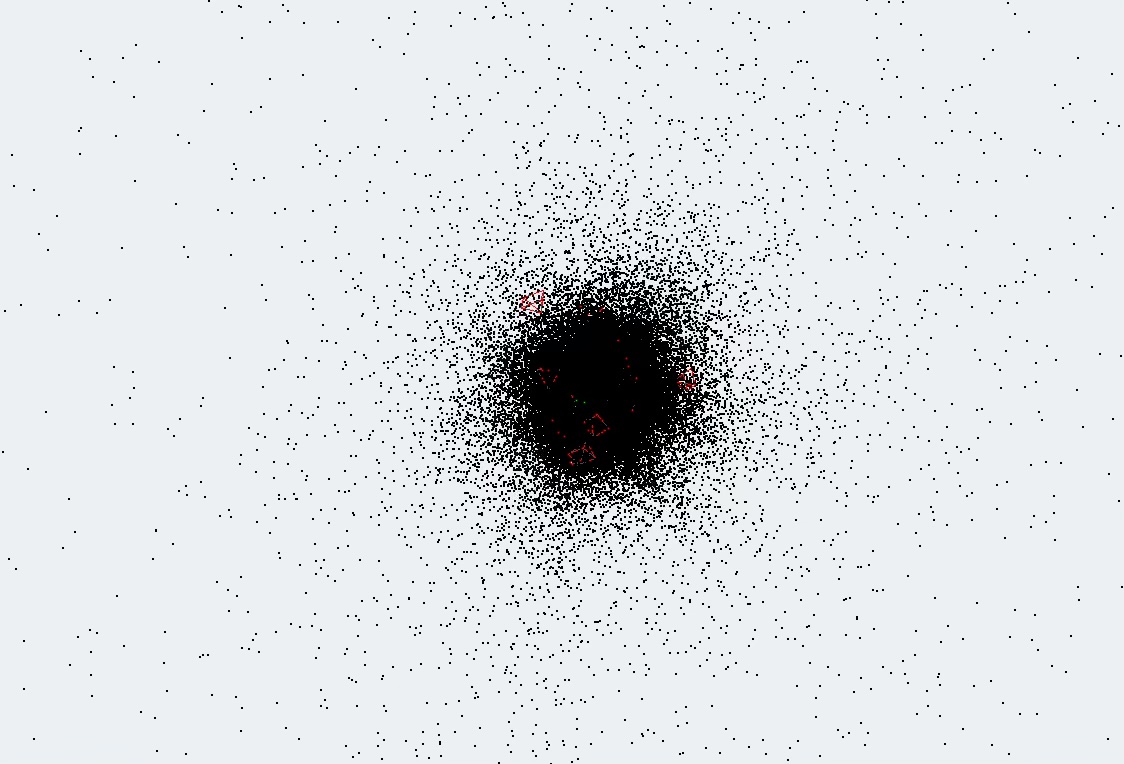} &
\includegraphics[width=0.31\textwidth]{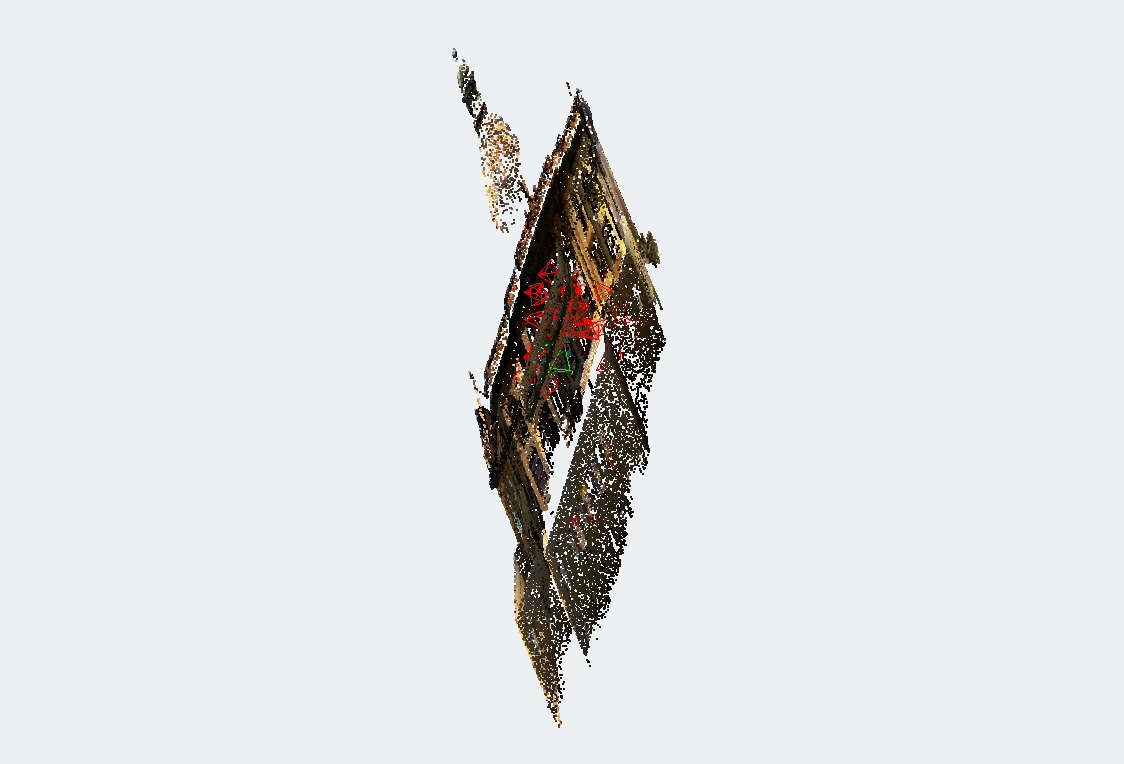} &
\includegraphics[width=0.31\textwidth]{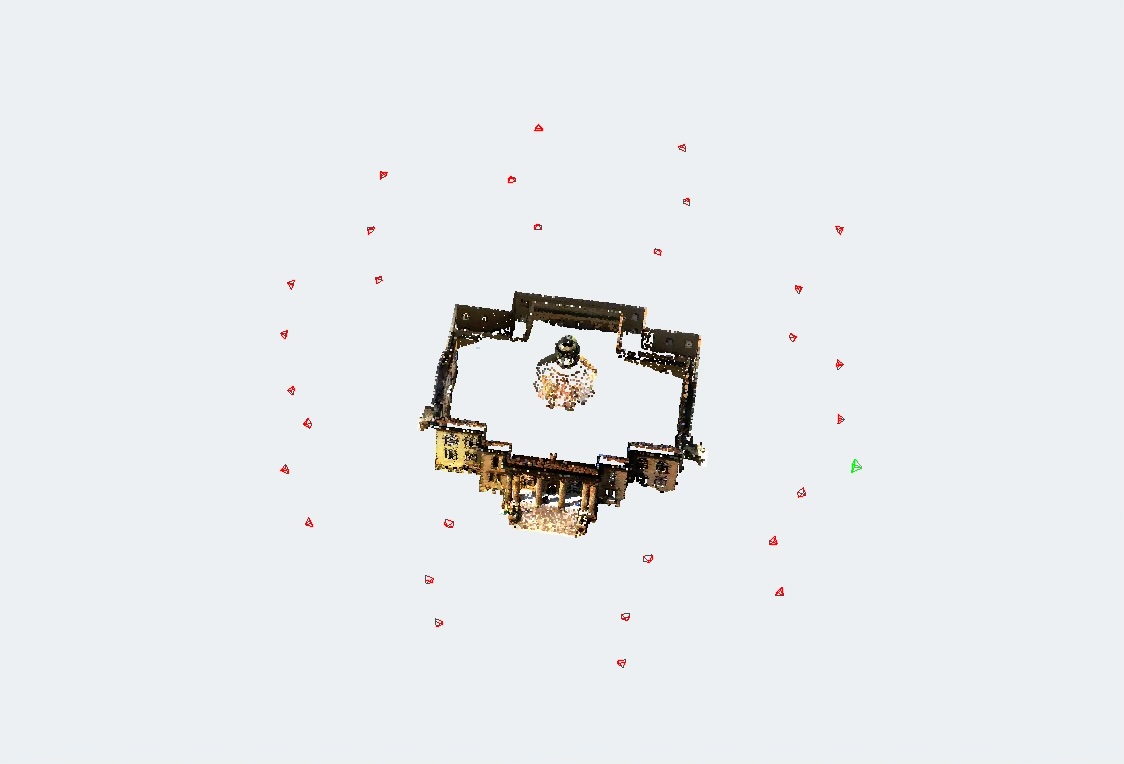}
\\[-1mm]
\small (a) Traditional BA fails &
\small (b) InitFree BA (before metric upgrade) &
\small (c) InitFree BA (after metric upgrade)
\end{tabular}
\vspace{-1mm}
\caption{\textbf{Initialization-free BA can recover structure where randomly initialized traditional BA fails.}
(a) We apply a recent traditional BA solver \cite{weber2023power} randomly initialized on our simple Set3 problem. It collapses to an invalid solution. InitFree BA, here with pOSE \cite{hong2018pose}, instead estimates (b) a projective reconstruction, and can then be upgraded (c) to a metric reconstruction.}
\label{fig:initfree_reconstruction_pipeline}
\vspace{-2mm}
\end{figure*}

\begin{figure*}[t]
\centering
\setlength{\tabcolsep}{2pt}
\renewcommand{\arraystretch}{0.92}
\begin{tabular}{cccc}
\includegraphics[width=0.245\textwidth]{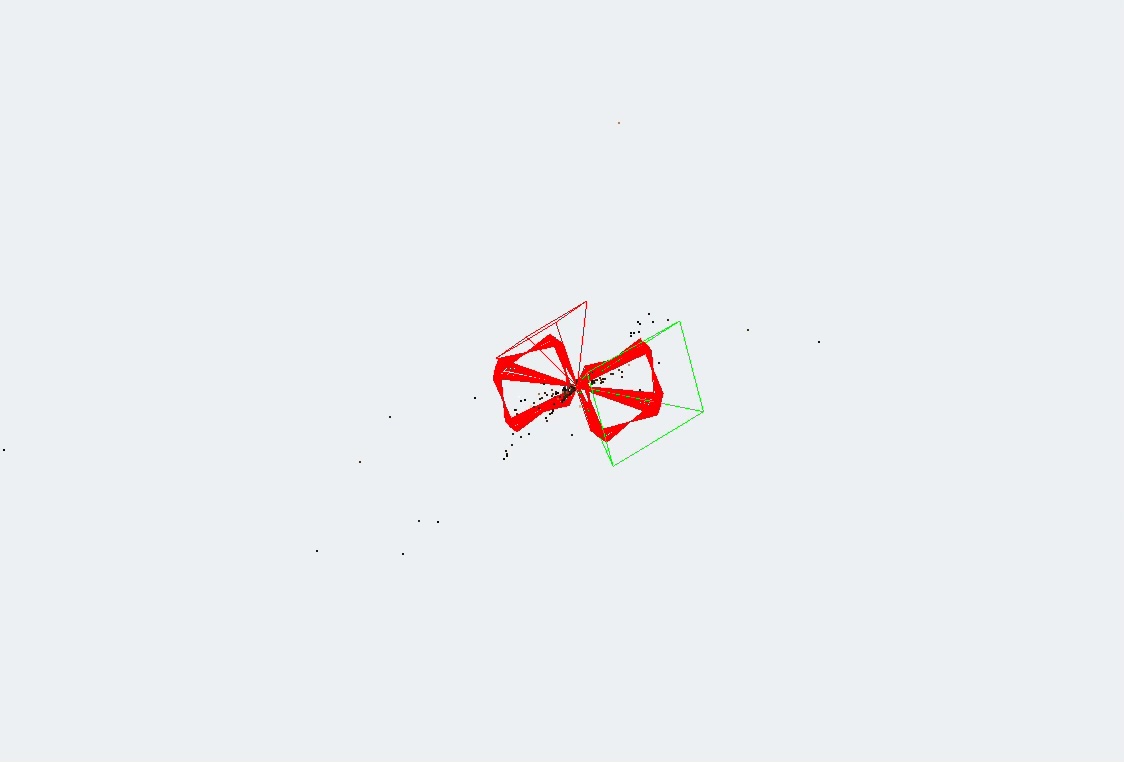} &
\includegraphics[width=0.245\textwidth]{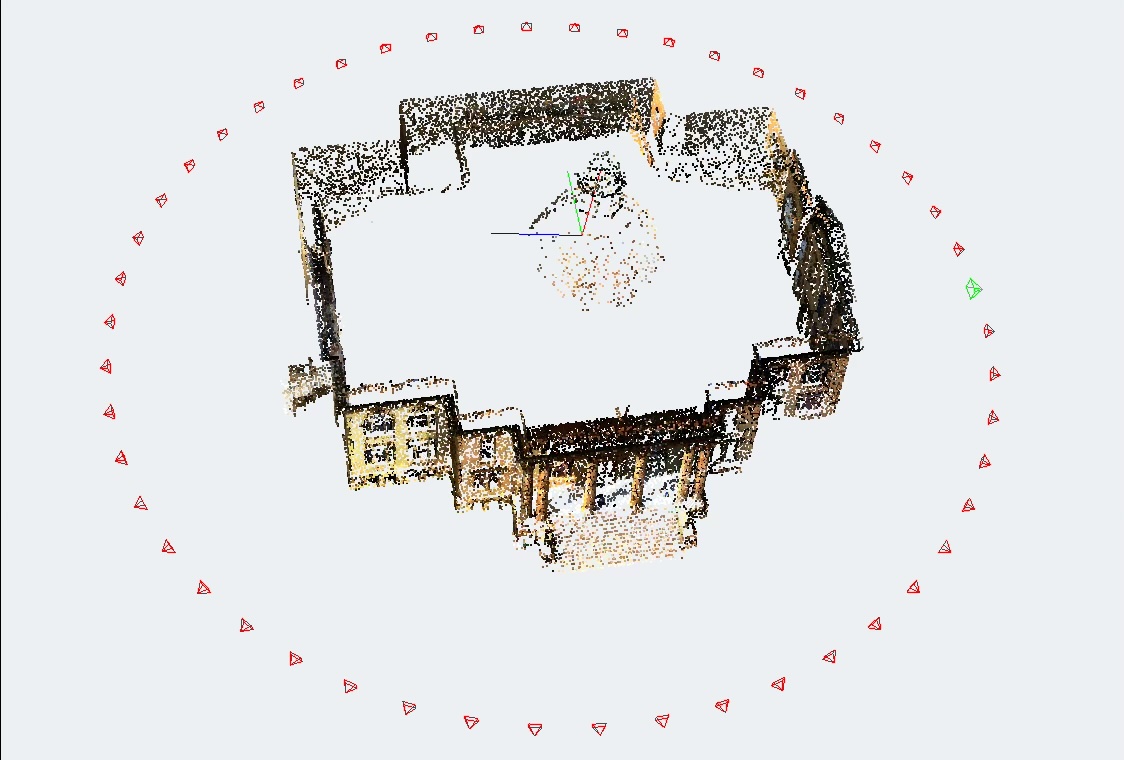} &
\includegraphics[width=0.245\textwidth]{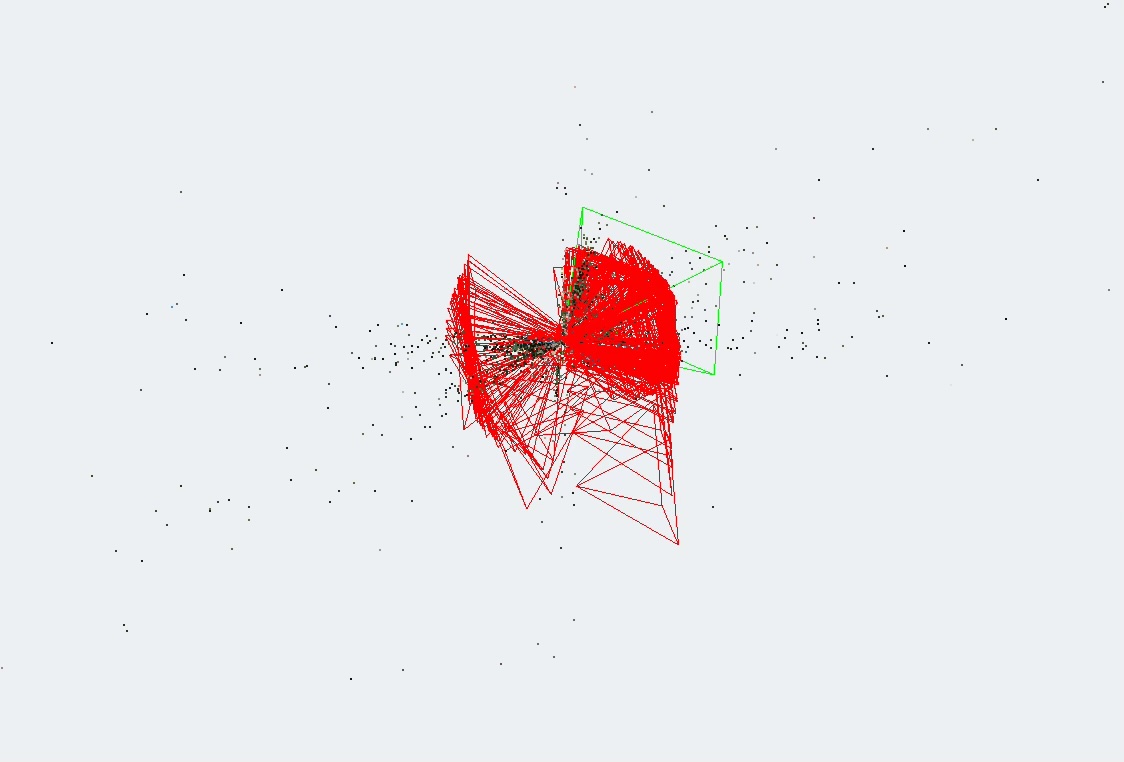} &
\includegraphics[width=0.245\textwidth]{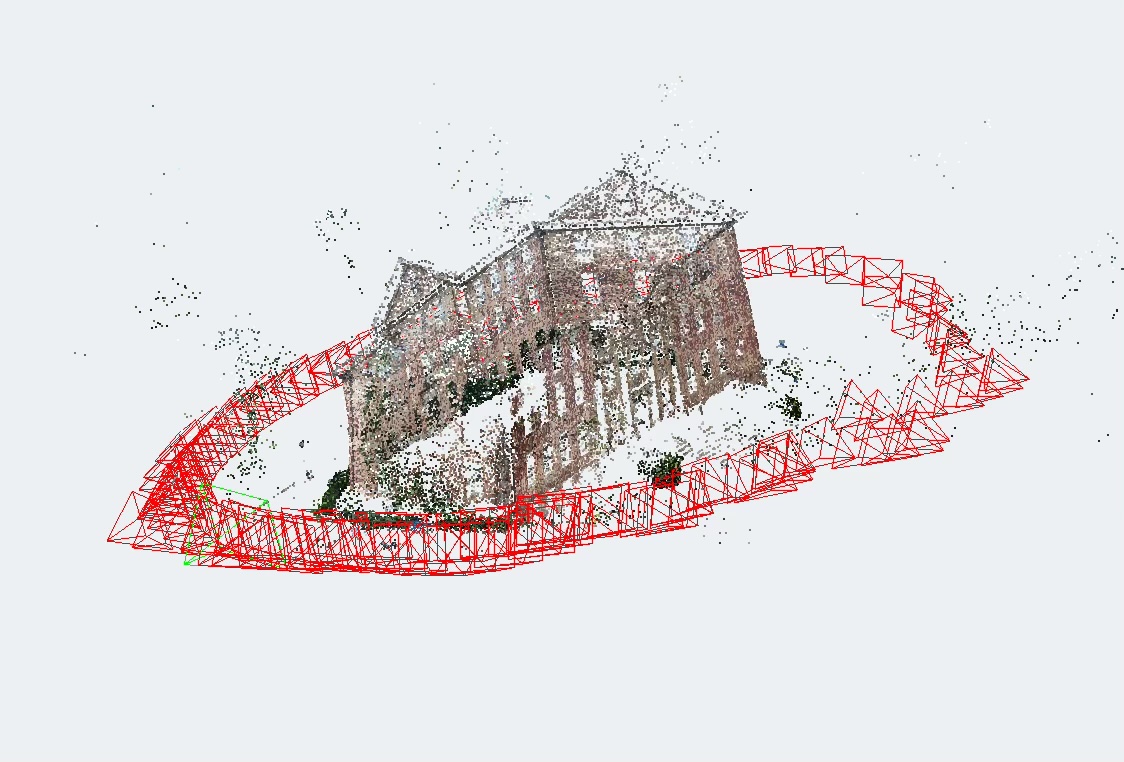}
\\[-1mm]
\small (a) pOSE, Uniform &
\small (b) pOSE, Normal &
\small (c) pOSE+rot, Normal &
\small (d) pOSE+rot, Unit Circle
\\[1mm]
\includegraphics[width=0.245\textwidth]{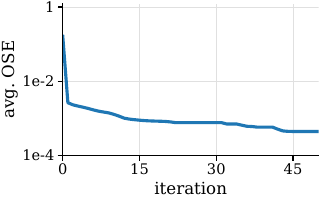} &
\includegraphics[width=0.245\textwidth]{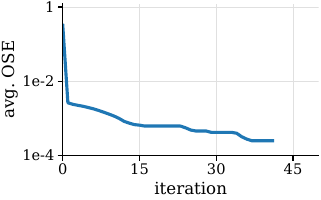} &
\includegraphics[width=0.245\textwidth]{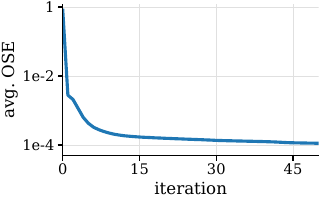} &
\includegraphics[width=0.245\textwidth]{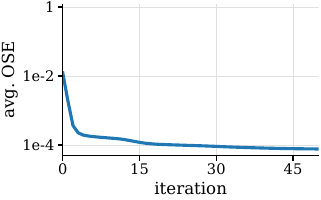}
\end{tabular}
\caption{\textbf{Initialization-free BA is highly sensitive to formulation and initialization, even on a controlled scene.}
Panels show (a), (b) Set2 and (c), (d) Set5 after optimization and linear metric upgrade, with the corresponding normalized OSE objective shown below each reconstruction. Although the OSE objective decreases rapidly and reaches close final errors for each corresponding pair, the resulting Euclidean reconstructions can differ dramatically. This illustrates that a decreasing OSE objective is not by itself a certificate of successful 3D reconstruction.}
\label{fig:teaser_set4}
\end{figure*}


\section{Introduction}
\label{sec:introduction}

Bundle adjustment (BA) is one of the fundamental optimization problems in geometric computer vision. Given image observations, BA jointly estimates camera poses and scene structure by minimizing a geometric reprojection error. In modern structure-from-motion (SfM) systems, BA is typically applied as a refinement stage after several geometric initialization steps, including feature matching, relative pose estimation, rotation averaging, triangulation, and global pose estimation (see \Cref{fig:traditional_vs_initfree_problem}, (a)). Since its introduction, BA has become a cornerstone of large-scale 3D reconstruction systems. Historically, much of the research on BA has focused on scalability. The emergence of Internet-scale image collections motivated the development of sparse linear algebra techniques, distributed solvers, and specialized optimization methods capable of handling increasingly large datasets \cite{agarwal2010bundle}. As a result, classical BA is now supported by mature algorithmic and engineering tools.

More recently, a different question has received increasing attention: can the need for geometric initialization be removed altogether? Initialization-free bundle adjustment (InitFree BA) aims to recover camera poses and scene structure directly from image observations, without relying on the standard SfM initialization pipeline (see \Cref{fig:traditional_vs_initfree_problem}, (b)). In this setting, BA is no longer merely a final refinement stage, but becomes the central optimization problem driving the reconstruction itself. A prominent line of work has revisited the Variable Projection (VarPro) framework and introduced object-space error (OSE) formulations that can be optimized from random camera configurations. Several variants have been proposed, including pOSE \cite{hong2018pose}, rOSE \cite{hong2018pose}, RpOSE \cite{iglesias2021radial}, expOSE \cite{iglesias2023expose}, and pOSE+rot \cite{olsson2025towards}. These methods have demonstrated encouraging optimization behavior and established InitFree BA as a promising research direction. By contrast, our simple experiment in \Cref{fig:initfree_reconstruction_pipeline} illustrates that conventional reprojection-based BA collapses when initialized from random camera configurations. However, a fundamental question remains insufficiently understood:
\begin{quote}
\emph{When does an initialization-free OSE pipeline produce a valid metric 3D reconstruction?}
\end{quote}

\begin{figure}[t]
\centering
\includegraphics[width=\columnwidth]{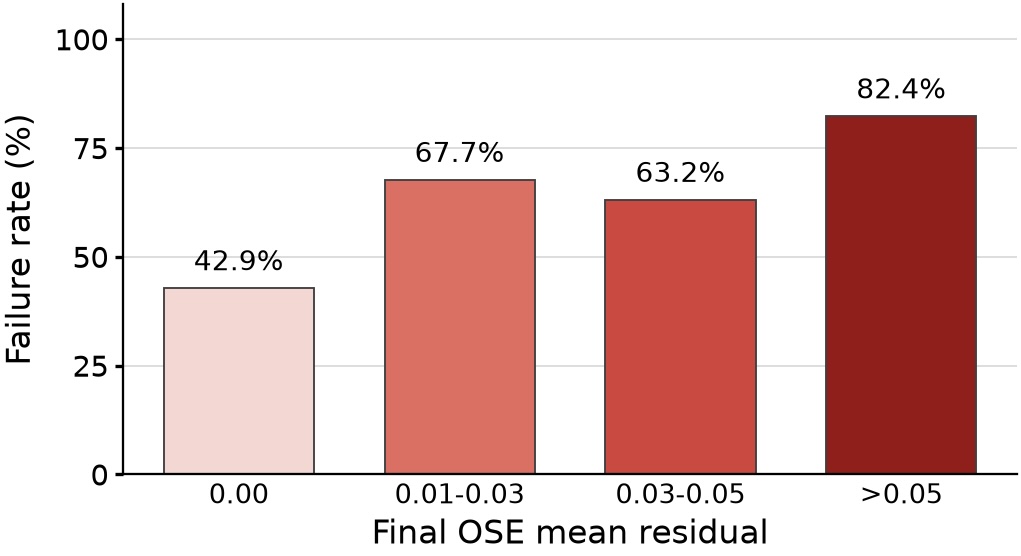}
\caption{\textbf{Metric upgrade failure rate as a function of the final OSE residual.} A run is counted as failed when Rot.~AUC@20 $<20$, Trans.~AUC@20 $<20$, or landmark RMSE $>30$. In contrast to an underlying assumption of previous works, a low OSE error does not necessarily ensure a valid reconstruction.}
\label{fig:metric_upgrade_failure_rate_ose}
\end{figure}

This distinction is important as existing work primarily evaluates optimization performance through objective values and convergence curves. Yet the ultimate goal of InitFree BA is not to minimize an OSE surrogate, but to recover a valid metric 3D reconstruction. Surprisingly, much less is known about the quality of the resulting reconstructions. Whether a low OSE objective actually corresponds to a successful Euclidean reconstruction is a major assumption of existing works. Our controlled experiments show that this assumption does not always hold: a low OSE error does not ensure a valid reconstruction, as it can be seen in \Cref{fig:metric_upgrade_failure_rate_ose}.
In practice, the complete pipeline also depends on landmark estimation, metric upgrade, cheirality, the distribution used to initialize the projective cameras, and the density of the observation graph. Studying this question experimentally is challenging. Real-world SfM benchmarks provide little control over camera trajectories, observation density, or initialization conditions, making it difficult to isolate the factors governing reconstruction success. Conversely, existing synthetic experiments, as in \cite{hong2016projective} for instance, are typically designed to validate individual formulations rather than to systematically analyze the entire reconstruction pipeline. This motivates the need for a controlled experimental framework.

In this paper, we introduce a unified experimental framework for initialization-free bundle adjustment, combining a C++ implementation of existing OSE formulations with a Blender-based dataset generator for controlled evaluation of camera configurations, observation densities, initialization strategies, robustification, and metric upgrade. Using this framework, we uncover an \emph{optimization--reconstruction gap}: similarly low OSE values can lead to substantially different Euclidean reconstructions after metric upgrade (see \Cref{fig:teaser_set4}). Our experiments show that initialization acts as an implicit geometric prior, observation density strongly affects reconstruction stability, and metric upgrade remains sensitive to the recovered projective solution. We further show that robust VarPro can prevent catastrophic failures, although, contrary to robust traditional BA, it does not consistently improve average reconstruction accuracy. We therefore study InitFree BA as a complete reconstruction pipeline rather than objective optimization alone.

The contributions of this paper are as follows:
\begin{itemize}

\item We introduce the first unified experimental framework for initialization-free bundle adjustment, combining existing OSE formulations with a Blender-based dataset generator\footnote{\url{https://github.com/mateosss/colder.git}} for controlled evaluation across camera configurations, observation densities, initialization strategies, and metric-upgrade protocols.

\item We reveal a previously overlooked optimization--reconstruction gap, showing that low OSE objective values do not necessarily imply successful metric reconstruction.

\item We identify the principal factors governing reconstruction success, including initialization priors, observation density, metric-upgrade stability, and robust VarPro, providing practical guidelines and stronger baselines for future InitFree BA research.

\end{itemize}

\section{Related Work}
\label{sec:related}

Bundle adjustment has a long history, originating in photogrammetry in the mid-nineteenth century and later formalized by Schmid~\cite{schmid1959allgemeine} and Brown~\cite{brown1958solution}. We focus here on initialization-free bundle adjustment and its main optimization strategy, Variable Projection. For broader reviews of classical and large-scale BA, we refer to~\cite{triggs1999bundle,weber2025krylov}.

\paragraph{Variable Projection}
Variable Projection (VarPro) exploits separable nonlinear least-squares structure by eliminating one block of variables through its conditional minimizer,
\[
    v^\star(u) = \arg\min_v f(u,v),
\]
and optimizing the reduced objective $f(u,v^\star(u))$ over $u$. Introduced by Golub and Pereyra~\cite{golub1973differentiation}, VarPro has since been extended through approximations of the reduced Jacobian~\cite{ruhe1980separable} and applied to regularized and nonsmooth inverse problems, large-residual problems, and neural-network training~\cite{van2021variable,espanol2023variable,chen2025variable,newman2020train}. 

\paragraph{Initialization-Free Bundle Adjustment}
VarPro has been explored in computer vision as a means to enlarge convergence basins in factorization and geometric estimation~\cite{okatani2011dampedwiberg,strelow2012l1wiberg,strelow2012l2wiberg,hong2017revisiting}. For initialization-free BA, Hong et al.~\cite{hong2016projective} studied projective BA with VarPro, followed by pOSE~\cite{hong2018pose}, which preserves a bilinear camera--point structure suitable for variable elimination. Subsequent formulations include RpOSE for radial distortion~\cite{iglesias2021radial}, expOSE with improved depth behavior~\cite{iglesias2023expose}, and pOSE+rot, which incorporates relative rotations in the calibrated setting~\cite{olsson2025towards}. Power Variable Projection~\cite{weber2024power} instead addresses scalability through a power-series approximation of the reduced system, extending PoBA \cite{weber2023power}. These methods differ in the auxiliary information they assume. We use \emph{initialization-free BA} to denote methods that are not given initial global camera poses or landmark positions; calibration, relative rotations, or metric-upgrade constraints may still be used. Existing work has primarily evaluated objective decrease and convergence from random initialization. In contrast, we investigate whether the resulting projective solution admits a valid metric reconstruction: during this work, and contrary to the underlying assumption in the previous works, we find that a low OSE objective does not necessarily imply geometrically meaningful cameras and landmarks (\Cref{fig:teaser_set4}). Moreover, except for the pOSE implementation in PoVar~\cite{weber2024power}, most existing methods lack public implementations, slowing down the future research on this problem. This motivates our unified evaluation framework.

\section{The InitFree BA Problem}
\label{sec:problem}
For bundle adjustment, existing initialization-free approaches rely on an \emph{object-space error} formulation. Rather than minimizing reprojection errors directly, OSE methods formulate residuals in the 3D object space and augment them with regularization terms that prevent trivial solutions. Over the last few years, several variants have been proposed, each introducing different regularization strategies and geometric constraints (see \Cref{subsec:ose}). All these InitFree BA methods are built upon the observation that the residual can be reformulated as a separable nonlinear least-squares problem,
\begin{align}\label{eq:residual_varpro}
\varepsilon(u,v) = G(u)v - z(u),
\end{align}
where the variables $u$ and $v$, representing respectively the poses and the landmarks, enter the residual nonlinearly and linearly, respectively. Such a structure enables the use of the Variable Projection algorithm (see \Cref{subsec:varpro}). In many applications, VarPro exhibits significantly larger basins of convergence than conventional Levenberg--Marquardt optimization solving for all variables jointly \cite{hong2017revisiting}.

\subsection{Object-Space Error Formulations}\label{subsec:ose}

Let $P_i \in \mathbb{R}^{3 \times 4}$ denote the projective camera of image $i$, let $\tilde{X}_j \in \mathbb{R}^4$ be the homogeneous coordinates of landmark $j$, and let $m_{ij} \in \mathbb{R}^2$ be the corresponding image observation. The object-space error replaces the reprojection residual by the object-space residual
\begin{equation}\label{eq:OSE}
r_{ij}^{\text{ose}}
=
P_{i,1:2}\tilde{X}_{j}
-
(p_{i,3}^{\top}\tilde{X}_{j})m_{ij},
\end{equation}
where $P_{i,1:2}$ contains the first two rows of $P_i$, and $p_{i,3}^{\top}$ its last row. This residual enforces consistency between the projected point and the observed image ray while preserving a bilinear structure in the camera and landmark variables, which makes it suitable for VarPro.
As discussed by Iglesias et al.~\cite{iglesias2021radial}, \Cref{eq:OSE} can be interpreted as a first-order approximation of the reprojection error. Writing
$\lambda = p_{i,3}^{\top}\tilde{X}_j$ and $z = P_{i,1:2}\tilde{X}_j$, the reprojection term $z/\lambda - m_{ij}$ is linearized around an equilibrium $(\tilde{\lambda},\tilde{z})$:
\begin{align}\label{eq:approximation_OSE}
    \frac{1}{\lambda} z
    \approx
    \frac{1}{\tilde{\lambda}}\tilde{z}
    + \frac{1}{\tilde{\lambda}}(z-\tilde{z})
    - \frac{1}{\tilde{\lambda}^{2}}\tilde{z}(\lambda-\tilde{\lambda}) .
\end{align}
Choosing the equilibrium to satisfy $(\tilde{z},\tilde{\lambda})=(m_{ij},1)$ yields the OSE residual in \Cref{eq:OSE}. However, this surrogate alone admits degenerate solutions, including the null solution, and existing formulations therefore add complementary constraints on depth, affine structure, radial geometry, or relative camera rotations.

\paragraph{pOSE and rOSE}
The original pOSE formulation \cite{hong2018pose} combines the object-space residual with an affine regularization term. This regularizer prevents trivial affine degeneracies and is weighted by a parameter $\alpha$. The rOSE variant \cite{hong2018pose} replaces this affine regularization by a depth-normalization constraint, encouraging landmarks to remain at a fixed normalized depth. Both methods preserve the VarPro structure and differ only in the regularization used to stabilize the projective reconstruction.

\paragraph{RpOSE and expOSE}
RpOSE \cite{iglesias2021radial} modifies the object-space residual to better account for radial geometry. This is particularly useful when the projection model includes radial distortion, while still retaining the separable structure required by VarPro. expOSE \cite{iglesias2023expose} instead introduces an exponential residual designed to better approximate perspective reprojection geometry and to reduce the bias induced by the original pOSE formulation.

\paragraph{pOSE+rot}
A more recent formulation incorporates additional rotation information into the OSE framework \cite{olsson2025towards}. In pOSE+rot, the cameras are constrained not only through image observations, but also through pairwise relative rotations. This extra information reduces the ambiguity of the projective reconstruction and improves the conditioning of the metric upgrade. In our implementation, we evaluate both relative rotations obtained from ground truth and relative rotations estimated from image correspondences (see Appendix). 

\paragraph{Common structure}
All these formulations can be written as weighted least-squares objectives of the form
\begin{equation}\label{eq:common_OSE}
\min_{P,X}
\sum_{i,j}
\left\|
\begin{bmatrix}
\sqrt{1-\alpha}\, \phi_{ij}(P_i,\tilde{X}_j,m_{ij})\\
\sqrt{\alpha}\, \psi_{ij}(P_i,\tilde{X}_j,m_{ij})
\end{bmatrix}
\right\|_2^2
+
\mathcal{R}_{\text{rot}},
\end{equation}
where $\phi_{ij}$ denotes the main OSE residual, $\psi_{ij}$ is a method-dependent regularization term, and $\mathcal{R}_{\text{rot}}$ is present only for rotation-constrained variants. $\phi_{ij}$ and  $\psi_{ij}$ are both nonlinear and separable in $P_{i}$ and $\tilde{X}_j$. The important point for this paper is not the specific algebraic form of each residual, but the fact that all these objectives define surrogate optimization problems whose minima must still be converted into a valid metric reconstruction. Due to the separable nonlinear structure of \Cref{eq:common_OSE}, the VarPro algorithm, with its wide convergence basin, is applied. 

\subsection{Variable Projection Algorithm}
\label{subsec:varpro}
The variable projection algorithm proceeds in two stages \cite{hong2017revisiting}. First, exploiting the separable structure of \Cref{eq:residual_varpro}, the linear variables $v$ can be eliminated in closed form. For a fixed value of $u$, the optimal solution is given by
\begin{equation}\label{eq:v}
\begin{split}
v^{*}(u)
&= \operatorname*{arg\,min}_{v}
   \lVert G(u)v - z(u) \rVert_{2}^{2}\\
&= G(u)^{\dagger}z(u) .
\end{split}
\end{equation}
where $G(u)^{\dagger}$ denotes the Moore--Penrose pseudoinverse of $G(u)$. Substituting this solution back into the residual yields the projected problem
\begin{equation}\label{eq:projected_residual}
\begin{split}
\min_{u} \lVert \varepsilon(u,v^{*}(u))\rVert_2^2
&= \min_{u} \lVert \varepsilon^{*}(u)\rVert_2^2\\
&= \min_{u}
\lVert (G(u)G(u)^{\dagger}-I)z(u)\rVert_2^2 .
\end{split}
\end{equation}
The second stage consists of minimizing the projected residual $\varepsilon^{*}(u)$ using the Levenberg--Marquardt (LM) algorithm. This requires the Jacobian of the projected residual with respect to $u$:
\begin{equation}
\begin{split}
J_u^{*}(u)
= \frac{d \varepsilon^{*}(u)}{du}
&= \frac{\partial \varepsilon(u,v^{*}(u))}{\partial v}
   \frac{d v^{*}(u)}{d u}\\
&\quad+
   \frac{\partial \varepsilon(u,v^{*}(u))}{\partial u}\\
&= J_v(u,v^{*}(u))\frac{d v^{*}(u)}{d u}\\
&\quad+J_u(u,v^{*}(u)) .
\end{split}
\end{equation}
By expanding the derivative of $v^{*}$ and applying the Ruhe--Wedin approximation (RW2) \cite{kaufman1975variable}, the projected Jacobian can be approximated as
\begin{equation}\label{eq:RW2}
\begin{split}
J_u^{*}(u)
= {}& \bigl(I - J_{v}(u,v^{*}(u))
J_{v}(u,v^{*}(u))^{\dagger}\bigr)\\
&\cdot J_u(u,v^{*}(u)) .
\end{split}
\end{equation}
where $J_v$ and $J_u$ denote the Jacobians of \Cref{eq:residual_varpro} with respect to $v$ and $u$, respectively. Let us highlight that by noting $Q_{v}(u) = I - J_{v}J_{v}^{\dagger}$, the following property holds:
\begin{equation}\label{eq:property}
    Q_{v}(u) Q_{v}(u) = Q_{v}(u) \,.
\end{equation}
It follows that the resulting LM step is
\begin{equation}\label{eq:lm_varpro}
\begin{split}
\bigl(J_{u}^{\top}(I-J_{v}J_{v}^{\dagger})J_{u}
+\lambda I\bigr)\Delta u
= -J_{u}^{*\top}\varepsilon .
\end{split}
\end{equation}
where all quantities are evaluated at $(u,v^{*}(u))$. Once $u$ is updated, $v$ is given by the closed-form equation \cref{eq:v}.
\paragraph{Solving the VarPro linear system}
Although VarPro eliminates the linear variables analytically, each LM iteration still requires solving the linear system above. In contrast to classical bundle adjustment, for which highly optimized solvers have been developed over several decades, little attention has been devoted to the design of dedicated solvers for VarPro. Weber et al.~\cite{weber2024power} adapted the power-series Schur complement framework originally proposed for large-scale BA \cite{weber2023power} and demonstrated competitive performance in terms of both speed and accuracy when solving the VarPro system
In contrast, preconditioned conjugate gradient methods, the golden standard for solving traditional BA, have been reported \cite{hong2015secrets} to perform poorly with VarPro.

\subsection{Refinement and Metric Upgrade}
\label{subsec:metric_upgrade}

After minimizing the separable non-linear least-squares objective
(\Cref{eq:common_OSE}) with VarPro, the recovered cameras and landmarks
define a projective reconstruction. This reconstruction can optionally be
refined by minimizing the classical reprojection error while retaining the
projective camera parameterization. However, the resulting cameras remain
defined up to a global projective transformation that recovers the intrinsics $K_i$, the rotation $R_i \in SO(3)$, and the translation $t_i \in \mathbb{R}^{3}$.

To obtain a Euclidean reconstruction, we estimate an ambiguity transform
$H \in \mathbb{R}^{4 \times 4}$ such that
\begin{equation}
    P_i H
    \approx
    K_i
    \begin{bmatrix}
    R_i & t_i
    \end{bmatrix}.
\end{equation}
We use the self-calibration approach of Pollefeys et
al.~\cite{pollefeys1999self,pollefeys2004visual}, which estimates $H$
through constraints on the absolute dual quadric $\Omega^{*}$,
\begin{equation}
\label{eq:qac}
    K_i K_i^{\top}
    \sim
    P_i \Omega^{*} P_i^{\top}.
\end{equation}
Under assumptions on the intrinsic parameters, these constraints yield a
linear estimate of the projective-to-metric transformation. We also
consider the nonlinear refinement proposed by Pollefeys et al. \cite{pollefeys1999self}, which
refines this transformation while keeping the projective reconstruction
fixed.

This metric upgrade is a critical stage of InitFree BA: a low OSE value
does not guarantee that the recovered projective cameras admit a stable
Euclidean upgrade (\Cref{fig:teaser_set4}). As our experiments show,
upgrade failures can therefore produce invalid metric reconstructions even
when the preceding OSE optimization appears successful.

\subsection{The Role of the Initialization Prior}
Existing InitFree BA methods typically initialize camera variables from a normal distribution. While these methods avoid scene-specific geometric initialization, the initialization distribution itself defines a prior over the camera parameters. Thus, if reconstruction quality changes substantially with this distribution, InitFree BA should not be regarded as initialization-independent, despite avoiding the classical SfM initialization pipeline. We evaluate this sensitivity using three generic initialization strategies: normal and uniform sampling of projective camera parameters, and a unit-circle configuration in which cameras are placed around an arbitrary origin and oriented toward it. The latter uses no information from the target scene but introduces a simple geometric prior. Comparing these strategies therefore tests whether reconstruction depends on the initialization prior rather than on scene-specific initialization. This distinction suggests a more precise interpretation of InitFree BA: avoiding problem-specific geometric initialization while remaining robust to generic, dataset-independent camera priors. To study this under independently controlled scene and initialization conditions with exact ground truth, we introduce the dataset generator described next.

\section{Dataset Generator}\label{sec:datagen}

A recurring obstacle in the analysis above was identifying and isolating failure cases: real-world SfM datasets offer no ground truth for the intermediate, unoptimized states of a reconstruction, and only a coarse and often noisy reference for the final one. To study InitFree BA under controlled conditions, we therefore build a controlled dataset generator that lets us progressively degrade data quality and observe its effect on the solvers, while retaining exact ground-truth cameras, landmarks, and observations, something particularly difficult to obtain from real-world data. The generator is designed to be fast and flexible, exposing a large number of parameters so that we can closely model the properties of real SfM datasets. It is built on top of Blender \cite{blender}, using the EEVEE rendering engine for speed and realism, and exports directly to the COLMAP \cite{schoenberger2016sfm,pan2024glomap} format for ease of use with current and future pipelines.

\begin{figure*}[t]
\centering
\includegraphics[width=\textwidth]{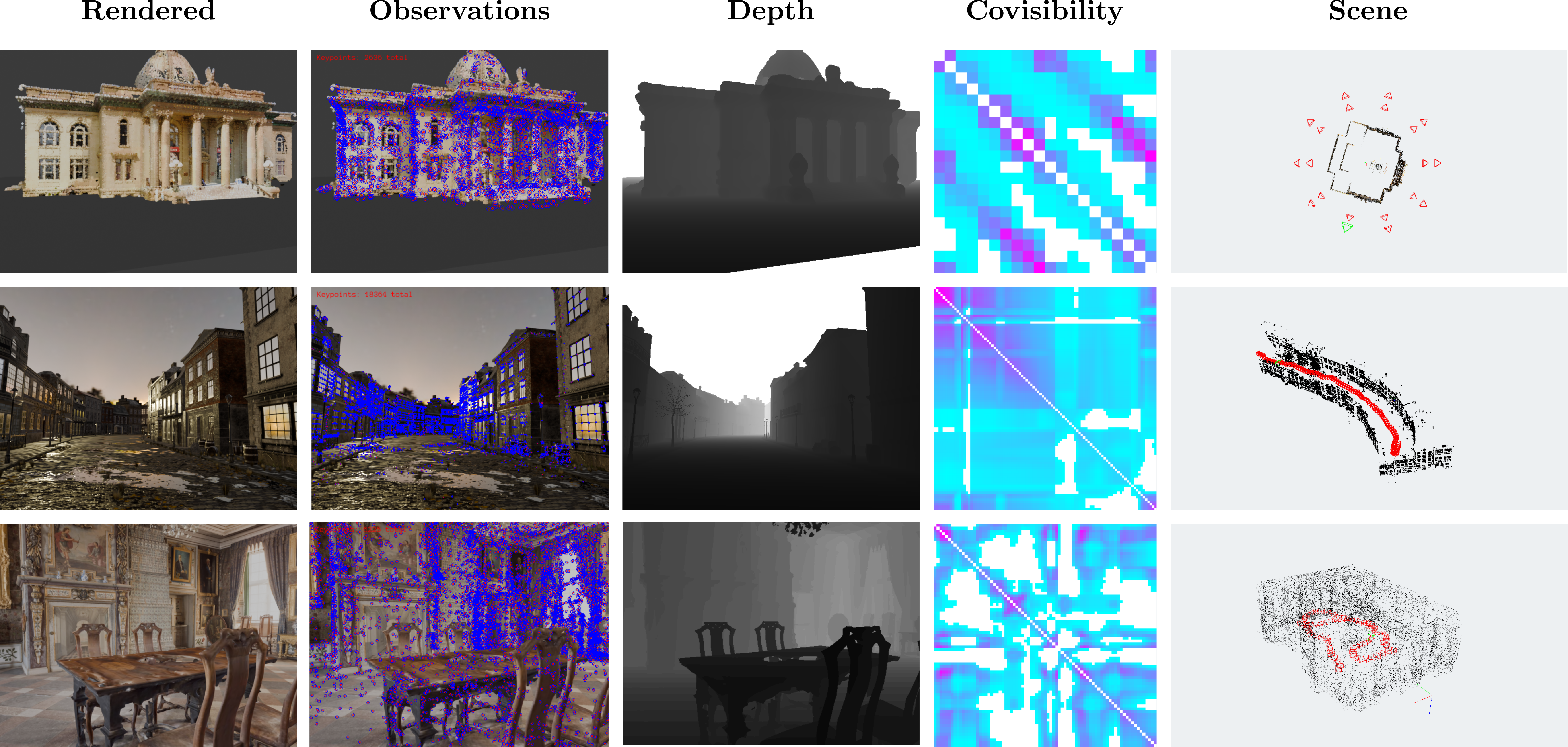}
\caption{\textbf{Overview of the dataset generator.} Each row shows a rendered image, 2D observations (blue) and projected 3D landmarks (red), depth map, covisibility matrix, and reconstructed points and cameras. From top to bottom: Set~6; an artist-made realistic scene\protect\footnote{Model courtesy of Aur\'elien Martel, licensed CC BY-NC 4.0, Sketchfab.}; and a photogrammetry scan of Skokloster Castle\protect\footnote{Scan courtesy of Skoklosters slott, licensed CC BY 4.0, Sketchfab.}. The examples illustrate that the generator supports both controlled and realistic scene geometries. Our generator is not restricted to InitFree BA problem, and is a reusable tool for controlled evaluation in SfM, SLAM, and related 3D reconstruction problems}
\label{fig:datagen}
\end{figure*}

\paragraph{Dataset generator}
Our Blender-based generator creates controlled SfM problems from 3D scenes represented as point clouds or meshes. For each camera, it renders an image and depth map, which can be used to account for landmark occlusions (\Cref{fig:datagen}). Camera intrinsics follow COLMAP's \texttt{SIMPLE\_RADIAL} model, while trajectories and orientations can be freely specified using Blender curves or animations. The generator supports both synthetic and photorealistic scenes and exports observations, cameras, and landmarks directly in COLMAP-compatible format. We will open-source it to facilitate the generation of additional sequences with controlled geometry, visibility, and camera configurations.

\begin{figure*}[t]
\centering
\includegraphics[width=\textwidth]{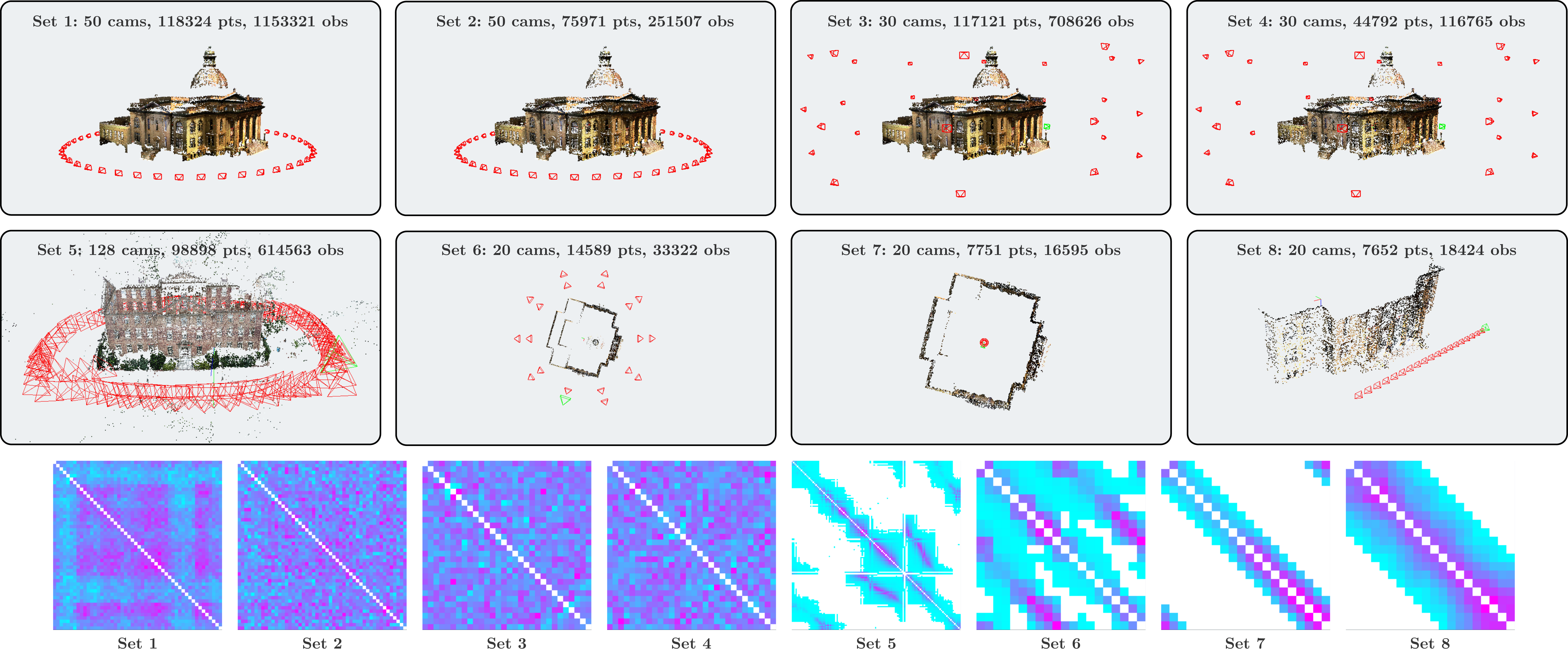}
\caption{\textbf{The eight sequences used in our experiments.} Sets~1--2 and~3--4 test the effect of observation sparsity on convergence, Set~5 is a real-world COLMAP reconstruction used to validate our controlled findings in a realistic setting, and Sets~6--8 test the effect of camera distribution and lookat targets on convergence. Bottom row: covisibility matrix for each sequence, giving an indication of the density of the Schur complement and of how well connected the cameras are (cyan: weakly covisible, magenta: highly covisible; normalized per matrix).}
\label{fig:sets}
\end{figure*}

\paragraph{Experimental sequences}
We evaluate the solvers on eight sequences (\Cref{fig:sets}) spanning different camera configurations and observation densities. Sets~1--2 use the same camera ring while decreasing the average landmark visibility from $\sim$10 to $\sim$3 observations; Sets~3--4 similarly use three perturbed camera rings with visibility decreasing from $\sim$6 to $\sim$3. These pairs isolate the effect of observation sparsity. Sets~6--8 instead evaluate different camera configurations: two coplanar rings looking toward the building center (Set~6), 20 outward-looking cameras inside the building (Set~7), and 20 laterally translating cameras with parallel optical axes (Set~8). These sequences use the ``Courthouse'' point cloud from Tanks and Temples \cite{knapitsch2017tanks}. Set~5 provides a more realistic setting using the ``South Building'' COLMAP reconstruction\footnote{COLMAP datasets: \url{https://demuc.de/colmap/datasets/}} \cite{schoenberger2016sfm}, which we feed directly to the solvers to retain realistic noise and outliers. Because VarPro eliminates landmarks before optimizing cameras, problem difficulty depends not only on the number of observations but also on how strongly landmarks couple cameras in the reduced system. The covisibility matrices in \Cref{fig:sets} provide a visual proxy for this structure.

\section{Experiments}\label{sec:experiments}

\subsection{Metrics}
\label{subsec:metrics}

In contrast to previous works, we evaluate reconstruction quality rather than the final OSE objective. Camera rotations are measured using pairwise relative rotation errors, removing the global rotation gauge, while translations are evaluated using pairwise directions between camera centers, removing global scale. Both are summarized by Area Under Curve (AUC) at $5^\circ$, $10^\circ$, and $20^\circ$. Scene structure is evaluated after similarity alignment \cite{umeyama1991least} using landmark median error, RMSE, and $90$th percentile error. Detailed metric definitions and per-dataset results before and after metric upgrade are provided in the appendix.

\subsection{Implementation}

We build on PoVar~\cite{weber2024power}, the only publicly available VarPro-based InitFree BA implementation to our knowledge, and extend it into a unified C++ framework implementing pOSE~\cite{hong2018pose}, rOSE~\cite{hong2018pose}, RpOSE~\cite{iglesias2021radial}, expOSE~\cite{iglesias2023expose}, and pOSE+rot~\cite{olsson2025towards} using the pOSE affine regularizer as detailed in
the appendix. All formulations share the same solver and evaluation pipeline, enabling comparison under identical numerical conditions. We additionally implement the metric upgrade of Pollefeys et al.~\cite{pollefeys1999self}, with optional nonlinear refinement, to evaluate the resulting Euclidean reconstructions rather than optimization costs alone. We use $\alpha=0.05$ and normalize image observations during OSE optimization before projecting them back to pixel space for refinement. Our framework will be released to facilitate reproducible evaluation of InitFree BA methods.

\begin{table}[t]
\centering
\caption{\textbf{Average pose accuracy by formulation.}
Results are averaged over all datasets, initialization distributions, and non-robust/Cauchy variants after the linear metric upgrade. We report AUC 5° and AUC 20°, for rotation and translation. Higher is better.}
\label{tab:summary_formulations_pose_linear1997}
\scriptsize
\setlength{\tabcolsep}{4pt}
\renewcommand{\arraystretch}{1.12}
\resizebox{\columnwidth}{!}{%
\begin{tabular}{@{}lcccc@{}}
\toprule
\textbf{Formulation}
& \textbf{Rot. A5}
& \textbf{Rot. A20}
& \textbf{Trans. A5}
& \textbf{Trans. A20} \\
\midrule
pOSE      & 25.43 & 34.25 & 26.43 & 34.10 \\
rOSE      & 24.33 & 31.99 & 28.82 & 37.56 \\
RpOSE     & 9.28  & 15.24 & 12.29 & 19.85 \\
expOSE    & 25.74 & 35.89 & 26.39 & 35.94 \\
pOSE+rot  & \textbf{42.14} & \textbf{53.87} & \textbf{45.80} & \textbf{57.08} \\
\bottomrule
\end{tabular}}
\end{table}

\begin{table}[t]
\centering
\caption{\textbf{Average structure accuracy by formulation.}
Results are averaged over all datasets, initialization distributions, and non-robust/Cauchy variants after the linear metric upgrade. Lower is better.}
\label{tab:summary_formulations_landmarks_linear1997}
\scriptsize
\setlength{\tabcolsep}{5pt}
\renewcommand{\arraystretch}{1.12}
\resizebox{\columnwidth}{!}{%
\begin{tabular}{@{}lccc@{}}
\toprule
\textbf{Formulation}
& \textbf{Lmk. Med.} $\downarrow$
& \textbf{Lmk. RMSE} $\downarrow$
& \textbf{Lmk. P90} $\downarrow$ \\
\midrule
pOSE      & 22.68 & 24.19 & 33.28 \\
rOSE      & 20.52 & 21.82 & 29.92 \\
RpOSE     & 27.93 & 29.58 & 40.66 \\
expOSE    & 19.71 & 21.40 & 29.81 \\
pOSE+rot  & \textbf{15.27} & \textbf{16.63} & \textbf{22.46} \\
\bottomrule
\end{tabular}}
\end{table}

\begin{figure}[t]
\centering
\setlength{\tabcolsep}{2pt}
\renewcommand{\arraystretch}{0.95}
\begin{tabular}{cc}
\includegraphics[width=0.485\columnwidth]{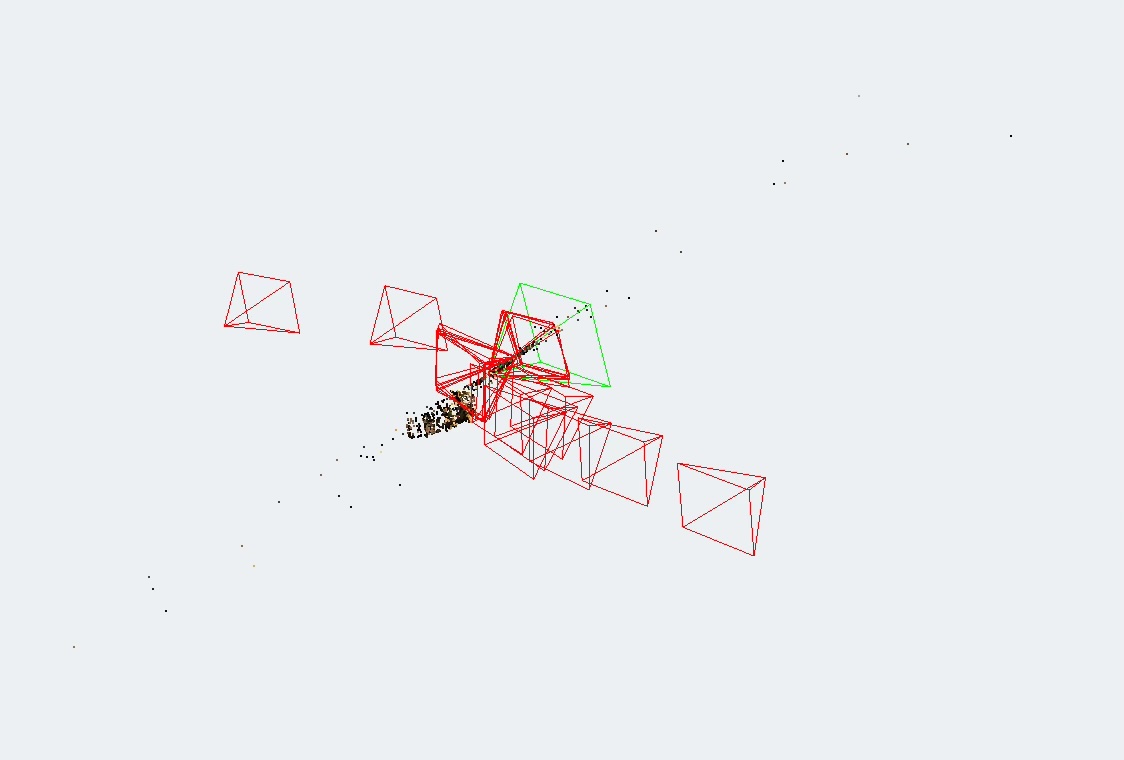} &
\includegraphics[width=0.485\columnwidth]{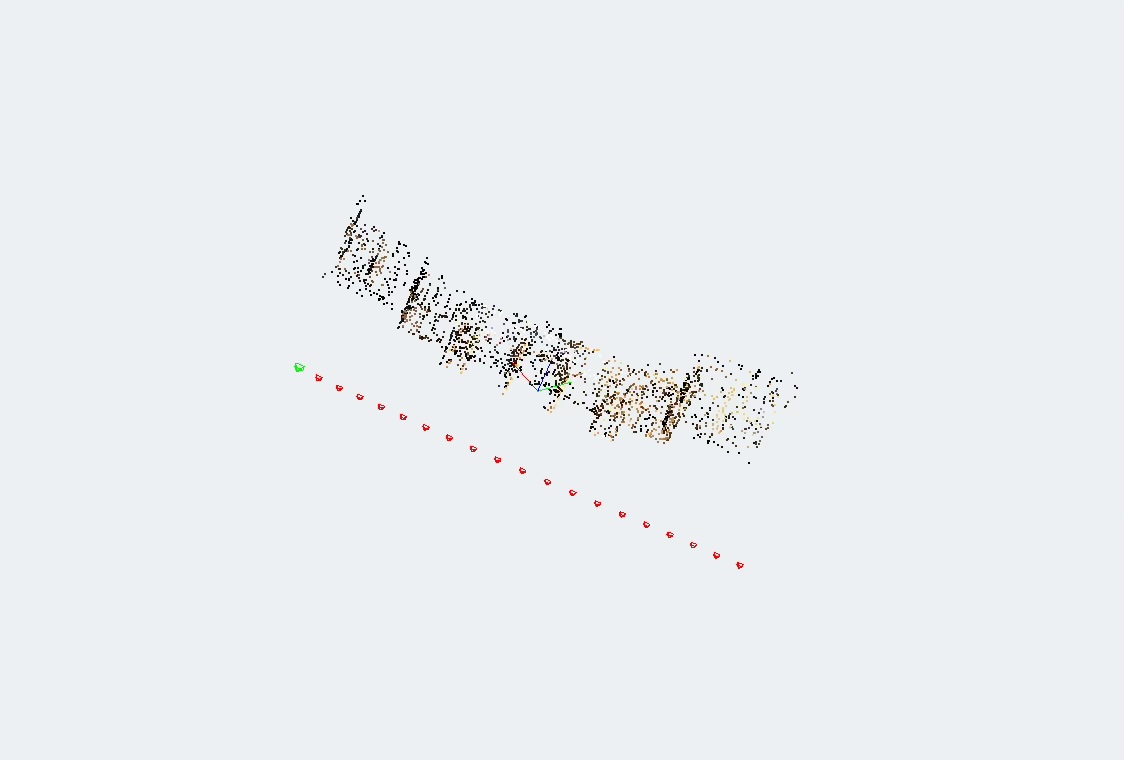}
\\[-1mm]
\scriptsize (a) Normal &
\scriptsize (b) Unit circle
\end{tabular}
\caption{\textbf{Effect of initialization.}
On Set8 with pOSE+rot, normal randomization collapses, whereas unit-circle initialization recovers a coherent reconstruction.}
\label{fig:effect_initialization}
\end{figure}

\begin{table}[t]
\centering
\caption{\textbf{Average effect of initialization.}
Results are averaged over all datasets, formulations, and non-robust/Cauchy variants after the linear metric upgrade. Higher is better for AUC; lower is better for landmark RMSE.}
\label{tab:init_average_linear1997}
\scriptsize
\setlength{\tabcolsep}{4pt}
\renewcommand{\arraystretch}{1.12}
\resizebox{\columnwidth}{!}{%
\begin{tabular}{@{}lccccc@{}}
\toprule
\textbf{Initialization}
& \textbf{Rot. A5}
& \textbf{Rot. A20}
& \textbf{Trans. A5}
& \textbf{Trans. A20}
& \textbf{Lmk. RMSE} $\downarrow$ \\
\midrule
Normal      & 24.23 & 29.60 & 26.39 & 32.21 & 24.63 \\
Uniform     & 23.47 & 30.54 & 25.71 & 33.04 & 24.27 \\
Unit circle & \textbf{28.45} & \textbf{42.60} & \textbf{31.74} & \textbf{45.47} & \textbf{19.27} \\
\bottomrule
\end{tabular}}
\end{table}

\begin{table}[t]
\centering
\caption{\textbf{Effect of initialization and Cauchy robustification on pose recovery.}
Results are averaged over all datasets and formulations after the linear metric upgrade. Each entry reports without Cauchy/with Cauchy, and the best value for each initialization is bolded.}
\label{tab:init_cauchy_pose_linear1997_pairs}
\scriptsize
\setlength{\tabcolsep}{4pt}
\renewcommand{\arraystretch}{1.12}
\resizebox{\columnwidth}{!}{%
\begin{tabular}{@{}lcccc@{}}
\toprule
\textbf{Initialization}
& \textbf{Rot. A5}
& \textbf{Rot. A20}
& \textbf{Trans. A5}
& \textbf{Trans. A20} \\
\midrule
Normal      & \textbf{25.01} / 23.46 & \textbf{30.18} / 29.01 & 26.29 / \textbf{26.49} & 31.64 / \textbf{32.79} \\
Uniform     & \textbf{25.72} / 21.22 & \textbf{32.59} / 28.49 & \textbf{27.35} / 24.08 & \textbf{34.27} / 31.80 \\
Unit circle & 27.95 / \textbf{28.94} & 42.24 / \textbf{42.96} & 31.58 / \textbf{31.89} & \textbf{45.80} / 45.13 \\
\bottomrule
\end{tabular}}
\end{table}

\begin{table}[t]
\centering
\caption{\textbf{Effect of initialization and Cauchy robustification on structure accuracy.}
Results are averaged over all datasets and formulations after the linear metric upgrade. Each entry reports without Cauchy/with Cauchy, and the best value for each initialization is bolded. Lower is better.}
\label{tab:init_cauchy_landmarks_linear1997_pairs}
\scriptsize
\setlength{\tabcolsep}{4pt}
\renewcommand{\arraystretch}{1.12}
\resizebox{\columnwidth}{!}{%
\begin{tabular}{@{}lccc@{}}
\toprule
\textbf{Initialization}
& \textbf{Lmk. Med.} $\downarrow$
& \textbf{Lmk. RMSE} $\downarrow$
& \textbf{Lmk. P90} $\downarrow$ \\
\midrule
Normal      & 22.66 / 23.19 & 24.50 / 24.76 & 33.96 / 34.17 \\
Uniform     & 22.50 / 22.86 & 23.98 / 24.56 & 33.18 / 33.98 \\
Unit circle & \textbf{17.81} / 18.31 & \textbf{18.99} / 19.55 & \textbf{25.60} / 26.48 \\
\bottomrule
\end{tabular}}
\end{table}

\begin{figure}[t]
\centering
\setlength{\tabcolsep}{2pt}
\renewcommand{\arraystretch}{0.95}
\begin{tabular}{cc}
\includegraphics[width=0.485\columnwidth]{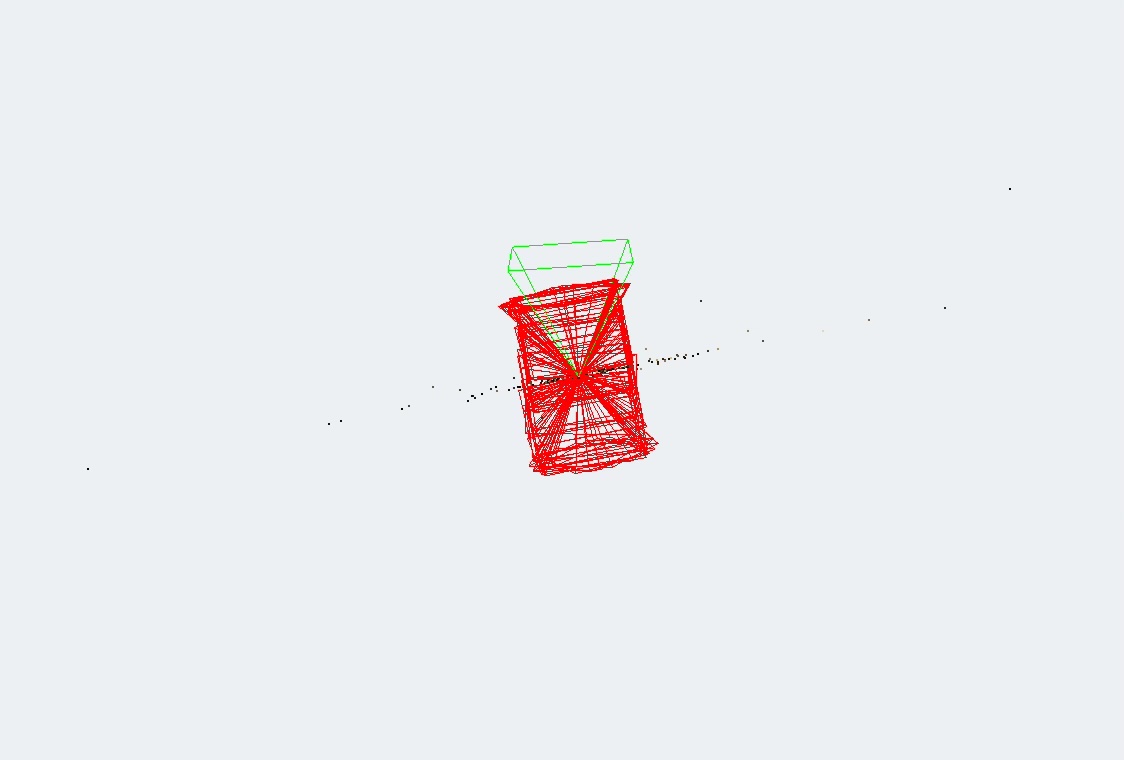} &
\includegraphics[width=0.485\columnwidth]{figures/fig_set2_cauchy_pose.jpg}
\\[-1mm]
\scriptsize (a) None &
\scriptsize (b) Cauchy
\end{tabular}
\caption{\textbf{Effect of robustification.}
On Set2 with pOSE and normal initialization, the non-robust solver collapses, while the Cauchy loss recovers a valid metric reconstruction.}
\label{fig:effect_cauchy}
\end{figure}

\begin{table}[t]
\centering
\caption{\textbf{Robustification can prevent catastrophic failures with linear metric upgrade.}
Each row compares the non-robust formulation with the Cauchy variant for the same dataset, formulation, and initialization.}
\label{tab:robust_rescue_cases_minobs3_linear1997}
\scriptsize
\setlength{\tabcolsep}{3pt}
\renewcommand{\arraystretch}{1.12}
\resizebox{\columnwidth}{!}{%
\begin{tabular}{@{}llllccc@{}}
\toprule
\textbf{Dataset} & \textbf{Form.} & \textbf{Init.} & \textbf{Loss} & \textbf{Rot. A20} & \textbf{Trans. A20} & \textbf{Lmk. RMSE} $\downarrow$ \\
\midrule
Set2 & pOSE & Normal & None   & 0.4 & 3.9 & 40.66 \\
     &      &        & Cauchy & 97.8 & 99.0 & 0.10 \\
\midrule
Set2 & pOSE+rot & Uniform & None   & 2.5 & 9.9 & 40.66 \\
     &          &         & Cauchy & 99.3 & 99.7 & 0.03 \\
\midrule
Set2 & rOSE & Normal & None   & 3.7 & 7.0 & 40.66 \\
     &      &        & Cauchy & 96.4 & 98.3 & 0.17 \\
\midrule
Set3 & RpOSE & Uniform & None   & 50.4 & 81.8 & 1.57 \\
     &       &         & Cauchy & 84.6 & 93.3 & 0.43 \\
\bottomrule
\end{tabular}}
\end{table}

\begin{figure}[t]
\centering
\setlength{\tabcolsep}{2pt}
\renewcommand{\arraystretch}{0.95}
\begin{tabular}{cc}
\includegraphics[width=0.485\columnwidth]{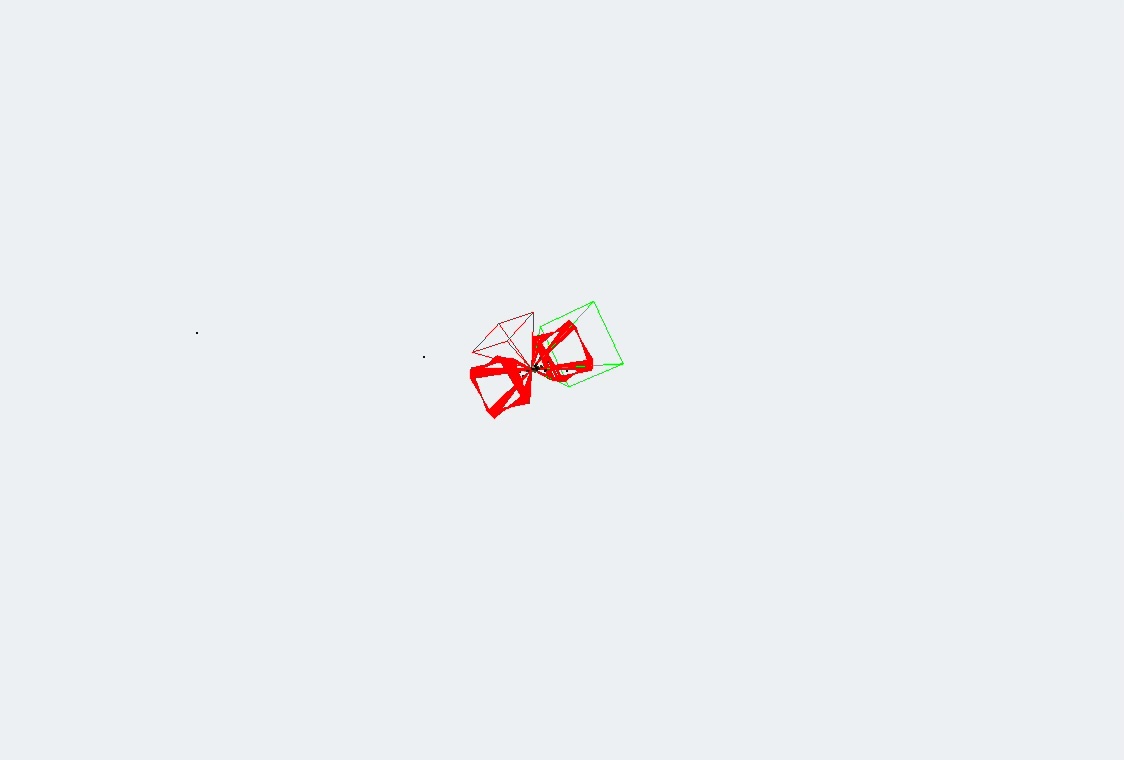} &
\includegraphics[width=0.485\columnwidth]{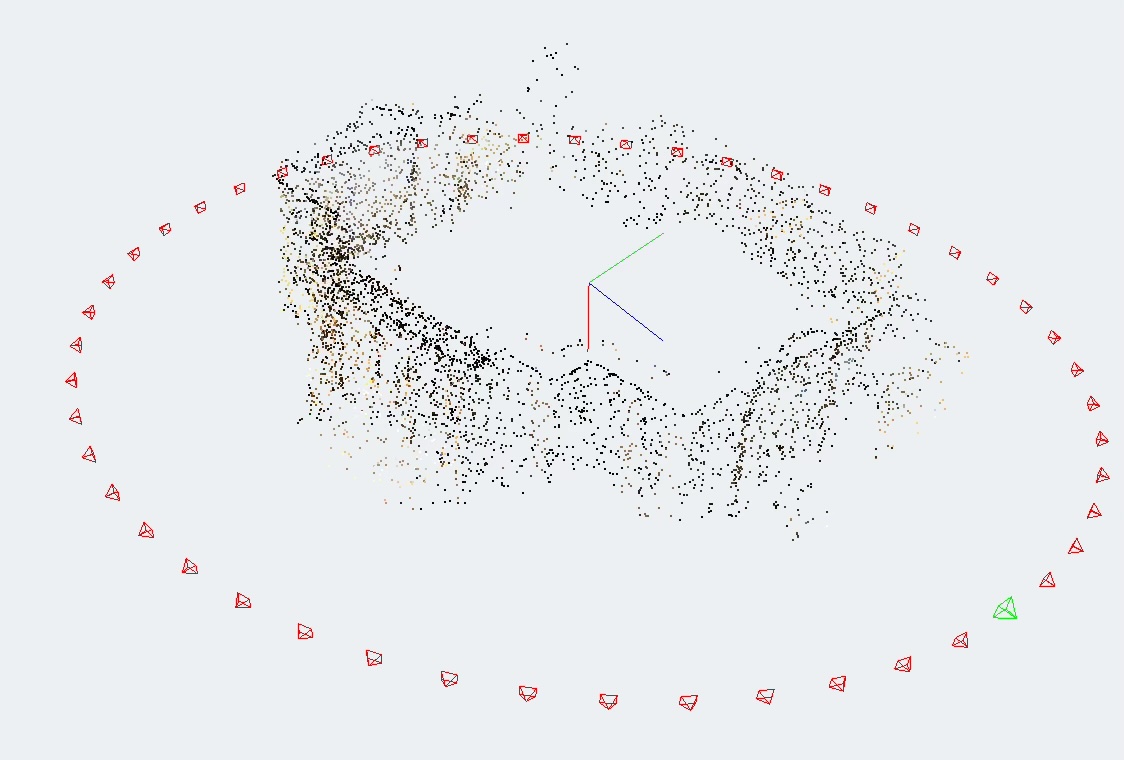}
\\[-1mm]
\scriptsize (a) Min. 3 obs. per landmark &
\scriptsize (b) Min. 6 obs. per landmark
\end{tabular}
\caption{\textbf{Effect of observation density.}
On Set2 with rOSE, filtering weakly observed landmarks stabilizes the metric upgrade and produces a more coherent reconstruction.}
\label{fig:effect_densification}
\end{figure}

\begin{table}[t]
\centering
\caption{\textbf{Effect of densification.}
We report the change obtained by increasing the landmark visibility threshold from three to six observations, computed on the six complete common datasets: Set1--Set5 and Set8. Positive values are better for AUC metrics; negative values are better for landmark RMSE. Results use linear metric upgrade.}
\label{tab:summary_densification_delta}
\scriptsize
\setlength{\tabcolsep}{3.5pt}
\renewcommand{\arraystretch}{1.12}
\resizebox{\columnwidth}{!}{%
\begin{tabular}{@{}lccccc@{}}
\toprule
\textbf{Loss}
& $\Delta$ \textbf{Rot. A5}
& $\Delta$ \textbf{Rot. A20}
& $\Delta$ \textbf{Trans. A5}
& $\Delta$ \textbf{Trans. A20}
& $\Delta$ \textbf{Lmk. RMSE} $\downarrow$ \\
\midrule
None   & +6.31 & +6.21  & +6.46 & +5.10 & -2.63 \\
Cauchy & +8.55 & +10.08 & +9.38 & +9.54 & -4.67 \\
\bottomrule
\end{tabular}}
\end{table}

\begin{table}[t]
\centering
\caption{\textbf{Model-wise effect of densification.}
We report the change induced by increasing the minimum landmark visibility from three to six observations, using the six complete common datasets. Results are averaged over initialization distributions and non-robust/Cauchy variants after linear metric upgrade.}
\label{tab:densification_by_formulation}
\scriptsize
\setlength{\tabcolsep}{4pt}
\renewcommand{\arraystretch}{1.12}
\resizebox{\columnwidth}{!}{%
\begin{tabular}{@{}lccc@{}}
\toprule
\textbf{Form.}
& $\Delta$ \textbf{Rot. A20}
& $\Delta$ \textbf{Trans. A20}
& $\Delta$ \textbf{Lmk. RMSE} $\downarrow$ \\
\midrule
pOSE     & +1.04  & +2.35  & -2.53 \\
rOSE     & +12.60 & +12.01 & -5.68 \\
RpOSE    & +18.34 & +14.17 & -5.15 \\
expOSE   & +1.99  & +3.61  & -1.37 \\
pOSE+rot & +6.75  & +4.46  & -3.53 \\
\bottomrule
\end{tabular}}
\end{table}

\begin{table}[t]
\centering
\caption{\textbf{Robustification can prevent catastrophic failures for minimum 6 obs. per landmark, with linear metric upgrade.}
Each row compares the non-robust formulation with the best robust variant for the same dataset, formulation, and initialization. Metrics are reported after the linear metric upgrade.}
\label{tab:robust_rescue_cases_minobs6_linear1997}
\scriptsize
\setlength{\tabcolsep}{3pt}
\renewcommand{\arraystretch}{1.12}
\resizebox{\columnwidth}{!}{%
\begin{tabular}{@{}llllccc@{}}
\toprule
\textbf{Dataset} & \textbf{Form.} & \textbf{Init.} & \textbf{Loss} & \textbf{Rot. A20} & \textbf{Trans. A20} & \textbf{Lmk. RMSE} $\downarrow$ \\
\midrule
Set4 & expOSE & Normal & None   & 0.6 & 0.3 & 41.52 \\
     &        &        & Cauchy & 100.0 & 100.0 & 0.00 \\
\midrule
Set2 & RpOSE & Normal & None   & 1.3 & 2.2 & 41.00 \\
     &       &        & Cauchy & 99.9 & 99.9 & 0.01 \\
\midrule
Set3 & pOSE & Normal & None   & 0.9 & 3.8 & 41.19 \\
     &      &        & Cauchy & 100.0 & 100.0 & 0.00 \\
\bottomrule
\end{tabular}}
\end{table}

\begin{table}[t]
\centering
\caption{\textbf{Effect of linear metric upgrade.}
We compare the projective solution before metric upgrade with the reconstruction obtained after the linear metric upgrade. Results are averaged over all datasets, formulations, initializations, and losses.}
\label{tab:metric_upgrade_effect}
\scriptsize
\setlength{\tabcolsep}{4pt}
\renewcommand{\arraystretch}{1.12}
\resizebox{\columnwidth}{!}{%
\begin{tabular}{@{}lccc@{}}
\toprule
\textbf{Stage}
& \textbf{Rot. A20}
& \textbf{Trans. A20}
& \textbf{Lmk. RMSE} $\downarrow$ \\
\midrule
Before metric upgrade & 16.34 & 6.23  & 25.86 \\
Linear metric upgrade & 36.30 & 39.06 & 22.14 \\
\midrule
Mean change & +19.96 & +32.83 & -3.73 \\
\bottomrule
\end{tabular}}
\end{table}

\begin{figure}[t]
\centering
\includegraphics[width=\columnwidth]{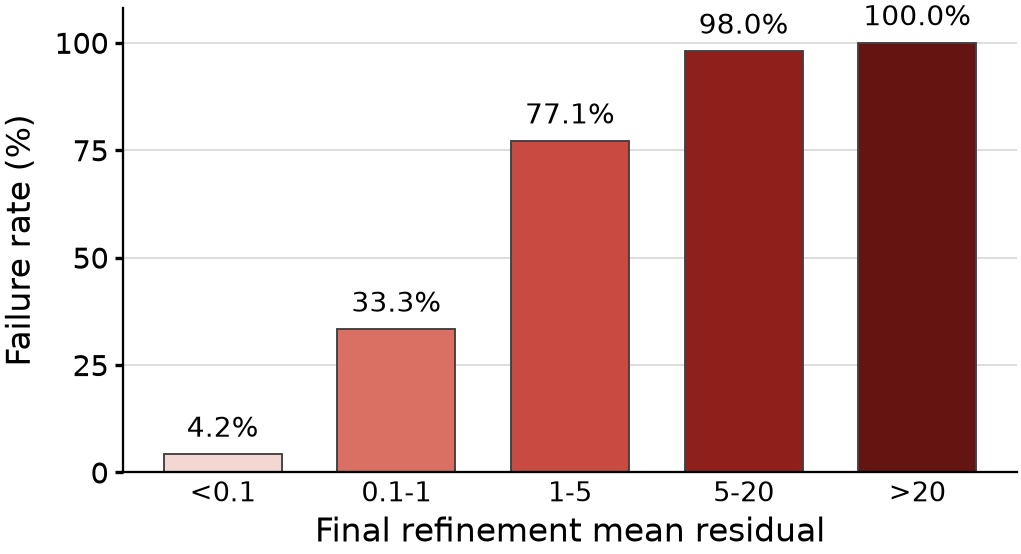}
\caption{\textbf{Metric upgrade failure rate as a function of the final refinement residual.}
We use the final mean residual after the refinement optimization and evaluate failure after linear metric upgrade with the same criterion as in \cref{fig:metric_upgrade_failure_rate_ose}.}
\label{fig:metric_upgrade_failure_rate_refinement}
\end{figure}

\subsection{Comparative Analysis}\label{subsec:comparative}

\subsubsection{Metric upgrade protocol}
Unless stated otherwise, the results in the main paper are reported after the linear metric upgrade. We use this protocol because it is simple, deterministic, and a strong baseline. Although we also enable the nonlinear refinement stage in our implementation, let us highlight that Pollefeys et al. \cite{pollefeys1999self} noticed that this refinement can be unstable. We also experiment with a geometric selection between the linear and nonlinear refined metric upgrades; these results are reported in the appendix. The selected variant gives similar aggregate trends, but the linear upgrade provides a cleaner reference point for comparing OSE formulations, initialization distributions, and robust VarPro.

\subsubsection{Comparison of OSE formulations}
Across the tested formulations, pOSE+rot gives the strongest overall performance (\Cref{tab:summary_formulations_pose_linear1997,tab:summary_formulations_landmarks_linear1997}). Averaged over all datasets, initialization distributions, and non-robust/Cauchy variants, it obtains the highest rotation AUC@20, 53.87, the highest translation-direction AUC@20, 57.08, and the lowest landmark RMSE, 16.63. This confirms that relative-rotation information provides a strong stabilizing constraint for the Euclidean reconstruction.

Among the formulations that do not use additional relative-rotation information, pOSE, rOSE, and expOSE are closer. expOSE gives the best average rotation AUC@20 among these methods, 35.89, while rOSE gives the best translation-direction AUC@20, 37.56. In terms of structure, expOSE and rOSE are again the strongest non-rotation formulations, with landmark RMSE values of 21.40 and 21.82, respectively. RpOSE is less competitive in these experiments, with lower pose AUCs and larger landmark errors on average. Overall, these results show that the different OSE objectives are not equivalent once evaluated after metric upgrade: their optimization behavior may be comparable, but their induced projective reconstructions can differ substantially in how well they support a Euclidean upgrade.

Dense, well-connected configurations (Sets 1, 3) are solved by most formulations regardless of initialization or robustness. On Set~1 (dense ring, $\sim$10 observations per landmark) and Set~3 (three rings, $\sim$6 observations per landmark), pOSE, rOSE, expOSE, and pOSE+rot all converge reliably under both normal and uniform initialization. RpOSE is the notable exception, remaining poor even on these favorable configurations (e.g., 13--50\% rotation AUC@20 on Set~3).

\subsubsection{Effect of initialization}
The initialization distribution (\Cref{tab:init_average_linear1997,fig:effect_initialization}) has a major impact on the final metric reconstruction, showing that initialization-free BA is not fully insensitive to the way projective cameras are randomized. Averaged over all formulations, unit-circle initialization gives the best results: rotation AUC@20 increases from 29.60 with normal initialization and 30.54 with uniform initialization to 42.60 with unit-circle initialization. The same trend appears for translation, where AUC@20 increases from 32.21 and 33.04 to 45.47, and for structure, where landmark RMSE decreases from 24.63 and 24.27 to 19.27.
This suggests that InitFree BA is better described as free from scene-specific initialization rather than initialization-independent. The solver does not require an external SfM initialization, but the sampling distribution used to initialize the projective cameras acts as an implicit prior. A favorable distribution can produce projective configurations that are much easier to optimize and upgrade to a metric frame.
Importantly, let us highlight that a geometric prior can help even when it does not match the true trajectory. Indeed, on Set~8 (lateral camera translation with parallel optical axes), unit-circle initialization is very effective (96--99\% for pOSE, rOSE, RpOSE) despite the true trajectory looking nothing like a circle, suggesting the benefit is not solely explained by matching scene geometry but also by avoiding degenerate or poorly separated starting configurations. 

Finally, sparse configurations (Sets 2, 4, 6) separate the formulations and initializations sharply. Set~2 shares Set~1's camera ring but reduces observation density to $\sim$3 per landmark, and normal/uniform initialization collapses almost completely for pOSE, rOSE, and expOSE (rotation AUC@20 $\leq$4\%). Unit-circle initialization substantially rescues these formulations (44--81\%), while pOSE+rot remains strong even under normal initialization (97.4\%) thanks to its relative-rotation constraints. Set~4 repeats this sparsity reduction on the three-ring geometry of Set~3, but here unit-circle initialization does not rescue pOSE, rOSE, or expOSE (all remain below 10\%), while pOSE+rot again stays near-perfect under all three initializations. Finally, Set~6 resists every combination of initialization (all formulations $\leq$11\%). These three datasets suggest that sparse configuration remains an important challenge for InitFree BA.

\subsubsection{Robust VarPro}
We also evaluate Cauchy robustification as an additional stabilization mechanism. Unlike in classical bundle adjustment, where robust losses are usually introduced to handle mismatched correspondences, our motivation is to reduce the influence of unstable projected residuals produced during the early iterations of VarPro. In practice, robustification is applied after eliminating the linear landmark variables (\cref{eq:projected_residual}): the projected residuals and their Jacobians are reweighted inside the reduced camera optimization, which preserves the VarPro elimination step. It is based on first-order Triggs correction (\cite{triggs1999bundle,zach2014robust}). The detailed reweighting scheme is given in the appendix.

The average effect of first-order Cauchy robustification is limited, with results close to non-robust VarPro (\Cref{tab:init_cauchy_pose_linear1997_pairs}). However, in several difficult configurations, the non-robust solver collapses while the robust variant recovers a valid metric reconstruction (\Cref{tab:robust_rescue_cases_minobs3_linear1997,tab:init_cauchy_landmarks_linear1997_pairs,fig:effect_cauchy}). For example, on Set2 with pOSE and normal initialization, the non-robust solver obtains only 0.4 rotation AUC@20 and 3.9 translation AUC@20, with landmark RMSE 40.66. The Cauchy variant reaches 97.8 rotation AUC@20, 99.0 translation AUC@20, and landmark RMSE 0.10. Similar rescue cases occur for rOSE and pOSE+rot on Set2. Thus, robustification is not the dominant factor in average performance, but it is valuable as a safeguard against some catastrophic projective configurations.

\subsubsection{Effect of landmark densification}
As we have seen, in contrast to traditional BA, the sparsity is particularly challenging for InitFree BA. Following this insight, we now show that increasing the landmark visibility threshold from three to six observations improves the stability of the metric upgrade (\Cref{tab:summary_densification_delta,fig:effect_densification}). On the common subset of six datasets, the average rotation AUC@20 increases by 6.21 points without robustification and by 10.08 points with Cauchy robustification. Translation-direction AUC@20 also improves, by 5.10 and 9.54 points respectively. The improvement is visible in structure as well: landmark RMSE decreases by 2.63 without robustification and by 4.67 with Cauchy.
The effect is especially strong for formulations that are sensitive to poorly constrained landmarks, and notably RpOSE (\Cref{tab:densification_by_formulation}). This behavior is expected since RpOSE achieves radial-distortion invariance by discarding the tangential component of each image observation. Each correspondence therefore contributes fewer independent constraints, making the optimization more sensitive to the total amount of available data. Increasing the observation density compensates for this reduction in information and substantially improves reconstruction quality. Averaged over robust and non-robust variants, rOSE improves by 12.60 rotation AUC@20 points and 12.01 translation AUC@20 points, while RpOSE improves by 18.34 and 14.17 points respectively. pOSE+rot also benefits from densification, improving by 6.75 rotation AUC@20 points and 4.46 translation AUC@20 points. These results suggest that weakly observed landmarks reduce the stability of the projective reconstruction, making the subsequent metric upgrade considerably more fragile. Similarly, a robust VarPro can prevent some failure cases even when coupled with a densification (\Cref{tab:robust_rescue_cases_minobs6_linear1997}).

\subsubsection{Metric upgrade}
\Cref{tab:metric_upgrade_effect} highlights the necessity of the metric upgrade stage, whereas \Cref{fig:metric_upgrade_failure_rate_ose,fig:metric_upgrade_failure_rate_refinement} show the percentage of failure cases with respect to the OSE error (stage 1) and to the final refinement error (stage 2), respectively. It is noteworthy that even with very small OSE residuals the failure rate is relatively high, showing the discrepancy between optimization and 3D reconstruction. In particular, regarding the plots for the final refinement residual, the gap appears between stage 1 and stage 2: a low OSE error does not translate into a low final residual. Conversely, a small final residual is a driver of the success of the metric upgrade. Consequently, the link between OSE optimization and reprojection error is not as straightforward as reported by previous works, as a small OSE error can lead to a large final refinement residual. As shown in the supplemental, this result is consistent when we analyze the selected metric upgrade, suggesting that the observed failures are primarily driven by the preceding stages of the InitFree pipeline rather than by the particular
metric-upgrade variant considered here.

\subsection{Discussion}

Our experiments reveal four main lessons. First, relative-rotation constraints provide the strongest stabilization; among unconstrained formulations, expOSE and rOSE perform best, while RpOSE is weakest. Second, unit-circle initialization outperforms normal and uniform initialization on average, showing that current methods remain dependent on a generic geometric prior despite requiring no scene-specific SfM initialization. This advantage persists even when the ground-truth camera distribution differs substantially from a unit circle, as in Set~8. Third, robust VarPro is not a general improvement, but rather a targeted safeguard against catastrophic failures. Fourth, filtering weakly observed landmarks substantially stabilizes the subsequent metric upgrade, despite the upgrade depending only on the recovered cameras. This suggests that small perturbations induced by poorly constrained landmarks can sufficiently affect the projective cameras to destabilize the upgrade. Finally, reconstruction success is strongly linked to the final refinement residual rather than the OSE residual, highlighting the discrepancy between the OSE approximation and the reprojection error (\Cref{eq:approximation_OSE}).

\begin{figure*}[t]
\centering
\setlength{\tabcolsep}{1.5pt}
\renewcommand{\arraystretch}{0.95}
\begin{tabular}{cccc}

\includegraphics[width=0.24\textwidth,height=0.118\textheight,keepaspectratio]{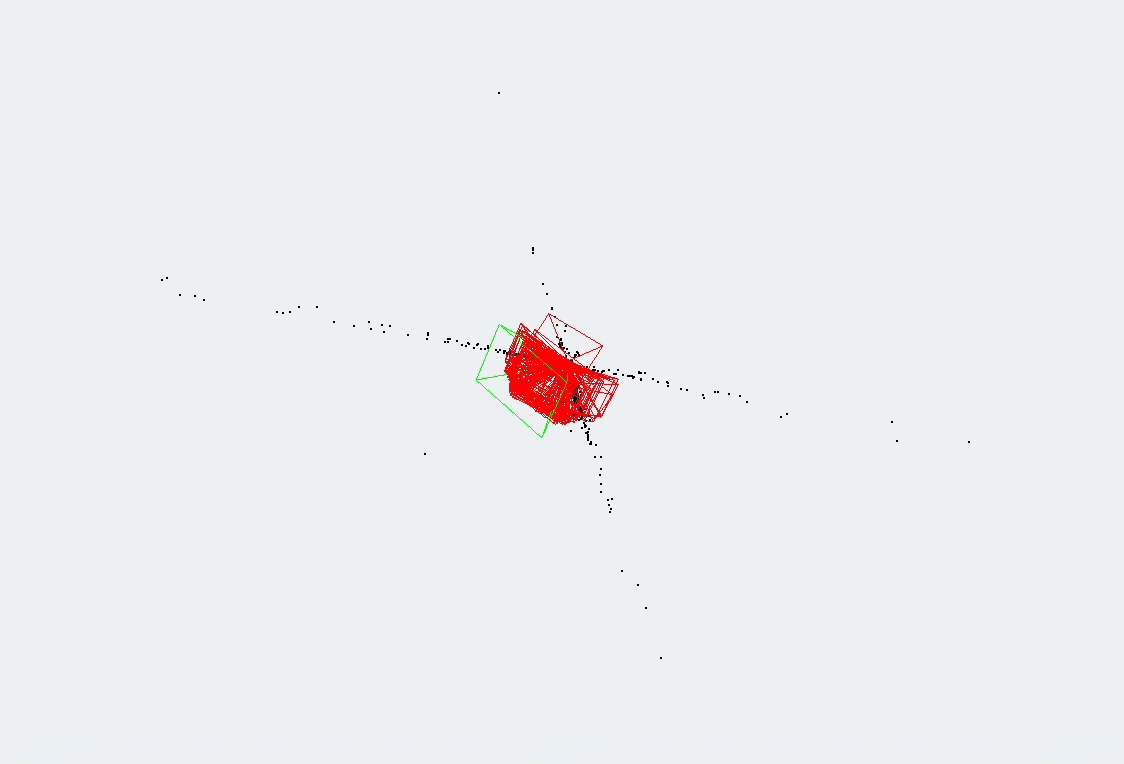} &
\includegraphics[width=0.24\textwidth,height=0.118\textheight,keepaspectratio]{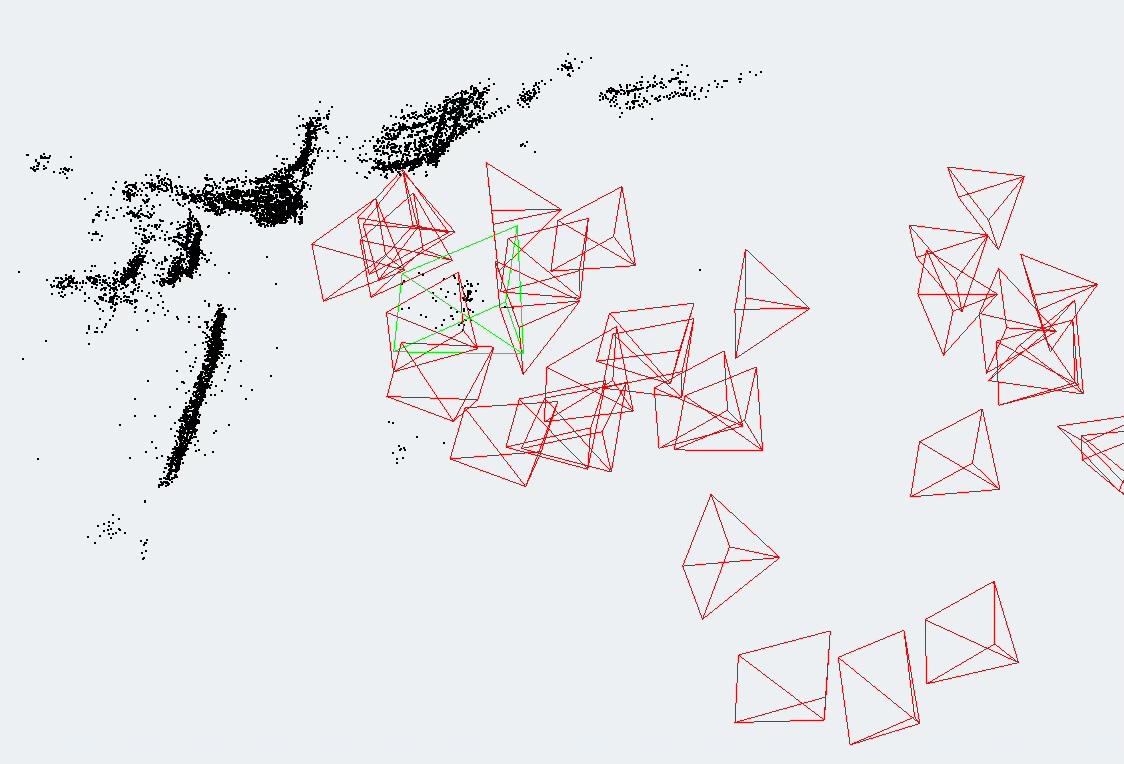} &
\includegraphics[width=0.24\textwidth,height=0.118\textheight,keepaspectratio]{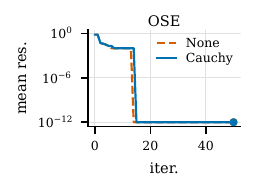} &
\includegraphics[width=0.24\textwidth,height=0.118\textheight,keepaspectratio]{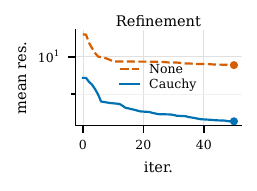}
\\[-1mm]
{\scriptsize (a) pOSE+rot, no robust loss} &
{\scriptsize (b) pOSE+rot, Cauchy loss} &
{\scriptsize (c) OSE residual} &
{\scriptsize (d) refinement residual}
\\[2mm]

\includegraphics[width=0.24\textwidth,height=0.118\textheight,keepaspectratio]{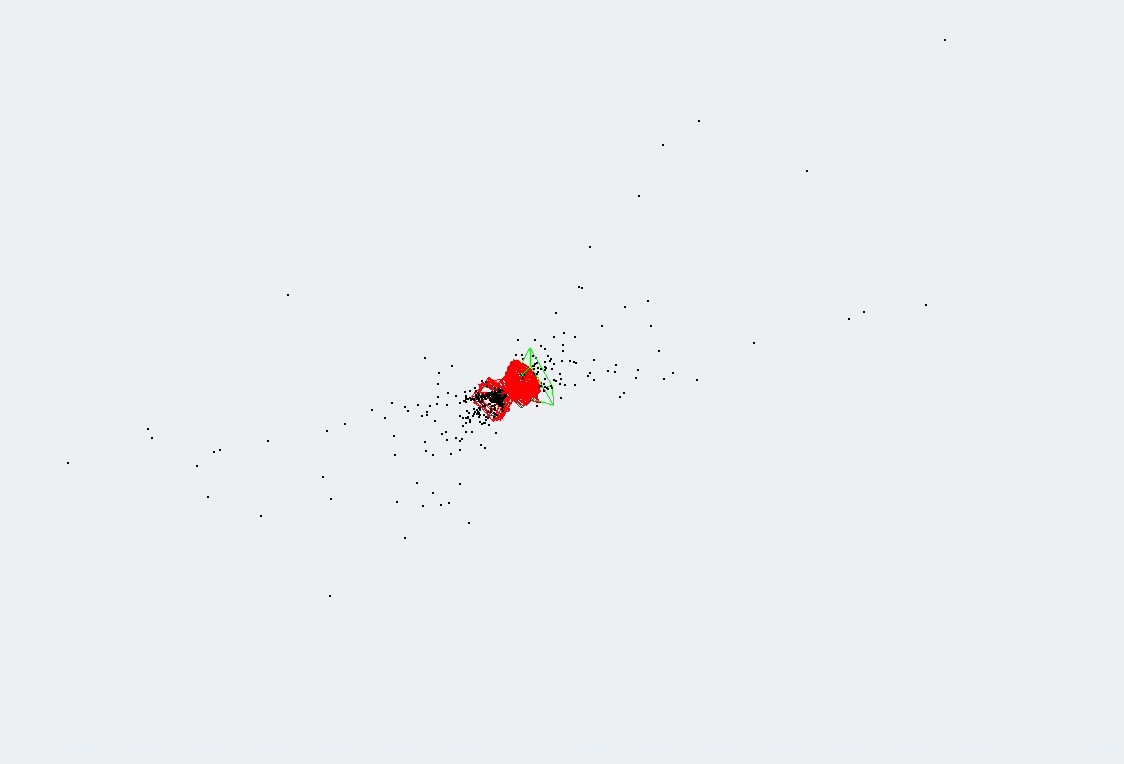} &
\includegraphics[width=0.24\textwidth,height=0.118\textheight,keepaspectratio]{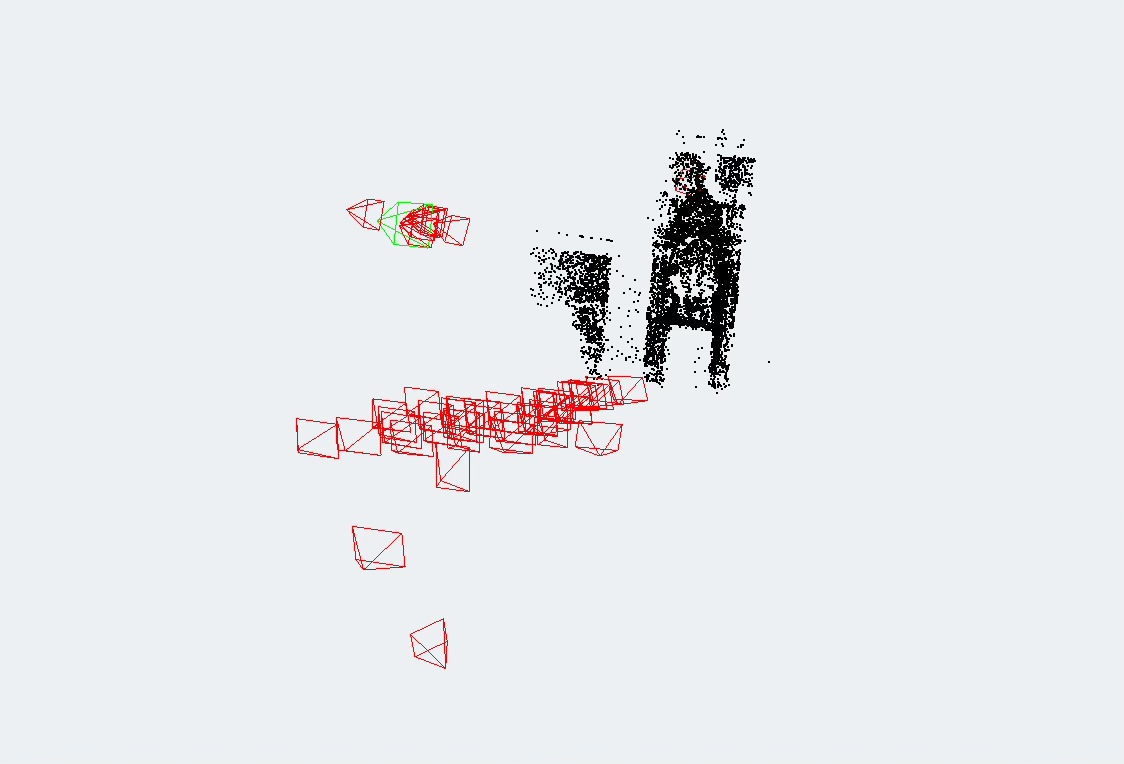} &
\includegraphics[width=0.24\textwidth,height=0.118\textheight,keepaspectratio]{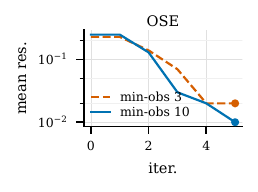} &
\includegraphics[width=0.24\textwidth,height=0.118\textheight,keepaspectratio]{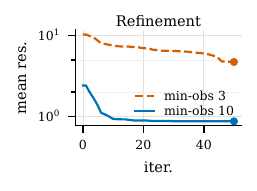}
\\[-1mm]
{\scriptsize (e) expOSE, at least 3 obs.} &
{\scriptsize (f) expOSE, at least 10 obs.} &
{\scriptsize (g) OSE residual} &
{\scriptsize (h) refinement residual}
\\[2mm]

\includegraphics[width=0.24\textwidth,height=0.118\textheight,keepaspectratio]{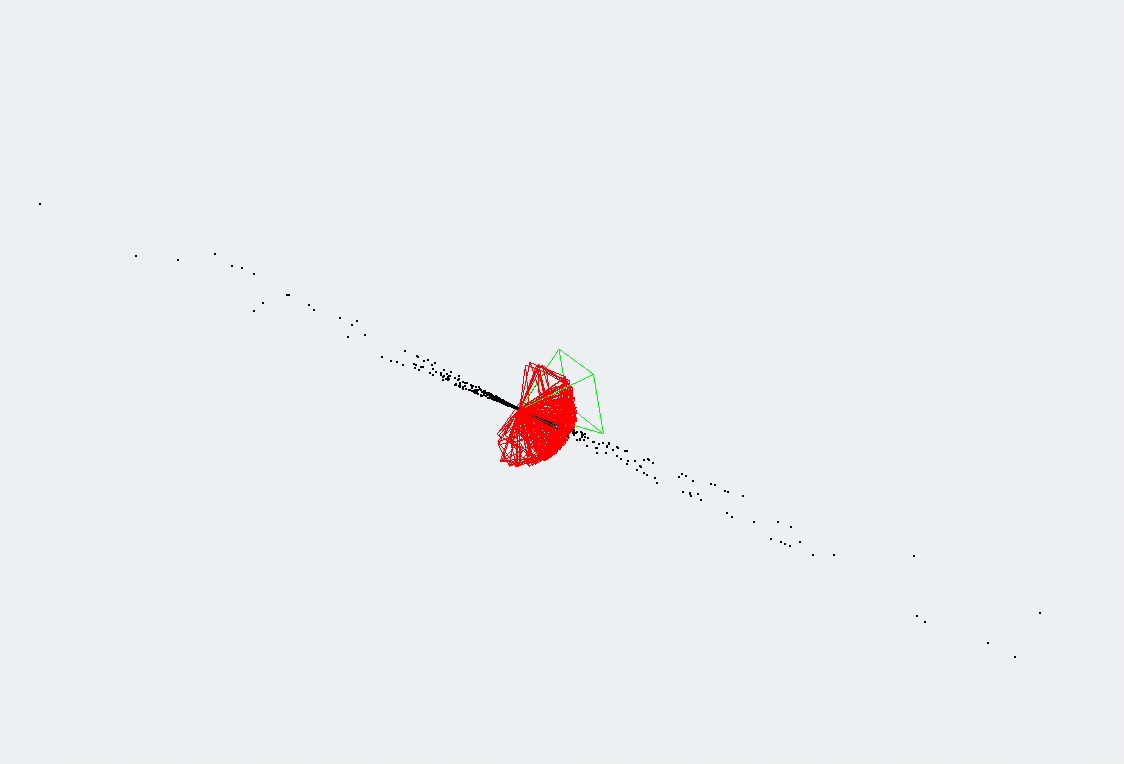} &
\includegraphics[width=0.24\textwidth,height=0.118\textheight,keepaspectratio]{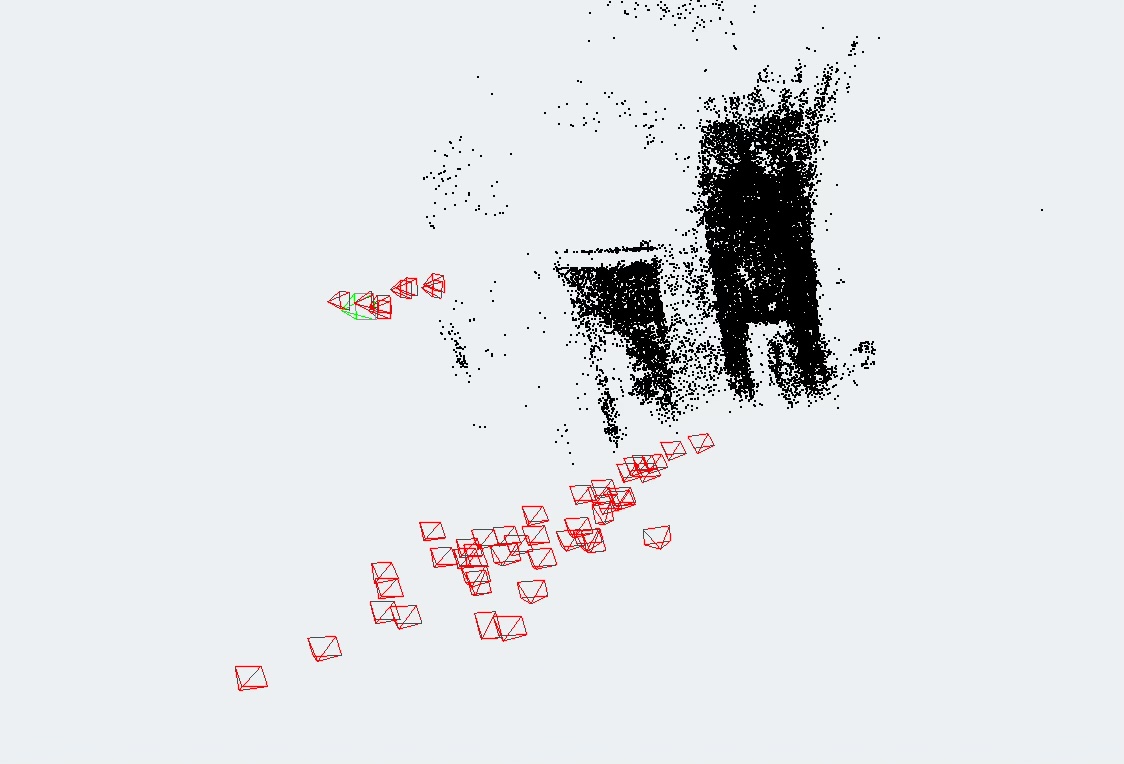} &
\includegraphics[width=0.24\textwidth,height=0.118\textheight,keepaspectratio]{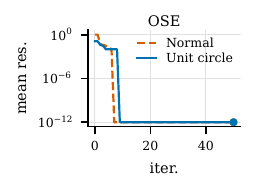} &
\includegraphics[width=0.24\textwidth,height=0.118\textheight,keepaspectratio]{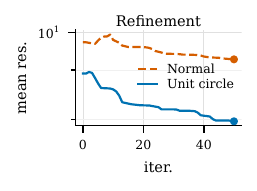}
\\[-1mm]
{\scriptsize (i) pOSE+rot, normal init.} &
{\scriptsize (j) pOSE+rot, unit-circle init.} &
{\scriptsize (k) OSE residual} &
{\scriptsize (l) refinement residual}

\end{tabular}
\caption{\textbf{Failure and recovery modes on BAL after linear metric upgrade.}
Each row varies one factor while keeping the scene and formulation fixed, for Trafalgar-150 (first row), and Venice-52 (other rows). The OSE residual decreases in both successful and failed cases, whereas the refinement residual more clearly separates successful metric reconstructions from collapsed ones.}
\label{fig:bal_real_failure_recovery}
\end{figure*}

\section{Beyond Controlled Benchmarks}
\label{sec:towards_bal}

Our controlled benchmark isolates the effects of initialization, observation density, robustification, and metric upgrade, but does not establish whether these findings transfer to established SfM data. We therefore complement it with qualitative experiments on BAL problems~\cite{agarwal2010bundle}, which provide heterogeneous camera configurations, observation graphs, and noise patterns. Unlike prior work using BAL primarily to evaluate OSE optimization and scalability~\cite{weber2024power}, we examine the resulting reconstructions after metric upgrade. As shown in \cref{fig:bal_real_failure_recovery}, the main failure modes identified in our controlled experiments persist on BAL. Reconstructions can collapse after metric upgrade despite successful OSE reduction, while changing observation density, initialization, or robustification can recover coherent solutions. Moreover, the refinement residual, rather than the OSE residual, clearly separates successful from failed reconstructions. This confirms that the gap between OSE optimization and metric reconstruction is not specific to our controlled benchmark, and highlights the discrepancy between the OSE surrogate and the reprojection error as a key indicator of reconstruction success.

\section{Conclusion}
\label{sec:conclusion}

In this paper, we revisited initialization-free bundle adjustment from the perspective of \emph{metric reconstruction} rather than objective minimization alone. Using a unified implementation and controlled evaluation framework, we show that optimization success does not necessarily imply successful Euclidean reconstruction: projective solutions with similarly low OSE values can lead to substantially different metric upgrades. Our experiments identify initialization priors, observation density, robustification, and metric upgrade stability as key factors governing this optimization--reconstruction gap. Among the evaluated methods, pOSE+rot achieves the strongest overall performance by exploiting relative-rotation information, while expOSE is the strongest purely OSE-based formulation. However, no method consistently succeeds across all configurations, and experiments on BAL indicate that the observed failure modes extend beyond our controlled benchmark. Overall, our results suggest that progress in InitFree BA requires considering the complete pipeline, from optimization and projective reconstruction to metric upgrade, rather than objective minimization alone. We hope that our unified implementation and benchmark will provide a solid foundation for future work in this direction. Finally, our results show that InitFree BA is a substantially more challenging problem than suggested by optimization success alone, and that reliable metric reconstruction remains far from solved.

\section*{Acknowledgments}
This work was supported by the European Research Council (ERC) Advanced Grant
SIMULACRON, by the DFG project CR 250/26-1 “4D-YouTube”, by the GNI Project
“AI4Twinning”, and by the Munich Center for Machine Learning.

\vspace{11pt}

{
    \small
    \bibliographystyle{IEEEtran}
    \bibliography{main}
}

\section{Biography Section}

\begin{IEEEbiography}[{\includegraphics[width=1in,height=1.25in,clip,keepaspectratio]{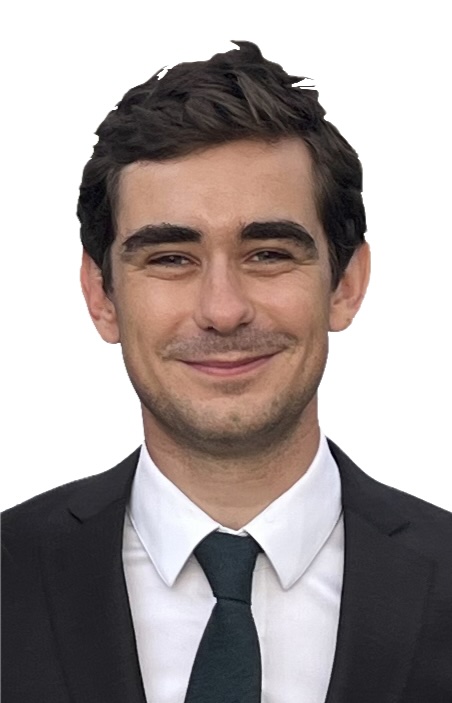}}]{Simon Weber} received his PhD in computer science from the Technical University of Munich, Germany, in 2025. He is currently a research associate in the department of computer science at the University of Oxford, United Kingdom, within the PIXL lab. He is also a member of the European Laboratory for Learning and Intelligent Systems (ELLIS). His research interests include 3D reconstruction, beyond-Euclidean computer vision, and optimization.

\end{IEEEbiography}

\begin{IEEEbiography}[{\includegraphics[width=1in,height=1.25in,clip,keepaspectratio]{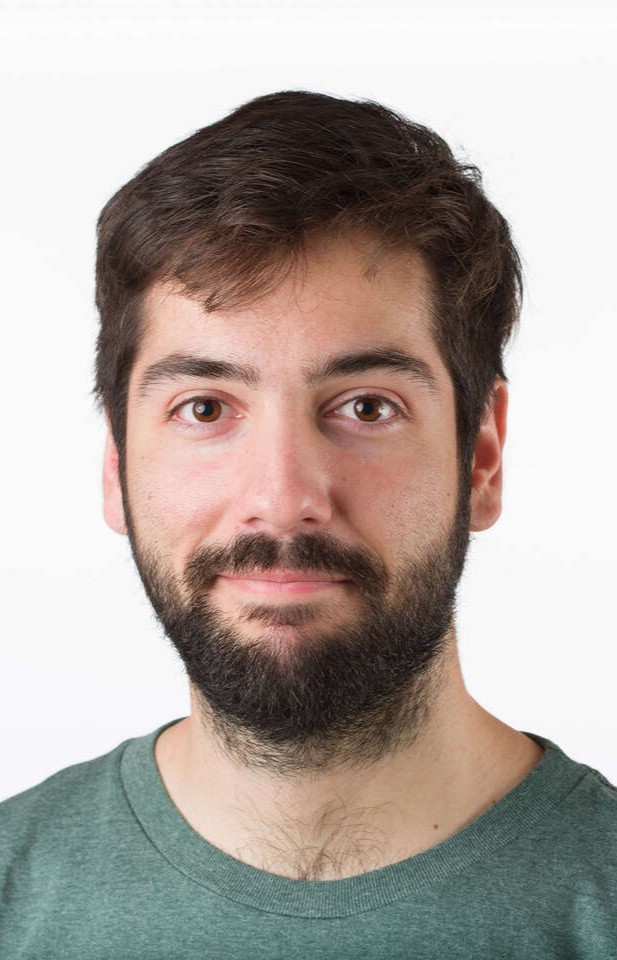}}]{Mateo de Mayo}
received his Licentiate degree in computer science from the National University of Córdoba, Argentina. He is currently pursuing a PhD at the Technical University of Munich, Germany, as a member of the Munich Center for Machine Learning and the Computer Vision Group under the supervision of Prof. Daniel Cremers. Before his PhD, he worked on open-source visual-inertial tracking and mixed reality systems. His research interests include embedded vision, low-latency tracking, and real-time 3D reconstruction.

\end{IEEEbiography}

\begin{IEEEbiography}[{\includegraphics[width=1in,height=1.25in,clip,keepaspectratio]{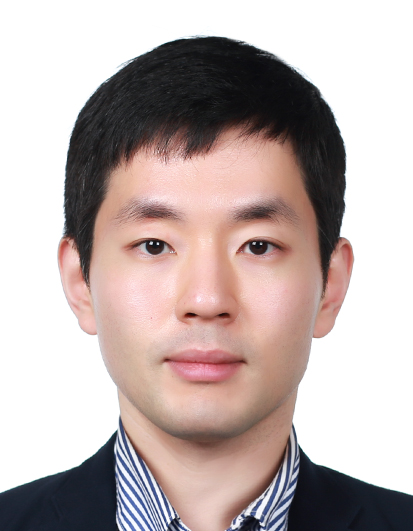}}]{Je Hyeong Hong}
received B.A. and M.Eng. degrees in Engineering (Electrical and Information Sciences) from the University of Cambridge, UK in 2011, and subsequently received a Ph.D. degree in Engineering (Computer Vision) from the University of Cambridge, UK, in 2018.
He served alternative military service in South Korea as a postdoctoral researcher at the Korea Institute of Science and Technology (KIST).
Currently, Je Hyeong is an Assistant Professor in the Department of Electronic Engineering at Hanyang University, Seoul, Korea.
His main research interests include computer vision, machine learning and optimization.
\end{IEEEbiography}

\begin{IEEEbiography}[{\includegraphics[width=1in,height=1.25in,clip,keepaspectratio]{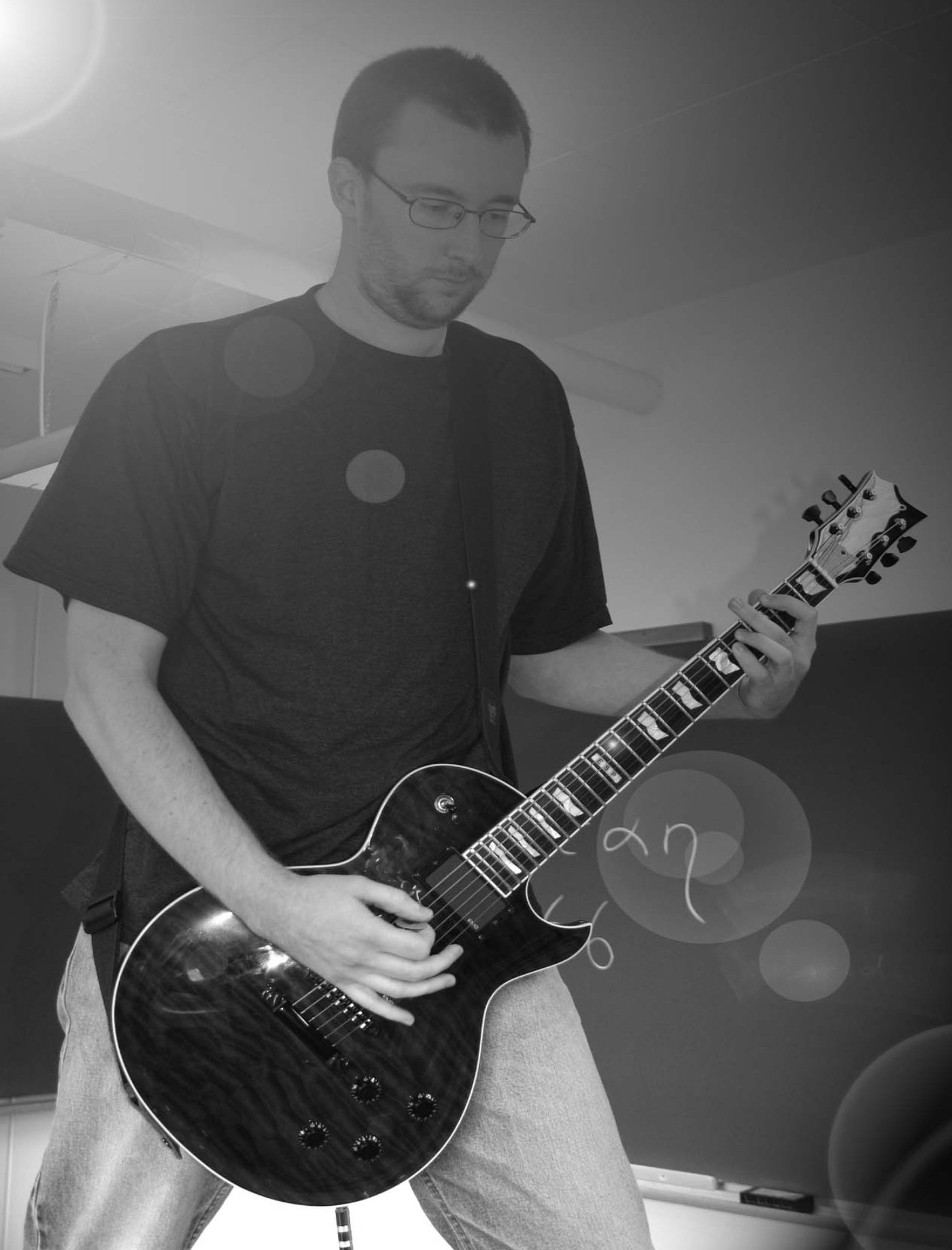}}]{Carl Olsson}
received his MSc degree in electrical engineering from the University of
Lund in 2004. He obtained his PhD degree in mathematics in 2009 at Lund University. Since then he has held positions at Lund University and Chalmers University of Technology.
He is currently a Professor at the Centre for
Mathematical Sciences, Lund University. His main research interests are optimization with
applications in computer vision, robotics, and related areas. In his spare time, he enjoys
working in the forest, playing the guitar and listening to rock’n roll.
\end{IEEEbiography}

\begin{IEEEbiography}[{\includegraphics[width=1in,height=1.25in,clip,keepaspectratio]{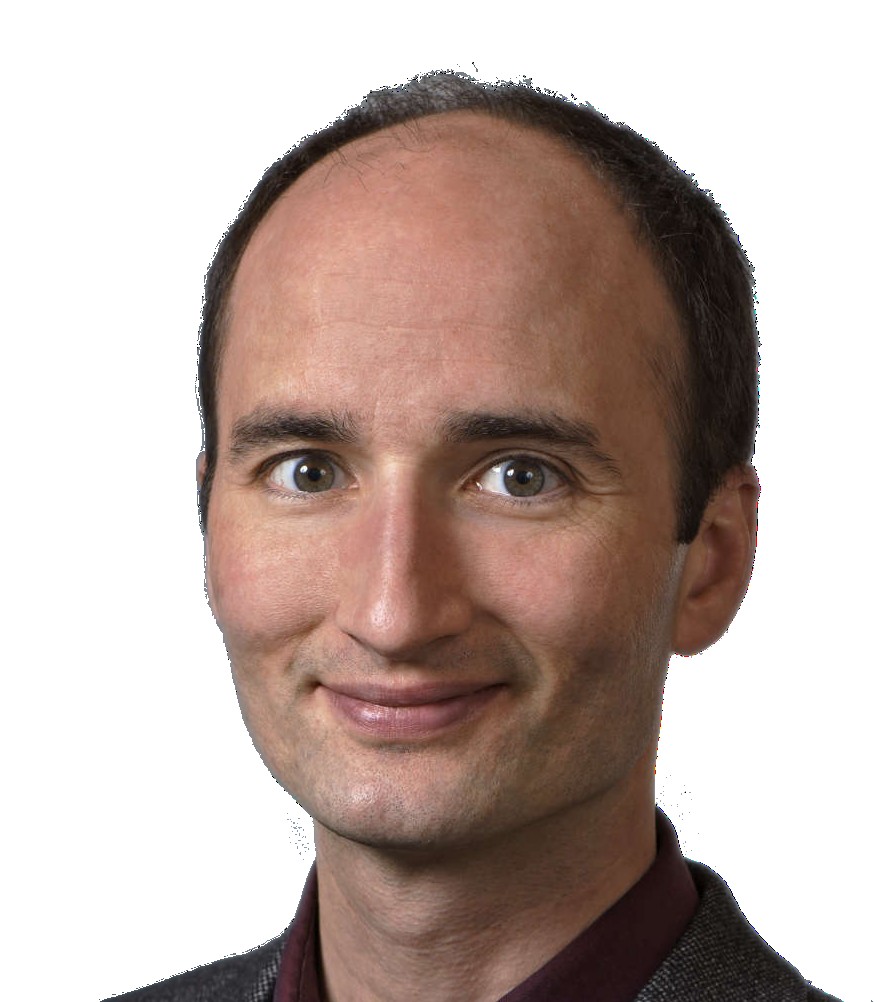}}]{Daniel Cremers}
received his PhD in computer science from the University of Mannheim, Germany, in 2002. He subsequently spent two years as a postdoctoral researcher at UCLA and one year at Siemens Corporate Research, before holding an associate professorship at the University of Bonn from 2005 to 2009. Since 2009, he holds the Chair of Computer Vision and Artificial Intelligence at the Technical University of Munich. He is a recipient of the Leibniz Award, the biggest award in German academia, as well as a Starting, Consolidator, and Advanced Grant from the European Research Council. Since 2023, he serves as President of the European Computer Vision Association. His research interests include 3D reconstruction, optimization, and machine learning for computer vision.

\end{IEEEbiography}

\begin{IEEEbiography}[{\includegraphics[width=1in,height=1.25in,clip,keepaspectratio]{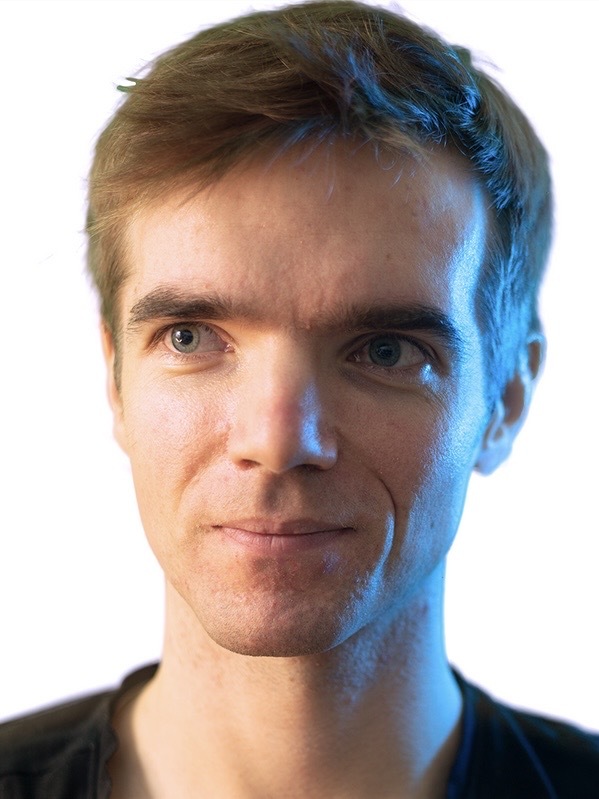}}]{Ronald Clark}
received his BSc and MSc from the University of the Witwatersrand in 2014 and his DPhil from the University of Oxford in 2017. He then joined the Dyson Robotics Lab and in 2019 was awarded an early-career Imperial College Research Fellowship (ICRF). In 2022 he joined the University of Oxford as an Associate Professor. His research interests include 3D computer vision at the intersection of geometry, optimization and generative models. His research received a best paper honourable mention at CVPR 2018.
\end{IEEEbiography}

\clearpage
\appendices

\section*{Supplementary Material}
\addcontentsline{toc}{section}{Supplementary Material}

This supplemental material is organized as follows:\\
\noindent \Cref{app:ose_residuals} surveys the explicit formulations of the OSE models discussed in \Cref{sec:problem}, and details the robust strategy for VarPro that we propose in \Cref{subsec:comparative}. \\
\noindent \Cref{app:metrics} details the metrics used in our experiments (\Cref{subsec:metrics}). \\
\noindent \Cref{app:selected_metric_ugprade} gives the results of our experiments by considering the selected metric upgrade strategy, instead of the linear metric upgrade. \\
\noindent \Cref{app:full_results} gives the details of the results for each dataset and each formulation. The tables given in the main paper are based on these tables. \\ 

\section{OSE Residual Definitions}
\label{app:ose_residuals}

Let $P_i \in \mathbb{R}^{3 \times 4}$ denote the projective camera associated with image $i$, let $\tilde{X}_j \in \mathbb{P}^3$ be the homogeneous coordinates of landmark $j$, and let $m_{ij} \in \mathbb{R}^2$ be the corresponding image observation. We write $P_{i,1:2}$ for the first two rows of $P_i$ and $p_{i,3}^{\top}$ for its third row.

\subsection{pOSE}

The original pOSE formulation combines an object-space residual with an affine regularization term:
\begin{equation}
l_\text{pOSE}
=
\sum_{i,j}
\left\|
\begin{bmatrix}
\sqrt{1-\alpha}\, r_{ij}^{\text{ose}}\\
\sqrt{\alpha}\, r_{ij}^{\text{aff}}
\end{bmatrix}
\right\|_2^2 ,
\end{equation}
where
\begin{equation}
r_{ij}^{\text{ose}}
=
P_{i,1:2}\tilde{X}_{j}
-
(p_{i,3}^{\top}\tilde{X}_{j})m_{ij},
\end{equation}
and
\begin{equation}
r_{ij}^{\text{aff}}
=
P_{i,1:2}\tilde{X}_{j}
-
m_{ij}.
\end{equation}
The parameter $\alpha$ controls the strength of the affine regularization.

\subsection{rOSE}

The rOSE formulation replaces the affine regularization used in pOSE by a depth-normalization term:
\begin{equation}
l_\text{rOSE}
=
\sum_{i,j}
\left\|
\begin{bmatrix}
\sqrt{1-\alpha}\, r_{ij}^{\text{ose}}\\
\sqrt{\alpha}\, r_{ij}^{\text{depth}}
\end{bmatrix}
\right\|_2^2 ,
\end{equation}
with
\begin{equation}
r_{ij}^{\text{depth}}
=
p_{i,3}^{\top}\tilde{X}_{j} - 1 .
\end{equation}
This term encourages landmarks to remain at a fixed normalized depth.

\subsection{RpOSE}

RpOSE modifies the OSE residual by introducing a radial constraint. In the formulation used in our implementation, the objective is
\begin{equation}
l_\text{RpOSE}
=
\sum_{i,j}
\left\|
\begin{bmatrix}
\sqrt{1-\alpha}\, r_{ij}^{\text{rad}}\\
\sqrt{\alpha}\, r_{ij}^{\text{aff}}
\end{bmatrix}
\right\|_2^2 ,
\end{equation}
where
\begin{equation}
r_{ij}^{\text{rad}}
=
\dfrac{\overline{m}_{ij}^{\top}P_{i,1:2}\tilde{X}_{j}}
       {\lVert m_{ij} \rVert}.
\end{equation}
Here $\overline{m}_{ij}$ denotes the direction orthogonal to the observed image point $m_{ij}$. This residual constrains the projected point along the radial direction while preserving the separable VarPro structure.

\subsection{expOSE}

expOSE replaces the original pOSE residual by an exponential residual that better approximates the perspective reprojection geometry:
\begin{equation}
l_\text{expOSE}
=
\sum_{i,j}
\left\|
\begin{bmatrix}
\sqrt{1-\alpha}\, r_{ij}^{\exp}\\
\sqrt{\alpha}\, r_{ij}^{\text{aff}}
\end{bmatrix}
\right\|_2^2 .
\end{equation}
Let
\begin{equation}
x_{ij} = P_{i,1:2}\tilde{X}_j,
\qquad
z_{ij} = p_{i,3}^{\top}\tilde{X}_j ,
\end{equation}
and let $(\overline{x}_{ij},\overline{z}_{ij})$ denote the linearization point used by expOSE. The exponential residual is
\begin{equation}
\begin{split}
r_{ij}^{\exp}
= {}& \frac{1}{\sqrt{2}}
\exp\!\left(
-\frac{m_{ij}^{\top}\overline{x}_{ij}+\overline{z}_{ij}}
       {2\sqrt{\lVert m_{ij}\rVert^{2}+1}}
\right)\\
&\cdot
\left(
\frac{m_{ij}^{\top}(x_{ij}-\overline{x}_{ij})
      +z_{ij}-\overline{z}_{ij}}
     {\sqrt{\lVert m_{ij}\rVert^2+1}}
-1
\right).
\end{split}
\end{equation}
This residual is designed to reduce the depth bias of the original pOSE formulation while keeping landmarks linear for fixed cameras.

\subsection{pOSE+rot}

The pOSE+rot formulation augments an OSE-style objective with pairwise relative rotation constraints:
\begin{equation}
\begin{split}
l_\text{pOSE+rot}
=
&\sum_{i,j}
\left\|
\begin{bmatrix}
\sqrt{1-\alpha}\, r_{ij}^{\text{ose}}\\
\sqrt{\alpha}\, r_{ij}^{\text{rot}}
\end{bmatrix}
\right\|_2^2 \\
&+
\beta
\sum_{(k,l)}
\left\|
\sqrt{W_{kl}}\,
\left(R_k R_l^{\top} - R_{kl}\right)
\right\|_F^2 ,
\end{split}
\end{equation}
where $R_k$ and $R_l$ are the estimated camera rotations, $R_{kl}$ is a pairwise relative rotation estimate, and $W_{kl}$ is a confidence weight. The residual used for the projection term is
\begin{equation}
r_{ij}^{\text{rot}}
=
\frac{
m_{ij}^{\top}P_{i,1:2}\tilde{X}_{j}
+
p_{i,3}^{\top}\tilde{X}_{j}}
{\lVert m_{ij} \rVert^2 + 1}
- 1 .
\end{equation}
In our experiments, we evaluate a variant in which this projection residual is replaced by the affine regularization used in pOSE, as it leads to significantly better results and better isolates the effect of the rotation constraints. The relative rotations are either taken from the ground-truth camera poses or estimated from image correspondences, depending on the experimental setting.

\subsection{Observation Normalization}

In all experiments, image observations are normalized before the OSE optimization so that coordinates lie in a comparable range, approximately $[-1,1]$. This preprocessing improves numerical conditioning and makes the scale of the residuals comparable across datasets and image resolutions.

For uncalibrated formulations, this normalization is an image-domain rescaling of the observations. The optimized cameras remain general projective cameras in $\mathbb{R}^{3 \times 4}$. For calibrated formulations, such as pOSE+rot, observations are additionally expressed in calibrated camera coordinates using the intrinsic calibration. The relative rotation constraints in pOSE+rot are therefore defined in the calibrated camera frame.

When a second-stage reprojection refinement is performed in pixel coordinates, the observations and cameras are converted back to the corresponding image coordinate convention before refinement. This ensures that the OSE stage benefits from normalized coordinates, while the final refinement is evaluated in the intended geometric domain.

\begin{table}[t]
\centering
\caption{\textbf{Median runtime decomposition.} Values are seconds and are computed over all datasets, initialization distributions, and the non-robust/Cauchy variants. The pre-stage-2 column includes formulation-specific preprocessing and intermediate steps before reprojection refinement, including the RpOSE local radial refinement and the pOSE+rot pair/matrix construction.}
\label{tab:appendix_runtime_decomposition}
\scriptsize
\setlength{\tabcolsep}{3pt}
\renewcommand{\arraystretch}{1.12}
\resizebox{\columnwidth}{!}{%
\begin{tabular}{@{}lcccc@{}}
\toprule
\textbf{Formulation} & \textbf{Pre-stage-2} & \textbf{Stage 2} & \textbf{Solver total} & \textbf{Wall-clock} \\
\midrule
pOSE & 4.64 & 4.15 & 8.87 & 9.38 \\
rOSE & 4.79 & 7.35 & 10.36 & 10.93 \\
RpOSE & 15.39 & 3.79 & 19.43 & 19.98 \\
expOSE & 4.44 & 3.48 & 7.69 & 8.03 \\
pOSE+rot & 4.85 & 4.93 & 11.74 & 12.39 \\
\bottomrule
\end{tabular}}
\end{table}

\begin{table}[t]
\centering
\caption{\textbf{pOSE+rot with ground-truth and estimated relative rotations.} Results use the linear metric upgrade and are averaged over all datasets, initialization distributions, and the non-robust/Cauchy variants. Higher is better for AUC; lower is better for landmark RMSE.}
\label{tab:appendix_poserot_gt_vs_estimated}
\scriptsize
\setlength{\tabcolsep}{3pt}
\renewcommand{\arraystretch}{1.12}
\resizebox{\columnwidth}{!}{%
\begin{tabular}{@{}lcccccc@{}}
\toprule
\textbf{Variant} & \textbf{Rot. A5} & \textbf{Rot. A20} & \textbf{Trans. A5} & \textbf{Trans. A20} & \textbf{Lmk. RMSE} & \textbf{Wall time} \\
\midrule
pOSE+rot & 42.14 & 53.87 & 45.81 & 57.08 & 16.63 & 38.39 \\
pOSE+rot est. & 41.80 & 52.90 & 45.30 & 54.83 & 17.04 & 69.03 \\
\bottomrule
\end{tabular}}
\end{table}

\subsection{Runtime and pOSE+rot Variants}
\label{app:runtime_poserot}

We report additional runtime statistics in \Cref{tab:appendix_runtime_decomposition}. 
The reported pre-stage-2 time includes all formulation-specific operations before the reprojection refinement. 
This is important for a fair comparison: RpOSE includes its local radial refinement and matrix-completion step before Stage 2, while pOSE+rot includes the construction of camera pairs and the corresponding pairwise matrices. 
The second-stage time corresponds to the subsequent reprojection-based refinement, and the wall-clock time includes the full solver execution excluding external table generation.

For pOSE+rot, we also compare two sources of relative rotations. 
The first variant uses the relative rotations provided by the ground-truth camera poses, while the second estimates them from the image data before constructing the pOSE+rot objective. 
As shown in \Cref{tab:appendix_poserot_gt_vs_estimated}, the two variants give very similar global reconstruction accuracy. 
The estimated variant is slightly weaker on average, but the gap is small compared to the differences observed between formulations and initialization distributions. 
This suggests that the conclusions drawn from pOSE+rot with ground-truth relative rotations are not merely an artifact of using privileged relative-pose information, although estimating these rotations introduces additional computational overhead.

The pOSE+rot construction proceeds as follows. 
For each admissible camera pair, we first obtain a relative rotation, either from the ground-truth poses or from an estimated two-view relation. 
Given this relative rotation, the method triangulates the landmarks observed by the pair and forms a pairwise object-space residual. 
This residual defines a matrix $W_{ij}$ associated with the camera pair $(i,j)$, which contributes to the reduced VarPro system. 
The full pOSE+rot objective is then obtained by aggregating these pairwise contributions over all selected camera pairs. 
Thus, pOSE+rot adds geometric information before the reduced camera optimization, but this comes at the cost of an additional pair-construction and matrix-building stage.

\subsection{Cauchy Robustification of Projected Residuals}
\label{app:robust_varpro}

For fixed camera variables $u$, VarPro eliminates the linear variables by solving
\begin{equation}
    v^{*}(u)
    =
    \arg\min_v
    \left\lVert
    G(u)v - z(u)
    \right\rVert^{2}.
\end{equation}
This gives the projected residual
\begin{equation}
    \varepsilon^{*}(u)
    =
    \varepsilon(u,v^{*}(u)).
\end{equation}
We robustify the reduced problem with a Cauchy loss,
\begin{equation}
    \min_u
    \sum_k
    \rho\left(
    \left\lVert
    \varepsilon^{*}_k(u)
    \right\rVert^2
    \right).
\end{equation}

We use a first-order, truncated Triggs correction \cite{triggs1999bundle},\cite{zach2014robust}. Let
\begin{equation}
    w_k
    =
    \sqrt{
    \rho'\left(
    \left\lVert
    \varepsilon^{*}_k
    \right\rVert^2
    \right)
    } .
\end{equation}
Each projected residual and its Jacobian are reweighted as
\begin{equation}
    \varepsilon^{*}_k \leftarrow w_k \varepsilon^{*}_k,
    \qquad
    J^{*}_k \leftarrow w_k J^{*}_k .
\end{equation}
The reduced normal equations therefore become
\begin{equation}
    (WJ_u^{*})^{\top}(WJ_u^{*})\Delta u
    =
    -
    (WJ_u^{*})^{\top}W\varepsilon^{*},
\end{equation}
where $W$ is the diagonal matrix of robust weights. Since this only reweights residual and Jacobian blocks, the RW2 approximation and the reduced camera system keep the same structure as in the non-robust solver.

\section{Detailed Metrics}\label{app:metrics}

We evaluate each method at the level of the final reconstructed geometry, rather than only through the value of the optimized OSE objective. This is important because, as shown in the previous sections, a low OSE value does not necessarily imply a valid metric reconstruction. We therefore report pose and structure metrics both before and after metric upgrade whenever applicable.

\paragraph{Rotation accuracy.}
We measure camera orientation accuracy using pairwise relative rotations. Given estimated rotations $\{\hat{R}_i\}$ and ground-truth rotations $\{R_i^\star\}$, we compare the relative rotation between every valid camera pair:
\begin{equation}
    \Delta R_{ij}
    =
    \left(\hat{R}_i \hat{R}_j^\top\right)
    \left(R_i^\star {R_j^\star}^{\top}\right)^\top .
\end{equation}
The angular error is
\begin{equation}
    e^R_{ij}
    =
    \arccos\left(
    \frac{\operatorname{tr}(\Delta R_{ij}) - 1}{2}
    \right).
\end{equation}
Using relative rotations makes the metric invariant to the global rotation gauge of the reconstruction. We summarize the distribution of errors using AUC scores at thresholds $5^\circ$, $10^\circ$, and $20^\circ$.

\paragraph{Translation-direction accuracy.}
We also evaluate the recovered camera trajectory through pairwise translation directions. Let $\hat{c}_i$ and $c_i^\star$ denote the estimated and ground-truth camera centers. For each camera pair, we compare the directions
\begin{equation}
    \hat{t}_{ij} =
    \frac{\hat{c}_j - \hat{c}_i}{\|\hat{c}_j - \hat{c}_i\|},
    \qquad
    t^\star_{ij} =
    \frac{c^\star_j - c^\star_i}{\|c^\star_j - c^\star_i\|}.
\end{equation}
The translation-direction error is
\begin{equation}
    e^t_{ij}
    =
    \arccos\left(
    \left| \hat{t}_{ij}^{\top} t^\star_{ij} \right|
    \right),
\end{equation}
where the absolute value removes the sign ambiguity of the translation direction. As for rotations, we report AUC at $5^\circ$, $10^\circ$, and $20^\circ$.

\paragraph{AUC computation.}
For both rotation and translation, we use the standard pose-estimation AUC computed from the empirical cumulative distribution of angular errors up to a threshold $\tau$. Given errors $\{e_k\}$, the AUC at threshold $\tau$ is
\begin{equation}
    \operatorname{AUC}@\tau
    =
    \frac{1}{\tau}
    \int_0^\tau
    \frac{1}{N}
    \sum_{k=1}^{N}
    \mathds{1}[e_k \leq \theta]
    \, d\theta .
\end{equation}
This metric rewards not only the number of estimates below the threshold, but also how small their errors are. Higher values indicate better pose accuracy.

\paragraph{Landmark accuracy.}
To evaluate the recovered structure, we compare estimated landmarks to the ground-truth landmarks that remain after the same observation filtering used in the optimization. Since reconstructions are defined only up to a global similarity transform, we first align the estimated landmark cloud to the ground truth using a similarity alignment. In details, let $\hat X_j \in \mathbb{R}^3$ denote the reconstructed landmark and $X_j \in \mathbb{R}^3$ the corresponding ground-truth landmark. We estimate a similarity transform $(s,R,t)$ by solving
\[
\min_{s,R,t} \sum_{j \in \mathcal{V}}
\left\| s R \hat X_j + t - X_j \right\|^2,
\]
where $R \in SO(3)$, $s>0$, and $\mathcal{V}$ is the set of landmarks present in both reconstructions after visibility filtering. We solve this Procrustes/Umeyama alignment in closed form. Landmark errors are then computed as
\[
e_j =
\left\| s R \hat X_j + t - X_j \right\|_2 .
\]

We then compute Euclidean landmark errors and report the median error, root mean squared error (RMSE), and the $90$th percentile error. These metrics complement the pose AUCs by measuring whether the recovered cameras also support an accurate 3D reconstruction.

\section{Results for Selected metric upgrade} \label{app:selected_metric_ugprade}

\begin{figure}[t]
\centering
\includegraphics[width=\columnwidth]{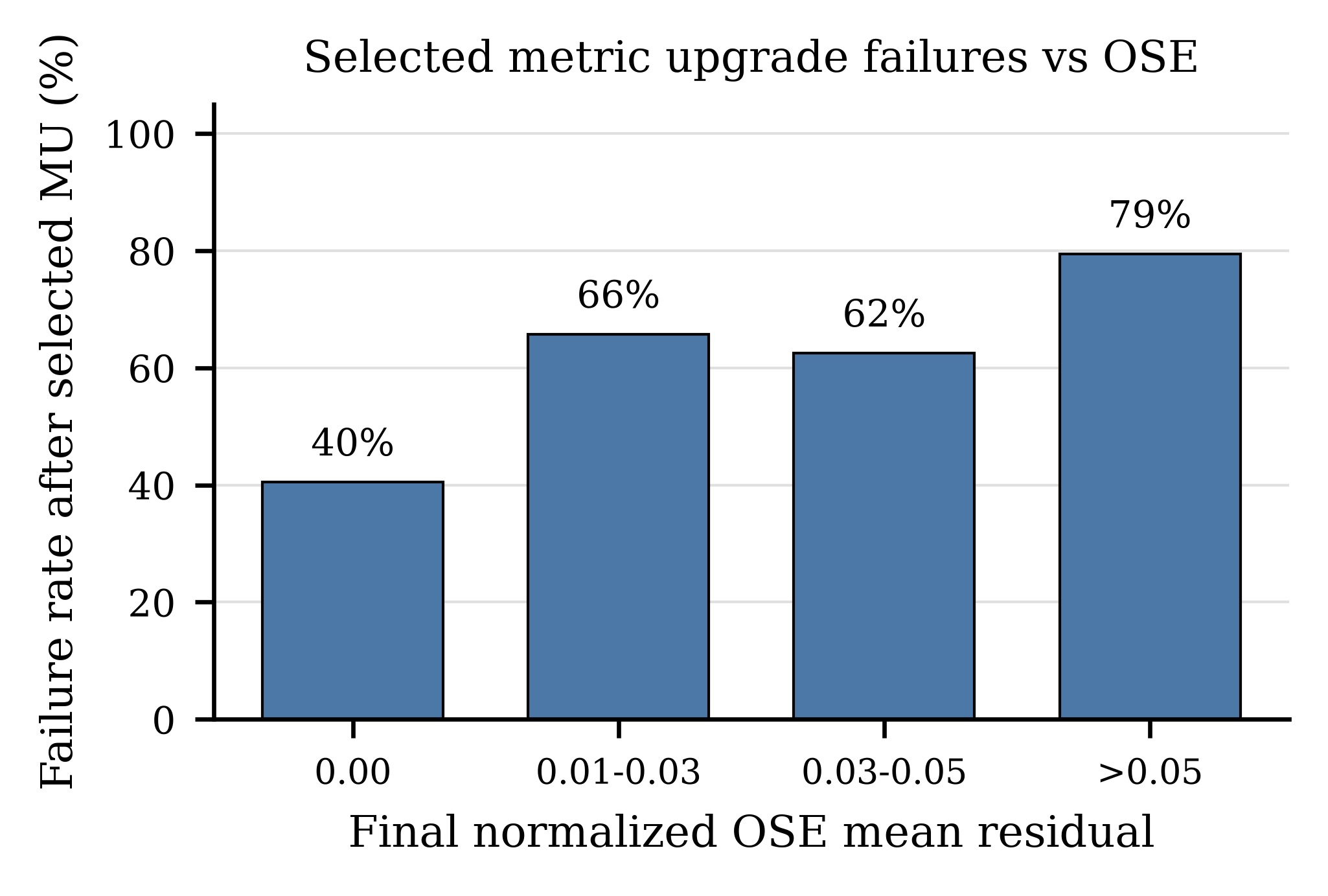}
\caption{\textbf{Metric upgrade failure rate as a function of the final OSE residual.}
We consider non-robust and Cauchy losses, followed by selected metric upgrade. A run is counted as failed when Rot.~AUC@20 $<20$, Trans.~AUC@20 $<20$, or landmark RMSE $>30$.}
\label{fig:metric_upgrade_failure_rate_ose_selected}
\end{figure}

\begin{figure}[t]
\centering
\includegraphics[width=\columnwidth]{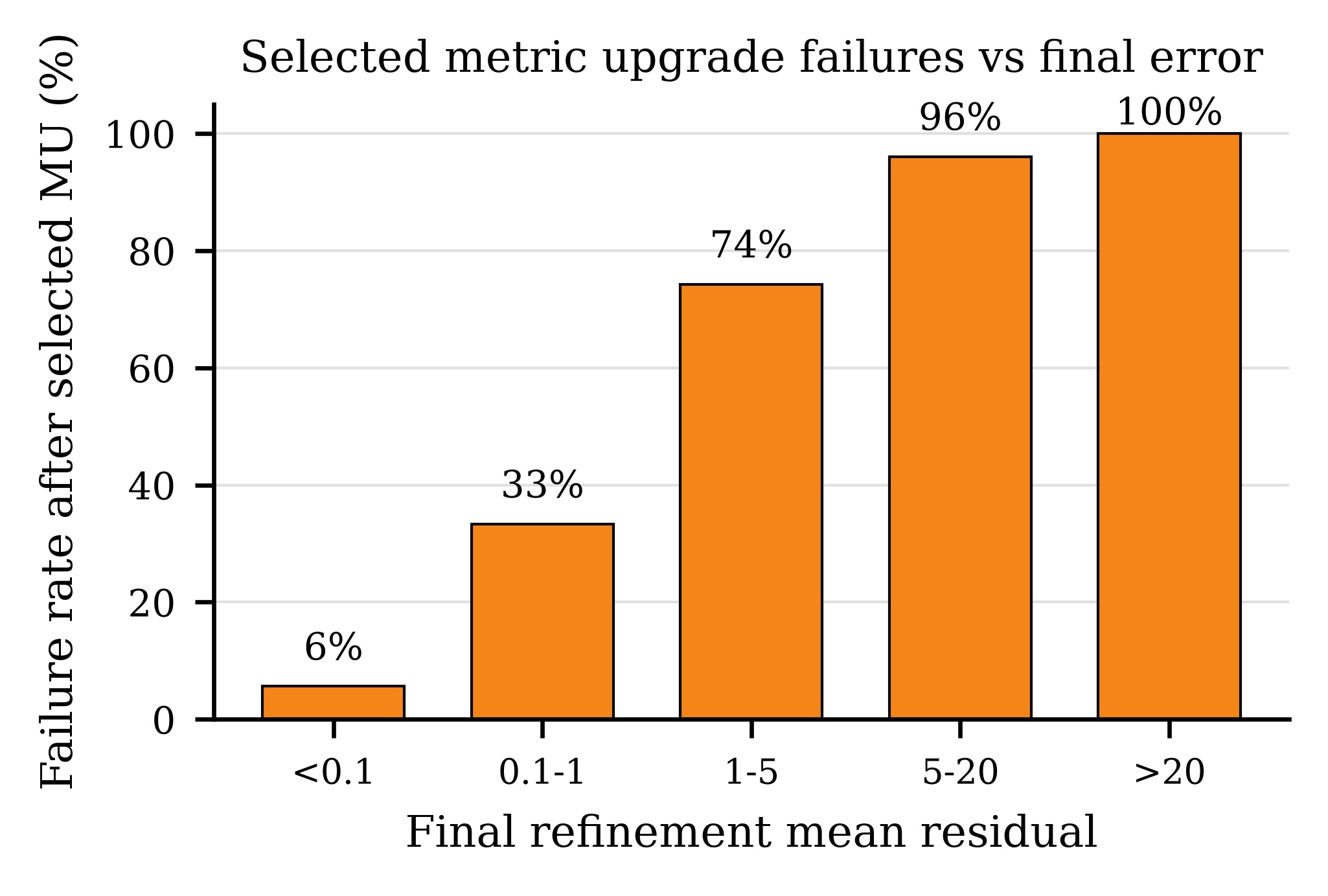}
\caption{\textbf{Metric upgrade failure rate as a function of the final refinement residual.}
We use the final mean residual after the refinement optimization and evaluate failure after selected metric upgrade with the same criterion as in \cref{fig:metric_upgrade_failure_rate_ose_selected}.}
\label{fig:metric_upgrade_failure_rate_refinement_selected}
\end{figure}

As explained in the main paper, Pollefeys et al. \cite{pollefeys1999self} propose not only a linear metric upgrade, but also a nonlinear approach that can be used as a refinement after a linear initialization. In practice, and as noted by Pollefeys et al., this refinement can lead to unstable results. Alongside the linear metric upgrade proposed in the main experiments, we also implement this nonlinear refinement. The selected metric upgrade performs the best metric upgrade between the linear approach and the nonlinear refinement based on the cheirality of the resulting solution. Following \Cref{tab:summary_formulations_pose_selected1997,tab:summary_formulations_landmarks_selected1997,tab:init_average_selected1997,tab:robust_average_selected1997_compact}, our main conclusions remain valid with this metric upgrade strategy. Notably, the percentage of failure cases in \Cref{fig:metric_upgrade_failure_rate_ose_selected,fig:metric_upgrade_failure_rate_refinement_selected} is very close to the results of the linear metric upgrade (\Cref{fig:metric_upgrade_failure_rate_ose,fig:metric_upgrade_failure_rate_refinement}), meaning that the main issues come from the output of the optimization itself, and not from the metric upgrade step.

\begin{table}[t]
\centering
\caption{\textbf{Average pose accuracy by formulation.}
Results are averaged over all datasets, initialization distributions, and non-robust/Cauchy variants after the selected linear/refined metric upgrade. Higher is better.}
\label{tab:summary_formulations_pose_selected1997}
\scriptsize
\setlength{\tabcolsep}{4pt}
\renewcommand{\arraystretch}{1.12}
\resizebox{\columnwidth}{!}{%
\begin{tabular}{@{}lcccc@{}}
\toprule
\textbf{Formulation}
& \textbf{Rot. A5}
& \textbf{Rot. A20}
& \textbf{Trans. A5}
& \textbf{Trans. A20} \\
\midrule
pOSE      & 24.89 & 34.15 & 26.73 & 35.64 \\
rOSE      & 24.95 & 33.48 & 31.24 & 41.45 \\
RpOSE     & 9.18  & 15.01 & 12.49 & 20.36 \\
expOSE    & 25.08 & 35.04 & 26.68 & 37.28 \\
pOSE+rot  & \textbf{41.99} & \textbf{54.16} & \textbf{46.08} & \textbf{57.63} \\
\bottomrule
\end{tabular}}
\end{table}

\begin{table}[t]
\centering
\caption{\textbf{Average structure accuracy by formulation.}
Results are averaged over all datasets, initialization distributions, and non-robust/Cauchy variants after the selected linear/refined metric upgrade. Lower is better.}
\label{tab:summary_formulations_landmarks_selected1997}
\scriptsize
\setlength{\tabcolsep}{5pt}
\renewcommand{\arraystretch}{1.12}
\resizebox{\columnwidth}{!}{%
\begin{tabular}{@{}lccc@{}}
\toprule
\textbf{Formulation}
& \textbf{Lmk. Med.} $\downarrow$
& \textbf{Lmk. RMSE} $\downarrow$
& \textbf{Lmk. P90} $\downarrow$ \\
\midrule
pOSE      & 21.06 & 22.80 & 31.03 \\
rOSE      & 18.67 & 20.06 & 27.30 \\
RpOSE     & 28.01 & 29.62 & 40.75 \\
expOSE    & 18.39 & 20.37 & 28.14 \\
pOSE+rot  & \textbf{15.34} & \textbf{16.67} & \textbf{22.44} \\
\bottomrule
\end{tabular}}
\end{table}

\begin{table}[t]
\centering
\caption{\textbf{Average effect of initialization.}
Results are averaged over all datasets, formulations, and non-robust/Cauchy variants after the selected linear/refined metric upgrade. Higher is better for AUC; lower is better for landmark RMSE.}
\label{tab:init_average_selected1997}
\scriptsize
\setlength{\tabcolsep}{4pt}
\renewcommand{\arraystretch}{1.12}
\resizebox{\columnwidth}{!}{%
\begin{tabular}{@{}lccccc@{}}
\toprule
\textbf{Initialization}
& \textbf{Rot. A5}
& \textbf{Rot. A20}
& \textbf{Trans. A5}
& \textbf{Trans. A20}
& \textbf{Lmk. RMSE} $\downarrow$ \\
\midrule
Normal      & 24.30 & 30.29 & 27.48 & 34.62 & 22.81 \\
Uniform     & 23.30 & 30.38 & 26.57 & 34.68 & 23.81 \\
Unit circle & \textbf{28.06} & \textbf{42.43} & \textbf{31.89} & \textbf{46.10} & \textbf{19.09} \\
\bottomrule
\end{tabular}}
\end{table}

\begin{table}[t]
\centering
\caption{\textbf{Average effect of robustification.}
Results are averaged over all datasets, formulations, and initialization distributions after the selected linear/refined metric upgrade. Higher is better for AUC; lower is better for landmark errors.}
\label{tab:robust_average_selected1997_compact}
\scriptsize
\setlength{\tabcolsep}{4pt}
\renewcommand{\arraystretch}{1.12}
\resizebox{\columnwidth}{!}{%
\begin{tabular}{@{}lccccc@{}}
\toprule
\textbf{Loss}
& \textbf{Rot. A20}
& \textbf{Trans. A20}
& \textbf{Lmk. Med.} $\downarrow$
& \textbf{Lmk. RMSE} $\downarrow$
& \textbf{Lmk. P90} $\downarrow$ \\
\midrule
None    & \textbf{34.87} & 37.69 & 20.53 & 22.06 & 30.22 \\
Cauchy  & 33.87 & \textbf{39.24} & \textbf{20.07} & \textbf{21.74} & \textbf{29.65} \\
\bottomrule
\end{tabular}}
\end{table}

\section{Full Per-Dataset Results}
\label{app:full_results}

This appendix reports the full per-dataset AUC and landmark tables. 
The main paper reports compact averages to keep the analysis readable.

\begin{table*}[t]
\centering
\caption{\textbf{Rotation AUC results without robust loss. Each entry reports before/linear/selected metric upgrade.}}
\label{tab:ose_rot_none_current}
\scriptsize
\setlength{\tabcolsep}{2.2pt}
\renewcommand{\arraystretch}{1.12}
\resizebox{\textwidth}{!}{%
\begin{tabular}{ll|ccc|ccc|ccc|ccc|ccc}
\toprule
& & \multicolumn{3}{c|}{\textbf{pOSE}} & \multicolumn{3}{c|}{\textbf{rOSE}} & \multicolumn{3}{c|}{\textbf{RpOSE}} & \multicolumn{3}{c|}{\textbf{expOSE}} & \multicolumn{3}{c}{\textbf{pOSE+rot}}\\
\cmidrule(lr){3-5}\cmidrule(lr){6-8}\cmidrule(lr){9-11}\cmidrule(lr){12-14}\cmidrule(l){15-17}
\textbf{Dataset} & \textbf{Initialization} & A5 & A10 & A20 & A5 & A10 & A20 & A5 & A10 & A20 & A5 & A10 & A20 & A5 & A10 & A20\\
\midrule
\multirow{3}{*}{Set1} & Normal & 0.2/86.6/86.6\% & 1.7/93.3/93.3\% & 5.5/96.6/96.6\% & 0.0/76.8/76.8\% & 1.4/88.4/88.4\% & 4.5/94.2/94.2\% & 1.3/0.1/0.0\% & 4.3/1.9/1.2\% & 10.5/5.4/3.6\% & 0.2/71.2/71.2\% & 1.8/85.6/85.6\% & 5.2/92.8/92.8\% & 0.4/56.3/56.3\% & 2.5/78.1/78.1\% & 6.7/89.1/89.1\%\\
 & Uniform & 0.9/83.1/83.1\% & 3.4/91.6/91.6\% & 8.7/95.8/95.8\% & 2.6/70.8/70.8\% & 6.7/85.4/85.4\% & 15.0/92.7/92.7\% & 0.8/1.7/1.7\% & 3.6/5.2/5.2\% & 10.3/12.6/12.6\% & 1.8/88.5/88.5\% & 5.1/94.2/94.2\% & 12.1/97.1/97.1\% & 2.0/90.8/90.8\% & 5.8/95.4/95.4\% & 14.9/97.7/97.7\%\\
 & Unit circle & 81.8/6.2/6.2\% & 90.9/14.9/14.9\% & 95.5/34.4/34.4\% & 17.6/5.6/5.6\% & 50.1/13.5/13.5\% & 75.0/31.6/31.6\% & 86.3/0.0/0.0\% & 93.1/1.3/1.4\% & 96.6/3.9/4.1\% & 83.5/8.0/8.0\% & 91.8/19.4/19.4\% & 95.9/41.4/41.4\% & 32.2/18.5/18.5\% & 63.0/37.8/37.8\% & 81.5/64.6/64.6\%\\
\midrule
\multirow{3}{*}{Set2} & Normal & 0.0/0.0/0.0\% & 0.1/0.0/0.0\% & 0.2/0.4/0.4\% & 0.4/0.0/0.0\% & 2.7/1.2/1.2\% & 9.0/3.7/3.7\% & 0.0/0.0/0.0\% & 0.4/0.4/0.4\% & 1.9/1.6/1.6\% & 0.1/0.0/0.0\% & 0.8/0.4/0.4\% & 2.7/1.4/1.4\% & 3.0/89.5/89.5\% & 7.2/94.7/94.7\% & 15.4/97.4/97.4\%\\
 & Uniform & 0.0/0.0/0.0\% & 0.5/1.1/1.0\% & 2.5/3.3/3.2\% & 0.0/0.0/0.0\% & 0.4/0.7/0.7\% & 2.2/2.6/2.6\% & 0.0/0.0/0.0\% & 0.3/0.5/0.5\% & 1.8/1.5/1.5\% & 0.0/0.0/0.0\% & 0.2/1.2/1.2\% & 1.8/3.6/3.6\% & 0.1/0.0/0.0\% & 0.6/0.7/0.7\% & 2.3/2.5/2.3\%\\
\bottomrule
\end{tabular}}
\end{table*}

\begin{table*}[t]
\centering
\caption{\textbf{Rotation AUC results with Cauchy loss. Each entry reports before/linear/selected metric upgrade.}}
\label{tab:ose_rot_cauchy_current}
\scriptsize
\setlength{\tabcolsep}{2.2pt}
\renewcommand{\arraystretch}{1.12}
\resizebox{\textwidth}{!}{%
\begin{tabular}{ll|ccc|ccc|ccc|ccc|ccc}
\toprule
& & \multicolumn{3}{c|}{\textbf{pOSE}} & \multicolumn{3}{c|}{\textbf{rOSE}} & \multicolumn{3}{c|}{\textbf{RpOSE}} & \multicolumn{3}{c|}{\textbf{expOSE}} & \multicolumn{3}{c}{\textbf{pOSE+rot}}\\
\cmidrule(lr){3-5}\cmidrule(lr){6-8}\cmidrule(lr){9-11}\cmidrule(lr){12-14}\cmidrule(l){15-17}
\textbf{Dataset} & \textbf{Initialization} & A5 & A10 & A20 & A5 & A10 & A20 & A5 & A10 & A20 & A5 & A10 & A20 & A5 & A10 & A20\\
\midrule
\multirow{3}{*}{Set1} & Normal & 0.7/0.6/0.0\% & 3.0/2.8/1.5\% & 10.1/7.0/4.5\% & 0.5/42.7/42.7\% & 2.5/69.0/69.0\% & 7.4/84.5/84.5\% & 0.8/0.0/0.0\% & 3.4/1.4/1.4\% & 11.2/4.2/4.2\% & 0.2/26.3/26.3\% & 1.7/48.9/48.9\% & 5.6/73.8/73.8\% & 0.1/79.3/79.3\% & 1.5/89.7/89.7\% & 4.6/94.8/94.8\%\\
 & Uniform & 0.6/19.6/19.6\% & 3.2/39.3/39.3\% & 10.5/66.0/66.0\% & 3.0/64.4/44.7\% & 7.6/82.2/71.5\% & 16.4/91.1/85.7\% & 0.1/0.1/0.1\% & 0.8/0.9/0.9\% & 3.3/3.5/3.5\% & 0.6/35.7/2.8\% & 3.1/61.2/7.2\% & 10.1/80.6/16.7\% & 0.1/19.8/19.8\% & 1.9/37.8/37.8\% & 7.5/63.6/63.6\%\\
 & Unit circle & 82.2/5.7/5.7\% & 91.1/13.7/13.7\% & 95.6/32.0/32.0\% & 17.6/5.9/5.9\% & 49.9/14.1/14.1\% & 75.0/32.8/32.8\% & 86.3/0.0/0.0\% & 93.1/1.3/1.3\% & 96.6/3.9/4.0\% & 83.5/7.2/7.2\% & 91.8/17.4/17.4\% & 95.9/38.5/38.5\% & 31.4/18.3/18.3\% & 62.3/37.5/37.5\% & 81.2/64.3/64.3\%\\
\midrule
\multirow{3}{*}{Set2} & Normal & 0.1/91.4/91.4\% & 1.1/95.7/95.7\% & 3.9/97.8/97.8\% & 3.1/85.7/85.7\% & 9.7/92.9/92.9\% & 35.5/96.4/96.4\% & 0.0/0.0/0.0\% & 0.5/0.8/0.8\% & 2.3/3.0/3.0\% & 0.5/0.0/0.0\% & 2.7/1.2/1.2\% & 7.2/3.6/3.6\% & 0.1/0.0/0.0\% & 0.6/0.7/0.4\% & 2.4/2.4/1.9\%\\
 & Uniform & 0.0/0.0/0.0\% & 0.5/1.1/1.2\% & 2.7/3.5/3.7\% & 0.0/0.0/0.0\% & 0.3/1.1/1.0\% & 2.3/3.3/3.2\% & 0.1/0.0/0.0\% & 0.8/0.8/0.8\% & 3.3/3.0/3.0\% & 0.0/0.1/0.0\% & 0.6/0.4/0.5\% & 2.6/1.9/2.2\% & 0.8/97.2/97.2\% & 3.3/98.6/98.6\% & 8.5/99.3/99.3\%\\
\bottomrule
\end{tabular}}
\end{table*}

\begin{table*}[t]
\centering
\caption{\textbf{Translation-direction AUC results without robust loss. Each entry reports before/linear/selected metric upgrade.}}
\label{tab:ose_trans_none_current}
\scriptsize
\setlength{\tabcolsep}{2.2pt}
\renewcommand{\arraystretch}{1.12}
\resizebox{\textwidth}{!}{%
\begin{tabular}{ll|ccc|ccc|ccc|ccc|ccc}
\toprule
& & \multicolumn{3}{c|}{\textbf{pOSE}} & \multicolumn{3}{c|}{\textbf{rOSE}} & \multicolumn{3}{c|}{\textbf{RpOSE}} & \multicolumn{3}{c|}{\textbf{expOSE}} & \multicolumn{3}{c}{\textbf{pOSE+rot}}\\
\cmidrule(lr){3-5}\cmidrule(lr){6-8}\cmidrule(lr){9-11}\cmidrule(lr){12-14}\cmidrule(l){15-17}
\textbf{Dataset} & \textbf{Initialization} & T5 & T10 & T20 & T5 & T10 & T20 & T5 & T10 & T20 & T5 & T10 & T20 & T5 & T10 & T20\\
\midrule
\multirow{3}{*}{Set1} & Normal & 0.2/93.8/93.8\% & 0.5/96.9/96.9\% & 1.9/98.5/98.5\% & 0.2/89.3/89.3\% & 1.0/94.6/94.6\% & 4.1/97.3/97.3\% & 0.2/1.8/1.9\% & 1.1/3.3/3.6\% & 3.8/6.4/7.2\% & 1.5/86.6/86.6\% & 6.7/93.3/93.3\% & 19.5/96.7/96.7\% & 0.9/79.9/79.9\% & 4.0/89.9/89.9\% & 11.2/95.0/95.0\%\\
 & Uniform & 0.4/92.2/92.2\% & 1.2/96.1/96.1\% & 3.3/98.1/98.1\% & 0.2/86.5/86.5\% & 0.5/93.3/93.3\% & 2.1/96.6/96.6\% & 0.6/12.1/12.1\% & 1.6/21.4/21.4\% & 3.6/36.6/36.6\% & 0.2/94.7/94.7\% & 0.8/97.3/97.3\% & 3.1/98.7/98.7\% & 0.3/95.7/95.7\% & 1.3/97.9/97.9\% & 4.1/98.9/98.9\%\\
 & Unit circle & 1.0/26.9/26.9\% & 4.2/43.5/43.5\% & 10.7/66.8/66.8\% & 0.0/25.6/25.6\% & 0.2/41.5/41.5\% & 1.5/64.3/64.3\% & 2.0/2.0/1.8\% & 4.2/3.8/3.7\% & 8.0/7.5/7.3\% & 0.3/30.2/30.2\% & 1.5/48.6/48.6\% & 4.5/72.1/72.1\% & 1.4/43.8/43.8\% & 3.1/68.1/68.1\% & 6.6/84.0/84.0\%\\
\midrule
\multirow{3}{*}{Set2} & Normal & 0.1/0.7/0.7\% & 0.2/1.7/1.7\% & 1.1/3.9/3.9\% & 0.1/1.6/1.6\% & 0.3/3.5/3.5\% & 0.7/7.0/7.0\% & 0.2/0.6/0.6\% & 0.4/2.2/2.2\% & 1.0/5.5/5.5\% & 0.2/0.2/0.2\% & 0.3/0.8/0.8\% & 1.1/2.8/2.8\% & 0.2/95.1/95.1\% & 0.8/97.6/97.6\% & 2.9/98.8/98.8\%\\
 & Uniform & 0.2/2.2/2.0\% & 0.7/4.3/3.5\% & 1.7/8.3/6.6\% & 0.1/0.5/0.5\% & 0.1/2.6/2.6\% & 0.5/7.4/7.4\% & 0.1/0.6/0.6\% & 0.3/1.8/1.8\% & 1.0/4.9/4.9\% & 0.1/1.6/1.6\% & 0.3/3.5/3.5\% & 2.4/7.2/7.2\% & 0.6/2.1/1.5\% & 1.8/4.5/3.9\% & 5.9/9.9/8.9\%\\
\bottomrule
\end{tabular}}
\end{table*}

\begin{table*}[t]
\centering
\caption{\textbf{Translation-direction AUC results with Cauchy loss. Each entry reports before/linear/selected metric upgrade.}}
\label{tab:ose_trans_cauchy_current}
\scriptsize
\setlength{\tabcolsep}{2.2pt}
\renewcommand{\arraystretch}{1.12}
\resizebox{\textwidth}{!}{%
\begin{tabular}{ll|ccc|ccc|ccc|ccc|ccc}
\toprule
& & \multicolumn{3}{c|}{\textbf{pOSE}} & \multicolumn{3}{c|}{\textbf{rOSE}} & \multicolumn{3}{c|}{\textbf{RpOSE}} & \multicolumn{3}{c|}{\textbf{expOSE}} & \multicolumn{3}{c}{\textbf{pOSE+rot}}\\
\cmidrule(lr){3-5}\cmidrule(lr){6-8}\cmidrule(lr){9-11}\cmidrule(lr){12-14}\cmidrule(l){15-17}
\textbf{Dataset} & \textbf{Initialization} & T5 & T10 & T20 & T5 & T10 & T20 & T5 & T10 & T20 & T5 & T10 & T20 & T5 & T10 & T20\\
\midrule
\multirow{3}{*}{Set1} & Normal & 0.1/2.7/2.2\% & 0.2/4.8/4.2\% & 1.3/9.3/8.4\% & 0.3/71.5/71.5\% & 1.0/85.8/85.8\% & 4.0/92.9/92.9\% & 0.1/1.9/1.9\% & 0.9/3.8/3.8\% & 3.4/7.6/7.6\% & 0.2/53.4/53.4\% & 0.7/76.2/76.2\% & 3.6/88.1/88.1\% & 0.1/90.5/90.5\% & 0.2/95.2/95.2\% & 1.0/97.6/97.6\%\\
 & Uniform & 0.1/44.8/44.8\% & 0.5/69.2/69.2\% & 4.0/84.6/84.6\% & 3.9/83.5/76.3\% & 8.1/91.7/88.2\% & 15.3/95.9/94.1\% & 0.2/0.1/0.1\% & 0.3/0.5/0.5\% & 1.5/3.8/3.8\% & 0.2/63.9/33.2\% & 0.7/81.9/54.7\% & 3.0/91.0/77.1\% & 0.9/41.9/41.9\% & 3.5/66.1/66.1\% & 8.7/83.0/83.0\%\\
 & Unit circle & 1.0/25.8/25.8\% & 4.1/41.8/41.8\% & 10.5/64.6/64.6\% & 0.0/26.2/26.2\% & 0.2/42.4/42.4\% & 1.8/65.4/65.4\% & 2.0/1.9/1.8\% & 4.2/3.7/3.7\% & 8.0/7.4/7.3\% & 0.3/28.8/28.8\% & 1.6/46.4/46.4\% & 4.5/70.0/70.0\% & 1.4/43.5/43.5\% & 3.0/67.8/67.8\% & 6.4/83.9/83.9\%\\
\midrule
\multirow{3}{*}{Set2} & Normal & 0.3/96.0/96.0\% & 1.0/98.0/98.0\% & 3.4/99.0/99.0\% & 0.1/93.4/93.4\% & 0.3/96.7/96.7\% & 1.7/98.3/98.3\% & 1.0/1.7/1.7\% & 2.1/4.0/4.0\% & 4.7/7.7/7.7\% & 0.1/1.8/1.8\% & 0.4/3.6/3.6\% & 1.4/7.2/7.2\% & 0.1/1.5/0.5\% & 0.1/3.1/2.1\% & 1.1/5.8/6.6\%\\
 & Uniform & 0.1/1.9/1.7\% & 0.3/3.7/3.4\% & 0.5/7.4/6.9\% & 0.1/1.4/1.7\% & 0.2/3.6/3.6\% & 0.6/7.5/7.4\% & 0.1/0.3/0.3\% & 0.5/1.2/1.2\% & 2.3/4.0/4.0\% & 0.4/1.2/0.5\% & 1.3/3.1/2.2\% & 3.5/7.0/6.0\% & 0.6/98.7/98.7\% & 1.9/99.3/99.3\% & 5.6/99.7/99.7\%\\
\bottomrule
\end{tabular}}
\end{table*}

\begin{table*}[t]
\centering
\caption{\textbf{Landmark accuracy without robust loss. Each entry reports before/linear/selected metric upgrade; lower is better.}}
\label{tab:landmark_none_current}
\scriptsize
\setlength{\tabcolsep}{2.2pt}
\renewcommand{\arraystretch}{1.12}
\resizebox{\textwidth}{!}{%
\begin{tabular}{ll|ccc|ccc|ccc|ccc|ccc}
\toprule
& & \multicolumn{3}{c|}{\textbf{pOSE}} & \multicolumn{3}{c|}{\textbf{rOSE}} & \multicolumn{3}{c|}{\textbf{RpOSE}} & \multicolumn{3}{c|}{\textbf{expOSE}} & \multicolumn{3}{c}{\textbf{pOSE+rot}}\\
\cmidrule(lr){3-5}\cmidrule(lr){6-8}\cmidrule(lr){9-11}\cmidrule(lr){12-14}\cmidrule(l){15-17}
\textbf{Dataset} & \textbf{Initialization} & Med & RMSE & P90 & Med & RMSE & P90 & Med & RMSE & P90 & Med & RMSE & P90 & Med & RMSE & P90\\
\midrule
\multirow{3}{*}{Set1} & Normal & 27.89/0.13/0.13 & 32.26/0.17/0.17 & 45.94/0.24/0.24 & 21.73/0.23/0.23 & 26.31/0.29/0.29 & 38.77/0.41/0.41 & 28.22/39.68/39.68 & 32.39/41.16/41.16 & 44.84/58.11/58.11 & 32.51/0.29/0.29 & 35.84/0.36/0.36 & 51.25/0.52/0.52 & 24.64/0.44/0.44 & 26.34/0.54/0.54 & 34.17/0.78/0.78\\
 & Uniform & 30.02/0.17/0.17 & 33.92/0.21/0.21 & 45.62/0.30/0.30 & 22.07/0.29/0.29 & 27.38/0.36/0.36 & 38.55/0.52/0.52 & 27.49/7.19/7.19 & 31.32/8.85/8.85 & 43.70/13.31/13.31 & 27.39/0.12/0.12 & 31.82/0.14/0.14 & 42.62/0.20/0.20 & 22.26/0.09/0.09 & 26.60/0.11/0.11 & 35.64/0.16/0.16\\
 & Unit circle & 0.94/3.87/3.87 & 1.13/4.32/4.32 & 1.56/6.24/6.24 & 2.99/4.20/4.20 & 3.98/4.70/4.70 & 5.89/6.78/6.78 & 3.01/39.68/39.68 & 5.26/41.16/41.16 & 8.38/58.11/58.11 & 2.10/3.17/3.17 & 2.43/3.54/3.54 & 3.55/5.11/5.11 & 2.49/1.60/1.60 & 2.72/1.82/1.82 & 3.96/2.65/2.65\\
\midrule
\multirow{3}{*}{Set2} & Normal & 39.10/39.12/39.12 & 40.65/40.66/40.66 & 58.19/58.19/58.19 & 19.03/39.12/39.12 & 21.59/40.66/40.66 & 31.34/58.19/58.19 & 38.23/39.12/39.12 & 39.25/40.66/40.66 & 54.92/58.19/58.19 & 39.12/39.12/39.12 & 40.66/40.66/40.66 & 58.20/58.19/58.19 & 14.85/0.10/0.10 & 17.19/0.13/0.13 & 23.09/0.18/0.18\\
 & Uniform & 38.76/39.12/39.12 & 40.03/40.66/40.66 & 58.12/58.19/58.20 & 38.42/39.12/39.12 & 39.77/40.66/40.66 & 57.30/58.19/58.19 & 39.03/39.12/39.12 & 40.50/40.66/40.66 & 57.94/58.19/58.19 & 39.12/39.12/39.12 & 40.66/40.66/40.66 & 58.19/58.19/58.19 & 38.84/39.12/39.12 & 40.28/40.66/40.66 & 57.99/58.19/58.19\\
\bottomrule
\end{tabular}}
\end{table*}

\begin{table*}[t]
\centering
\caption{\textbf{Landmark accuracy with Cauchy loss. Each entry reports before/linear/selected metric upgrade; lower is better.}}
\label{tab:landmark_cauchy_current}
\scriptsize
\setlength{\tabcolsep}{2.2pt}
\renewcommand{\arraystretch}{1.12}
\resizebox{\textwidth}{!}{%
\begin{tabular}{ll|ccc|ccc|ccc|ccc|ccc}
\toprule
& & \multicolumn{3}{c|}{\textbf{pOSE}} & \multicolumn{3}{c|}{\textbf{rOSE}} & \multicolumn{3}{c|}{\textbf{RpOSE}} & \multicolumn{3}{c|}{\textbf{expOSE}} & \multicolumn{3}{c}{\textbf{pOSE+rot}}\\
\cmidrule(lr){3-5}\cmidrule(lr){6-8}\cmidrule(lr){9-11}\cmidrule(lr){12-14}\cmidrule(l){15-17}
\textbf{Dataset} & \textbf{Initialization} & Med & RMSE & P90 & Med & RMSE & P90 & Med & RMSE & P90 & Med & RMSE & P90 & Med & RMSE & P90\\
\midrule
\multirow{3}{*}{Set1} & Normal & 19.52/39.68/39.67 & 19.52/41.15/41.16 & 26.30/58.11/58.11 & 25.98/0.64/0.64 & 29.38/0.77/0.77 & 39.68/1.12/1.12 & 20.65/39.68/39.68 & 25.58/41.16/41.16 & 37.85/58.11/58.11 & 29.72/1.14/1.14 & 32.48/1.32/1.32 & 44.53/1.93/1.93 & 32.04/0.21/0.21 & 33.38/0.26/0.26 & 43.63/0.37/0.37\\
 & Uniform & 20.88/1.51/1.51 & 20.57/1.73/1.73 & 28.12/2.52/2.52 & 14.07/0.36/0.52 & 16.27/0.44/0.65 & 22.23/0.64/0.92 & 33.31/39.68/39.68 & 36.94/41.16/41.16 & 53.32/58.11/58.11 & 17.66/0.81/2.24 & 19.39/0.98/2.73 & 25.88/1.42/3.94 & 18.99/1.62/1.62 & 21.18/1.90/1.90 & 27.47/2.78/2.78\\
 & Unit circle & 0.94/4.14/4.14 & 1.13/4.63/4.63 & 1.55/6.69/6.69 & 3.00/4.05/4.05 & 3.99/4.53/4.53 & 5.90/6.54/6.54 & 3.01/39.68/39.68 & 5.26/41.16/41.16 & 8.38/58.11/58.11 & 2.12/3.43/3.43 & 2.44/3.84/3.84 & 3.57/5.53/5.53 & 2.53/1.61/1.61 & 2.77/1.83/1.83 & 4.04/2.68/2.68\\
\midrule
\multirow{3}{*}{Set2} & Normal & 27.78/0.09/0.09 & 31.38/0.10/0.10 & 44.39/0.15/0.15 & 6.38/0.14/0.14 & 8.07/0.17/0.17 & 12.05/0.25/0.25 & 36.04/39.12/39.12 & 37.71/40.66/40.66 & 52.40/58.18/58.18 & 26.61/39.12/39.12 & 27.55/40.66/40.66 & 34.47/58.19/58.19 & 38.66/39.12/39.12 & 39.99/40.66/40.66 & 57.89/58.19/58.19\\
 & Uniform & 38.61/39.12/39.12 & 39.59/40.66/40.66 & 54.64/58.19/58.20 & 38.62/39.12/39.12 & 39.90/40.66/40.66 & 57.95/58.19/58.19 & 36.73/39.12/39.12 & 38.51/40.66/40.66 & 54.20/58.19/58.19 & 38.94/39.12/39.12 & 40.46/40.66/40.66 & 57.60/58.19/58.19 & 30.97/0.03/0.03 & 33.76/0.03/0.03 & 44.19/0.05/0.05\\
\bottomrule
\end{tabular}}
\end{table*}

\begin{table*}[t]
\centering
\caption{\textbf{Rotation AUC with minimum 6 observations per landmark and Cauchy loss.} Each entry reports before/linear metric upgrade.}
\label{tab:minobs6_rot_cauchy_linear1997}
\scriptsize
\setlength{\tabcolsep}{2.5pt}
\renewcommand{\arraystretch}{1.15}
\resizebox{\textwidth}{!}{%
\begin{tabular}{ll|ccc|ccc|ccc|ccc|ccc}
\toprule
& & \multicolumn{3}{c|}{\textbf{pOSE}} & \multicolumn{3}{c|}{\textbf{rOSE}} & \multicolumn{3}{c|}{\textbf{RpOSE}} & \multicolumn{3}{c|}{\textbf{expOSE}} & \multicolumn{3}{c}{\textbf{pOSE+rot}}\\
\cmidrule(lr){3-5}\cmidrule(lr){6-8}\cmidrule(lr){9-11}\cmidrule(lr){12-14}\cmidrule(l){15-17}
\textbf{Dataset} & \textbf{Initialization} & A5&A10&A20 & A5&A10&A20 & A5&A10&A20 & A5&A10&A20 & A5&A10&A20\\
\midrule
\multirow{3}{*}{Set1} & Normal & 0.0/0.4\% & 1.5/2.4\% & 5.6/6.4\% & 0.3/43.4\% & 2.2/69.4\% & 6.9/84.7\% & 2.5/95.0\% & 7.0/97.5\% & 21.0/98.7\% & 0.7/33.0\% & 3.3/57.5\% & 10.3/78.7\% & 1.5/63.8\% & 5.7/81.9\% & 13.7/90.9\%\\
 & Uniform & 2.1/14.9\% & 8.7/30.2\% & 25.4/53.4\% & 3.0/74.2\% & 7.9/87.1\% & 18.2/93.6\% & 0.4/92.1\% & 2.5/96.1\% & 7.2/98.0\% & 1.1/58.8\% & 3.8/79.4\% & 9.2/89.7\% & 0.9/22.2\% & 3.6/41.6\% & 9.1/67.6\%\\
 & Unit circle & 82.2/8.4\% & 91.1/20.4\% & 95.6/42.9\% & 17.4/7.7\% & 49.7/18.6\% & 74.8/40.4\% & 86.2/0.0\% & 93.1/1.5\% & 96.5/4.3\% & 83.1/8.7\% & 91.5/21.2\% & 95.8/44.0\% & 30.9/21.3\% & 61.9/42.1\% & 80.9/68.6\%\\
\midrule
\multirow{3}{*}{Set2} & Normal & 0.0/0.0\% & 0.7/1.2\% & 3.0/3.5\% & 5.7/98.2\% & 14.8/99.1\% & 39.6/99.6\% & 0.8/99.4\% & 3.5/99.7\% & 9.0/99.9\% & 1.5/0.0\% & 4.7/1.5\% & 11.4/4.4\% & 1.7/84.9\% & 5.1/92.5\% & 12.3/96.2\%\\
 & Uniform & 1.4/85.7\% & 4.7/92.9\% & 14.0/96.4\% & 0.7/76.5\% & 3.0/88.2\% & 8.1/94.1\% & 0.0/0.0\% & 0.4/0.2\% & 2.6/1.3\% & 1.6/78.9\% & 5.0/89.5\% & 11.7/94.7\% & 1.4/85.5\% & 4.5/92.8\% & 11.1/96.4\%\\
 & Unit circle & 72.5/20.2\% & 86.2/38.7\% & 93.1/64.9\% & 12.0/16.2\% & 39.3/32.3\% & 69.2/56.6\% & 83.4/14.1\% & 91.7/29.1\% & 95.9/52.0\% & 74.2/18.2\% & 87.1/35.5\% & 93.5/61.1\% & 39.8/64.6\% & 68.8/82.3\% & 84.4/91.1\%\\
\bottomrule
\end{tabular}}
\end{table*}

\begin{table*}[t]
\centering
\caption{\textbf{Rotation AUC with minimum 6 observations per landmark without robust loss.} Each entry reports before/linear metric upgrade. Note that Set6 and Set7 are too sparse for supporting this densification.}
\label{tab:minobs6_rot_none_linear1997}
\scriptsize
\setlength{\tabcolsep}{2.5pt}
\renewcommand{\arraystretch}{1.15}
\resizebox{\textwidth}{!}{%
\begin{tabular}{ll|ccc|ccc|ccc|ccc|ccc}
\toprule
& & \multicolumn{3}{c|}{\textbf{pOSE}} & \multicolumn{3}{c|}{\textbf{rOSE}} & \multicolumn{3}{c|}{\textbf{RpOSE}} & \multicolumn{3}{c|}{\textbf{expOSE}} & \multicolumn{3}{c}{\textbf{pOSE+rot}}\\
\cmidrule(lr){3-5}\cmidrule(lr){6-8}\cmidrule(lr){9-11}\cmidrule(lr){12-14}\cmidrule(l){15-17}
\textbf{Dataset} & \textbf{Initialization} & A5&A10&A20 & A5&A10&A20 & A5&A10&A20 & A5&A10&A20 & A5&A10&A20\\
\midrule
\multirow{3}{*}{Set1} & Normal & 0.6/67.4\% & 3.0/83.7\% & 7.8/91.8\% & 1.0/78.7\% & 3.6/89.3\% & 9.0/94.7\% & 1.4/95.2\% & 4.4/97.6\% & 10.6/98.8\% & 1.1/81.8\% & 3.7/90.9\% & 9.3/95.5\% & 1.4/57.9\% & 4.5/78.9\% & 14.1/89.5\%\\
 & Uniform & 1.3/74.9\% & 4.1/87.5\% & 10.0/93.7\% & 2.9/65.9\% & 7.6/83.0\% & 17.8/91.5\% & 1.3/41.0\% & 4.4/67.2\% & 10.7/83.6\% & 2.6/70.0\% & 7.1/85.0\% & 16.7/92.5\% & 1.1/6.9\% & 3.9/16.9\% & 9.7/33.8\%\\
 & Unit circle & 81.8/8.9\% & 90.9/21.7\% & 95.5/44.7\% & 17.4/8.0\% & 49.8/19.5\% & 74.9/41.7\% & 86.2/0.0\% & 93.1/1.3\% & 96.6/3.9\% & 83.1/8.7\% & 91.5/21.3\% & 95.8/44.1\% & 31.7/17.7\% & 62.6/36.6\% & 81.3/63.4\%\\
\midrule
\multirow{3}{*}{Set2} & Normal & 2.1/92.7\% & 6.0/96.4\% & 14.1/98.2\% & 4.5/98.5\% & 11.8/99.3\% & 31.7/99.6\% & 0.1/0.0\% & 0.7/0.2\% & 2.8/1.3\% & 4.0/96.1\% & 9.9/98.1\% & 21.8/99.0\% & 2.2/54.3\% & 6.2/77.0\% & 14.7/88.5\%\\
 & Uniform & 0.0/0.0\% & 1.1/1.1\% & 3.7/3.4\% & 1.5/95.1\% & 4.9/97.6\% & 11.5/98.8\% & 0.0/0.0\% & 0.9/0.7\% & 3.0/2.6\% & 2.7/88.1\% & 6.9/94.0\% & 16.0/97.0\% & 1.7/94.1\% & 5.1/97.1\% & 12.2/98.5\%\\
 & Unit circle & 72.1/17.2\% & 86.1/34.0\% & 93.0/59.1\% & 12.1/23.0\% & 39.5/43.1\% & 69.3/69.2\% & 83.4/2.1\% & 91.7/6.3\% & 95.9/13.8\% & 74.2/18.6\% & 87.1/36.0\% & 93.5/61.8\% & 40.5/71.2\% & 69.2/85.6\% & 84.6/92.8\%\\
\bottomrule
\end{tabular}}
\end{table*}

\begin{table*}[t]
\centering
\caption{\textbf{Translation-direction AUC with minimum 6 observations per landmark with Cauchy loss.} Each entry reports before/linear metric upgrade.}
\label{tab:minobs6_trans_cauchy_linear1997}
\scriptsize
\setlength{\tabcolsep}{2.5pt}
\renewcommand{\arraystretch}{1.15}
\resizebox{\textwidth}{!}{%
\begin{tabular}{ll|ccc|ccc|ccc|ccc|ccc}
\toprule
& & \multicolumn{3}{c|}{\textbf{pOSE}} & \multicolumn{3}{c|}{\textbf{rOSE}} & \multicolumn{3}{c|}{\textbf{RpOSE}} & \multicolumn{3}{c|}{\textbf{expOSE}} & \multicolumn{3}{c}{\textbf{pOSE+rot}}\\
\cmidrule(lr){3-5}\cmidrule(lr){6-8}\cmidrule(lr){9-11}\cmidrule(lr){12-14}\cmidrule(l){15-17}
\textbf{Dataset} & \textbf{Initialization} & A5&A10&A20 & A5&A10&A20 & A5&A10&A20 & A5&A10&A20 & A5&A10&A20\\
\midrule
\multirow{3}{*}{Set1} & Normal & 0.1/6.9\% & 0.1/11.4\% & 0.6/18.3\% & 0.3/71.5\% & 1.6/85.8\% & 6.8/92.9\% & 0.1/97.7\% & 0.5/98.8\% & 3.7/99.4\% & 0.1/60.4\% & 0.5/80.2\% & 1.7/90.1\% & 0.6/83.2\% & 2.4/91.6\% & 6.9/95.8\%\\
 & Uniform & 0.1/34.9\% & 0.3/56.4\% & 0.9/78.0\% & 0.1/88.0\% & 0.4/94.0\% & 1.6/97.0\% & 0.1/96.4\% & 0.4/98.2\% & 1.8/99.1\% & 0.2/80.9\% & 1.0/90.4\% & 5.4/95.2\% & 5.5/45.4\% & 11.0/69.8\% & 19.7/84.9\%\\
 & Unit circle & 1.0/31.1\% & 4.1/49.8\% & 10.7/73.2\% & 0.0/29.8\% & 0.2/47.9\% & 1.6/71.4\% & 2.0/1.9\% & 4.2/3.8\% & 8.1/7.5\% & 0.3/31.6\% & 1.6/50.6\% & 4.6/73.8\% & 1.4/47.4\% & 2.9/71.6\% & 6.4/85.8\%\\
\midrule
\multirow{3}{*}{Set2} & Normal & 0.1/2.0\% & 0.3/3.8\% & 1.2/7.5\% & 0.0/99.2\% & 0.0/99.6\% & 0.5/99.8\% & 0.2/99.7\% & 0.5/99.9\% & 1.5/99.9\% & 0.2/1.5\% & 0.3/3.4\% & 0.5/7.3\% & 0.6/93.0\% & 1.6/96.5\% & 3.7/98.3\%\\
 & Uniform & 0.1/93.4\% & 0.4/96.7\% & 1.3/98.4\% & 0.3/89.1\% & 0.9/94.6\% & 3.3/97.3\% & 0.2/0.3\% & 0.5/1.0\% & 1.8/3.0\% & 1.1/90.2\% & 3.9/95.1\% & 11.2/97.6\% & 0.3/93.3\% & 1.1/96.6\% & 3.4/98.3\%\\
 & Unit circle & 0.7/42.8\% & 3.1/67.4\% & 8.5/83.7\% & 0.4/36.7\% & 1.3/59.6\% & 3.5/79.8\% & 1.5/33.8\% & 3.6/55.3\% & 7.3/77.4\% & 0.6/39.8\% & 1.9/63.9\% & 4.6/81.9\% & 1.0/83.6\% & 2.1/91.8\% & 4.5/95.9\%\\
\bottomrule
\end{tabular}}
\end{table*}

\begin{table*}[t]
\centering
\caption{\textbf{Translation-direction AUC with minimum 6 observations per landmark without robust loss.} Each entry reports before/linear metric upgrade.}
\label{tab:minobs6_trans_none_linear1997}
\scriptsize
\setlength{\tabcolsep}{2.5pt}
\renewcommand{\arraystretch}{1.15}
\resizebox{\textwidth}{!}{%
\begin{tabular}{ll|ccc|ccc|ccc|ccc|ccc}
\toprule
& & \multicolumn{3}{c|}{\textbf{pOSE}} & \multicolumn{3}{c|}{\textbf{rOSE}} & \multicolumn{3}{c|}{\textbf{RpOSE}} & \multicolumn{3}{c|}{\textbf{expOSE}} & \multicolumn{3}{c}{\textbf{pOSE+rot}}\\
\cmidrule(lr){3-5}\cmidrule(lr){6-8}\cmidrule(lr){9-11}\cmidrule(lr){12-14}\cmidrule(l){15-17}
\textbf{Dataset} & \textbf{Initialization} & A5&A10&A20 & A5&A10&A20 & A5&A10&A20 & A5&A10&A20 & A5&A10&A20\\
\midrule
\multirow{3}{*}{Set1} & Normal & 0.3/84.9\% & 1.3/92.4\% & 4.1/96.2\% & 2.2/90.1\% & 5.3/95.1\% & 13.2/97.5\% & 0.1/97.8\% & 1.6/98.9\% & 5.1/99.4\% & 0.3/91.6\% & 1.3/95.8\% & 3.7/97.9\% & 0.1/80.5\% & 0.3/90.2\% & 1.2/95.1\%\\
 & Uniform & 0.1/88.4\% & 0.3/94.2\% & 0.9/97.1\% & 0.2/84.2\% & 0.6/92.1\% & 2.1/96.0\% & 0.4/69.5\% & 1.5/84.8\% & 4.3/92.4\% & 0.3/86.1\% & 0.6/93.1\% & 1.8/96.5\% & 4.3/24.4\% & 11.2/39.7\% & 22.6/61.7\%\\
 & Unit circle & 1.0/32.0\% & 4.3/51.2\% & 10.8/74.3\% & 0.0/30.4\% & 0.1/48.9\% & 1.3/72.3\% & 2.0/2.0\% & 4.3/3.8\% & 8.1/7.4\% & 0.4/31.6\% & 1.6/50.7\% & 4.6/73.9\% & 1.4/42.7\% & 3.0/66.9\% & 6.6/83.5\%\\
\midrule
\multirow{3}{*}{Set2} & Normal & 0.8/96.7\% & 2.3/98.3\% & 5.8/99.2\% & 0.4/99.3\% & 1.4/99.7\% & 6.6/99.8\% & 0.2/0.2\% & 0.3/0.7\% & 0.9/2.2\% & 0.3/98.2\% & 1.1/99.1\% & 4.7/99.6\% & 0.7/78.7\% & 1.9/89.3\% & 4.5/94.7\%\\
 & Uniform & 0.6/1.6\% & 1.7/3.4\% & 4.0/6.7\% & 0.1/97.7\% & 0.5/98.9\% & 3.0/99.4\% & 0.2/1.6\% & 0.8/4.2\% & 2.0/9.6\% & 0.1/94.5\% & 0.5/97.3\% & 1.7/98.6\% & 0.7/97.3\% & 1.8/98.6\% & 4.0/99.3\%\\
 & Unit circle & 0.7/38.3\% & 3.2/62.0\% & 8.6/81.0\% & 0.4/47.0\% & 1.2/71.4\% & 3.4/85.7\% & 1.5/2.6\% & 3.6/4.4\% & 7.4/8.5\% & 0.6/40.2\% & 1.9/64.5\% & 4.6/82.2\% & 1.0/86.7\% & 2.1/93.4\% & 4.6/96.7\%\\
\bottomrule
\end{tabular}}
\end{table*}

\end{document}